\pdfoutput=1
\documentclass[acmsmall,screen,nonacm]{acmart}
\AtBeginDocument{%
  }

\setcopyright{none}
\usepackage{textcomp}
\usepackage{url}
\usepackage{booktabs}

\usepackage{newfloat}
\usepackage{graphicx}
\graphicspath{{./}}
\usepackage{listings}
\usepackage{enumitem}
\usepackage{bm}
\usepackage{array}
\usepackage{algorithm}
\usepackage[noend]{algorithmic}
\usepackage{amsmath}
\usepackage{bbding}
\usepackage{colortbl}
\usepackage{multirow}
\usepackage{multicol}
\usepackage{placeins}  
\usepackage{color}
\usepackage{xspace}

\usepackage{xcolor}

\newenvironment{yqc}{}{}

\usepackage[normalem]{ulem}
\colorlet{cladd}{purple}  
\definecolor{clnote}{HTML}{D55E00}
\newif\ifclfinal
\clfinaltrue
\ifclfinal
  \newenvironment{claude}{}{}
  
  \newcommand{\cldel}[1]{}
  
  \newcommand{\clnote}[1]{}
\else
  \newcommand{\cldel}[1]{{\color{gray}\sout{#1}}}          
  \newcommand{\clnote}[1]{{\color{clnote}\footnotesize\textsf{[Claude: #1]}}} 
\fi

\newcommand{\parasum}[1]{\textbf{\textit{#1}}}
\newcommand{\circled}[1]{\textcircled{\raisebox{-0.9pt}{#1}}}

\AtBeginDocument{\captionsetup[figure]{skip=4pt}\captionsetup[table]{skip=4pt}}

\graphicspath{{./}}

\begin{document}

\newcommand{\sys}{\textsc{CATF}}
\newcommand{\syss}{\textsc{CATF}\xspace}

\title{\sys: A Manager-Worker Framework for Context-Aware Time-Series Forecasting}


\author{Yueqiao Chen}
\affiliation{
  \institution{UC Davis}
  \city{Davis}
  \state{CA}
  \country{USA}
}
\email{yeqchen@ucdavis.edu}


\author{Dongyu Liu}
\affiliation{
  \institution{UC Davis}
  \city{Davis}
  \state{CA}
  \country{USA}
}
\email{dyuliu@ucdavis.edu}

\renewcommand{\shortauthors}{Chen et al.}

\begin{abstract}

We study context-aware time-series forecasting, in which a single target variable is predicted from its own history and a set of contextual variables, such as weather and calendar information. As the context becomes more diverse, a single model struggles to capture all of its regimes, while ensemble learning and Mixture-of-Experts approaches rely on manual data partitioning or soft routing, which increases the inference cost and hinders interpretability. We propose \sys, a \underline{C}ontext-\underline{A}ware \underline{T}ime-Series \underline{F}orecasting framework built on a Manager-Worker architecture: a Transformer-based Manager routes each input to a single Worker based on the context, using hard routing during both training and inference. The Workers thus specialize without manual data partitioning, and any single-model forecaster can serve as a Worker. On five context-rich datasets, four horizons, and two normalization schemes, \sys~improves over its standalone counterpart in 162 of 200 comparisons, the best \sys-enhanced model outperforms the best standalone baseline in 19 of 20 settings, and it outperforms the zero-shot Time-MoE foundation model in 15 of 20 settings. A case study on bike-sharing data shows that the Manager routes inputs by the calendar and that each Worker specializes in a different weekly pattern.

\end{abstract}

 


\keywords{Time-series Forecasting, Context-aware Learning, Exogenous Variables, Training Algorithm, Manager-Worker Architecture, Deep Learning}


\maketitle

\section{Introduction}
Time-series forecasting is critical to applications such as weather prediction, electricity load management, traffic analysis, financial forecasting, and industrial monitoring~\cite{weather, electricity, traffic, stock, Alnegheimish2022SintelAM}. However, growing temporal complexity and rich contextual dependencies make accurate forecasting increasingly difficult and computationally expensive. Contextual variables are often heterogeneous and dynamic, evolving abruptly or nonlinearly over time. For example, hourly bike-sharing demand depends on weather, temperature, and the calendar: on working days it peaks during the morning and evening commutes, whereas on weekends and holidays it follows a single midday peak, and a holiday on a weekday can switch the demand to the weekend pattern.
In this work, we focus on such context-aware forecasting: the goal is to forecast a single target variable from its history and the contextual variables that drive it, a setting also known as forecasting with exogenous variables. This differs from the setting of most multivariate forecasting benchmarks~\cite{thuml_tslib}, in which all variables are forecast jointly; here, the contextual variables serve only as inputs.

Recent advances in deep learning have led to impressive results in time-series forecasting. Convolutional networks~\cite{TimesNet}, graph neural networks~\cite{Huang2023CrossGNNCN}, and Transformer-based models~\cite{Liu2024iTransformerIT} models have been adapted to capture complex temporal dependencies.
More recently, large-scale time-series foundation models (TSFMs)~\cite{jin2023time,Qiao2025MultiScaleFF} have shown strong performance in zero-shot and few-shot settings. However, these models are expensive to fine-tune, struggle with distribution shifts, and often fail to generalize to domain-specific dynamics.
To improve adaptability, researchers have explored alternatives such as retrieval-augmented forecasting~\cite{liu2024retrieval}, meta learning~\cite{oreshkin2021meta}, and adaptive policies~\cite{rl_in_time_series}. 

Hybrid forecasting approaches offer another promising direction by allowing model specialization under diverse conditions. 
Ensemble methods~\cite{galicia2019multi,He2023FinancialTS} train specialized models on partitioned subsets and aggregate their predictions. However, they incur inference overhead and require manual segmentation and weighting strategies. Mixture-of-Experts (MoE) models~\cite{early_moe,Shi2025TimeMoEBT} improve efficiency through sparse activation but demand careful gating design and are prone to expert under-utilization without \mbox{strong regularization}.

To address these challenges, we propose \textbf{\sys}, a \underline{C}ontext-\underline{A}ware \underline{T}ime-Series \underline{F}orecasting framework based on a Manager-Worker (MW) architecture.
\syss consists of a high-level transformer-based \emph{Manager} and a set of \emph{Worker} models trained via joint optimization. The Manager dynamically routes each input to a single specialized Worker based on contextual features, enabling each Worker to focus on a distinct subset of conditions. 
Inspired by economic specialization~\cite{Hidalgo2021EconomicCT}, we formulate training as a competitive yet cooperative process: the Manager learns to assign inputs to the most suitable Worker, while Workers specialize in minimizing error within their assigned contexts.

Unlike conventional MoEs, \syss employs \emph{hard routing}, assigning each input to a single Worker during both training and inference. This improves interpretability, encourages Worker specialization, and eliminates the need for manual partitioning or complex gating mechanisms. In addition, \syss supports heterogeneous architectures, enabling seamless integration of state-of-the-art forecasting models with minimal tuning.
We evaluated \syss on five real-world context-rich datasets across four forecasting horizons, observed improvements over strong single-model baselines, and analyze its routing decisions. Our key contributions are as follows.
\begin{itemize}[leftmargin=10pt]
    \item \textbf{Modular framework.} We propose \sys, a Manager-Worker framework in which a Transformer-based Manager routes each input to one of several Worker models through hard routing, trained with a joint mechanism that lets the Workers specialize without manual data partitioning (Section~\ref{sec:model}). \sys~is model-agnostic: any single-model forecaster can serve as a Worker without modification, in homogeneous or heterogeneous configurations, and we instantiate it with five representative architectures, Autoformer~\cite{autoformer}, Informer~\cite{informer}, FEDformer~\cite{Fedformer}, TimesNet~\cite{TimesNet}, and iTransformer~\cite{Liu2024iTransformerIT}.
    \item \textbf{Context-aware interpretation.} Because each prediction is produced by a single, identifiable Worker, the routing decisions can be inspected and related to the context. In a case study on a real-world bike-sharing dataset, we show that the Manager learns to route inputs by the most relevant contextual variables (e.g., working days vs.\ weekends) and that the Workers specialize accordingly (Section~\ref{sec:context}); on synthetic data with known context-driven regimes, the same analysis recovers the ground truth (Appendix~\ref{app:context_syn}).
    \item \textbf{Performance improvement.} On five context-rich real-world datasets, four horizons, and two normalization schemes, \sys~improves over its standalone counterpart in 162 of the 200 comparisons (81\%), the best \sys-enhanced model outperforms the best standalone baseline in 19 of the 20 settings, \sys~reduces the run-to-run variance of unstable baselines, and it outperforms the zero-shot Time-MoE foundation model~\cite{Shi2025TimeMoEBT} in 15 of the 20 settings (Section~\ref{sec:comp1}).
\end{itemize}
\section{Related Work}

\subsection{Time Series Forecasting}
\textbf{Single-Model Approaches.}
Early forecasting models relied on convolutional (CNN)~\cite{Bai2018AnEE,Dempster2019ROCKETEF,Liu2022SCINetTS,TimesNet} and recurrent (RNN)~\cite{zhao2017lstm,rangapuram2018deep,salinas2020deepar} architectures to capture temporal dependencies. CNNs are efficient and effective for local patterns, while RNNs model autoregressive behavior. However, both struggle with long-range dependencies. Graph neural networks (GNNs)~\cite{Wu2020ConnectingTD,Huang2023CrossGNNCN} address inter-variable relationships but incur high computational cost in high-dimensional or long-sequence settings. More recently, multilayer perceptron (MLP) models~\cite{wang2024timemixer,Li2023RevisitingLT} have emerged as efficient alternatives, relying on token mixing or linear projections. Transformer-based models~\cite{informer, autoformer, Fedformer, Nie2022ATS, Liu2024iTransformerIT} have become more popular due to their ability to model global dependencies through self-attention. 

Despite their strengths, single-model architectures often struggle to generalize across heterogeneous and dynamic temporal regimes. Models tuned for specific conditions may underperform under unexpected shifts such as holidays or extreme events. Our proposed framework addresses this limitation by enabling dynamic model selection based on contextual cues.

\noindent
\textbf{Time-Series Foundation Models (TSFMs).}
These approaches aim to build universal forecasting models pretrained on large-scale datasets~\cite{liang2024foundation,jin2023time}. These models vary in architecture (e.g., Transformer, diffusion, or MLP), pretraining objectives (supervised or self-supervised), and downstream adaptation strategies (fine-tuning, prompting, zero-shot)~\cite{goswami2024moment,ansari2024chronos,liu2024timer,das2024decoder,das2024context,gao2024units,benechehab2025adapts}. While TSFMs offer general-purpose representations and cross-domain transferability, they are computationally expensive and often brittle under distribution shifts, particularly in domains with limited or \mbox{highly specialized data}.

\subsection{Hybrid Forecasting Architectures}
Hybrid frameworks combine multiple models to leverage their complementary strengths. 

\noindent
\textbf{Ensemble learning.}
Ensemble learning methods—including bagging, boosting, stacking, and hybrid approaches—consistently improve prediction accuracy in time series forecasting by reducing variance, mitigating bias, and modeling complex non-linear dependencies~\cite{Kumar2019EnsembleL,context-aware_ensemble,ensemble_for_time_series,finance_ensemble_learning,He2023FinancialTS,galicia2019multi}.
These methods combine predictions from multiple independent models, typically through averaging or weighting, which enhances robustness but often overlooks input-specific context. 
Their effectiveness can be constrained by the need for manual design of segmentation and aggregation strategies, and performance gains tend to diminish when base models are highly correlated. 
Moreover, ensemble methods are computationally more demanding and less interpretable compared to individual models. Achieving superior performance over the best single model requires careful tuning of integration strategies, such as model weighting and diversity enforcement.

\noindent
\textbf{Mixture-of-Experts (MoE).} 
MoE architectures~\cite{jacobs1991adaptive,early_moe,riquelme2021scaling,gshard,chen2024pathformer,mole,Shi2025TimeMoEBT} dynamically route inputs to specialized expert models via a gating mechanism. While sparse activation improves computational efficiency, these models require careful gating design and are prone to issues such as expert collapse or under-utilization without strong regularization. In time-series forecasting, recent MoE variants like Time-MOE~\cite{Shi2025TimeMoEBT} have shown promising results. Time-MOE is a decoder-only autoregressive Transformer trained on over 300 billion time points. It employs point-wise tokenization, multi-resolution attention heads, and top-$k$ expert activation with load-balancing to achieve scalability with minimal inference overhead. However, such models remain challenging to optimize and are often tightly coupled to specific architectural choices, limiting their flexibility across forecasting domains.

Our framework builds on the core principle of specialization, but replaces soft gating with hard routing implemented through a learned Transformer-based Manager. This design eliminates the instability commonly associated with gating mechanisms and enables transparent, deterministic input-to-expert assignments. Unlike prior MoE approaches, our framework naturally supports heterogeneous Workers and allows seamless integration of SOTA forecasting models. As a result, it achieves both scalability and interpretability—while also mitigating the challenges faced by ensemble methods, such as manual data partitioning and extensive tuning.

\section{Preliminaries}

\subsection{Data Description}
\label{sec:datadesc}

This study focuses on \textbf{context-aware multivariate time-series forecasting}: the model observes the history of a target variable together with multiple contextual variables, and predicts the future values of the target variable.

\textbf{\textit{Target Time-Series Data.}}
Let $x_t \in \mathbb{R}$ denote the value of the target variable at time $t$ (e.g., the PM2.5 concentration or the number of rented bikes). Its past observations up to time $t$ are
\[
X_{t, \Delta t, L} = (x_{t - (L-1)\Delta t}, x_{t - (L-2)\Delta t}, \ldots, x_t),
\]
where $\Delta t$ is the time interval between consecutive observations and $L$ is the length of the historical (lookback) window.

\textbf{\textit{Multivariate Contextual Data.}}
In addition to the target, we observe $N$ contextual variables that capture external factors influencing the target dynamics, such as weather conditions, co-pollutants, or calendar indicators. Let $\mathbf{c}_t \in \mathbb{R}^{N}$ denote the contextual observations at time $t$; the contextual history over the same window is
\[
C_{t, \Delta t, L} = (\mathbf{c}_{t - (L-1)\Delta t}, \mathbf{c}_{t - (L-2)\Delta t}, \ldots, \mathbf{c}_t).
\]
The target and contextual variables are temporally aligned, so the model input at each time step is the $(N{+}1)$-dimensional vector $(x_t, \mathbf{c}_t)$, and the full input is an $L \times (N{+}1)$ matrix.

\subsection{Problem Formulation}
Given the historical target sequence $X = X_{t,\Delta t,L}$ and the associated contextual sequence $C = C_{t,\Delta t,L}$, the objective is to forecast the future values of the target variable over a horizon of length $L_y$:
\[
Y = (x_{t+\Delta t}, x_{t+2\Delta t}, \ldots, x_{t+L_y\Delta t}) \in \mathbb{R}^{L_y}.
\]
A forecasting model $f$ learns the mapping
\[
\hat{Y} = f(X, C), \qquad \hat{Y} = (\hat{x}_{t+\Delta t}, \ldots, \hat{x}_{t+L_y\Delta t}),
\]
that is, a \emph{multivariate-input, univariate-output} mapping. The contextual variables are only used as inputs; their future values are neither predicted nor assumed to be known.

\textbf{\textit{Normalization and Evaluation.}}
Because the target and contextual variables have very different scales, the inputs are normalized by a transform $g$ before being fed to the model, and the model is trained to minimize the MSE between its output and $g(Y)$ in the normalized space. We consider two choices of $g$: \emph{global} normalization, which standardizes every variable with the mean and standard deviation of the training split, and \emph{sample} normalization, which standardizes each input window with its own per-variable statistics and scales the target horizon with the statistics of the target in the input window. At test time, predictions are mapped back to the original scale with the inverse transform, and the error is computed against the raw ground truth:
\begin{equation}
\label{eq:eval_mse}
\text{MSE} = \frac{1}{|\mathcal{T}|\,L_y} \sum_{t \in \mathcal{T}} \sum_{i=1}^{L_y} \left( g^{-1}(\hat{Y})_{i} - x_{t + i\Delta t} \right)^2,
\end{equation}
where $\mathcal{T}$ is the set of forecast origins in the test split. Reporting the error in the original units of the target makes the results directly interpretable (e.g., squared rentals per hour for Bike Share) and comparable across normalization schemes, but not across datasets.

While recent forecasting models have achieved strong performance by modeling complex temporal dependencies, their effectiveness often degrades in the presence of heterogeneous and context-dependent patterns. Different temporal regimes may exhibit distinct dynamics, making it difficult for a single model to capture all forecasting behaviors effectively. This motivates model specialization, where different forecasting models focus on different subsets of the data distribution.

To address this challenge, the proposed CATF framework adopts a Manager--Worker architecture. A manager model first extracts contextual characteristics from the target history $X$ and the context $C$ and dynamically assigns each instance to one of several specialized worker models. By routing each instance to the most suitable worker, CATF enables adaptive specialization without requiring manual data partitioning, thereby improving forecasting accuracy across diverse temporal patterns, while the routing decisions reveal which patterns each worker specializes in.

\section{CATF Framework}
\label{sec:model}

This section presents the technical details of \syss framework, a context-aware time-series forecasting approach based on a Manager–Worker (MW) architecture. We describe its model structure and training algorithm.

\subsection{Modeling}
\label{sec:arch}

\begin{figure}[t]
    \centering
    \includegraphics[width=\linewidth]{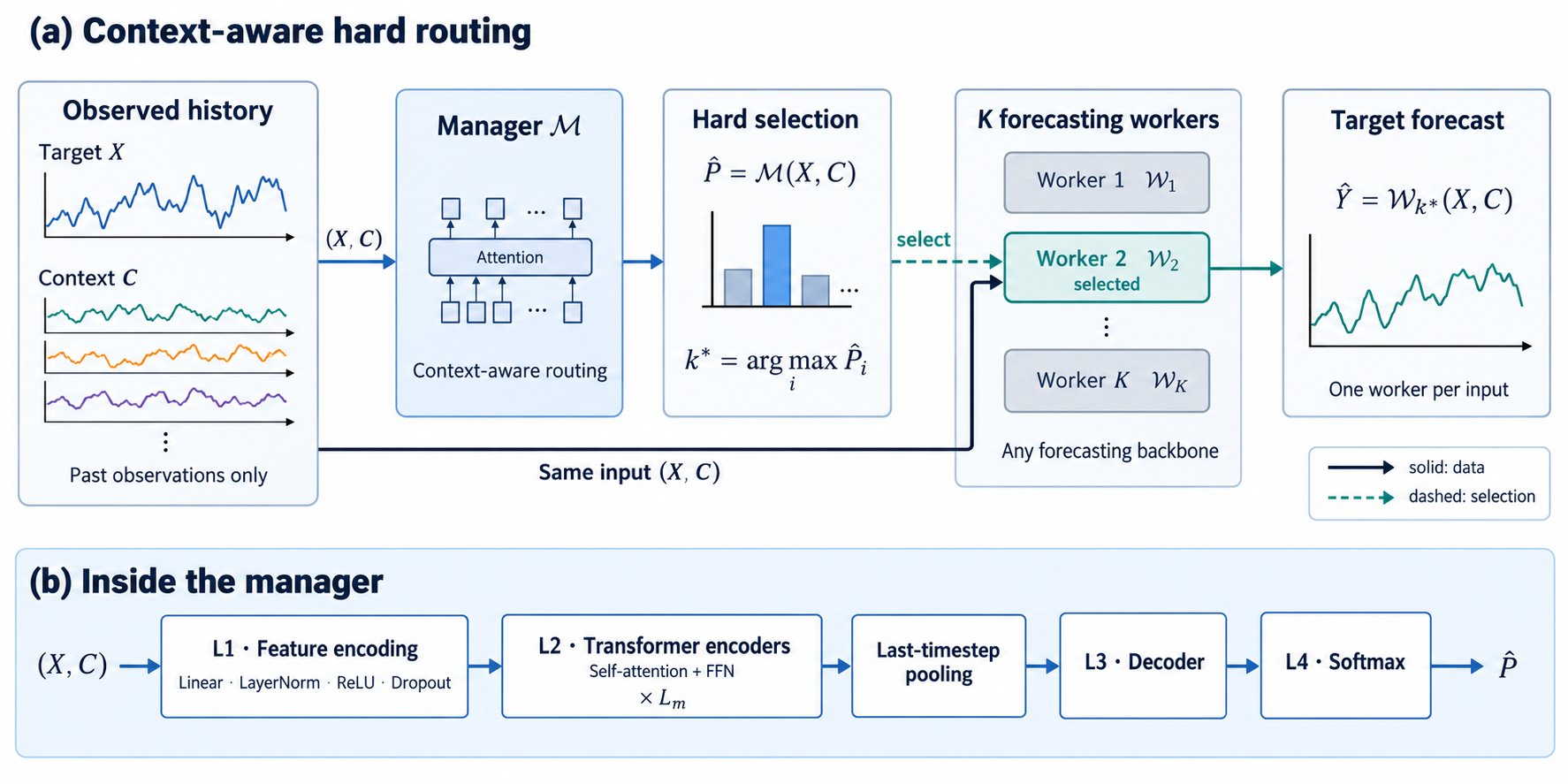}
    \caption{Manager--Worker architecture of \sys. (a) Given the target history $X$ and the context history $C$, the manager $\mathcal{M}$ outputs routing probabilities $\hat{P}$ over the $K$ workers. The worker with the highest probability, $\mathcal{W}_{k^*}$, receives the same input $(X, C)$ and alone predicts the future target sequence $\hat{Y}$. (b) Inside the manager: feature encoding (L1), $L_m$ Transformer encoder layers (L2), last-timestep pooling, a decoder (L3), and softmax (L4).}
    \Description{Two-panel architecture diagram. The upper panel shows the target history and the context history entering a manager that selects one worker by the largest routing probability. The same historical input is passed directly to that worker, which alone produces the target forecast. The lower panel expands the manager into feature encoding, Transformer encoders, last-timestep pooling, decoder, and softmax.}
    \label{fig:arch}
\end{figure}

\syss consists of a manager model $\mathcal{M}$ and $K$ worker models $\mathcal{W}_1, \dots, \mathcal{W}_K$, as illustrated in Figure~\ref{fig:arch}a. 
The manager $\mathcal{M}$ takes the target history $X$ and the context history $C$ as input and estimates, for each worker $\mathcal{W}_i$, the probability $\hat{P}_i$ that it is best suited to predict the future target sequence. 
The worker with the highest probability, $\mathcal{W}_{k^*}$ with $k^* = \arg\max_i \hat{P}_i$, is selected and generates the forecast $\hat{Y} = \mathcal{W}_{k^*}(X, C)$ from the same input; at inference time, the other workers are not executed.

\subsubsection{Manager Model}
The manager performs a $K$-class classification task, mapping input $(X, C)$ to a probability distribution over the $K$ worker models. As shown in Figure~\ref{fig:arch}b, the manager follows a Transformer-based architecture~\cite{Transformer}, consisting of:
\begin{itemize}[leftmargin=10pt]
    \item \textbf{Feature Encoding (L1):} A linear projection followed by layer normalization, ReLU activation, and dropout transforms raw input into high-dimensional embeddings.  
    This preprocessing ensures stable training and expressive representations, which are essential for downstream temporal modeling.

    \item \textbf{Transformer Encoder Stack (L2):} A stack of $L_m$ Transformer encoder layers models long-range and complex temporal-contextual dependencies through self-attention mechanisms. Each layer consists of multi-head self-attention and position-wise feedforward sub-layers, equipped with GELU activation~\cite{hendrycks2016gaussian} and residual connections.  
    GELU is chosen for its smooth, non-linear behavior, which preserves gradient flow and has been empirically shown to enhance optimization stability and performance in deep Transformer-based networks.

    \item \textbf{Output Layer (L3–L4):} Following Transformer-based classification designs~\cite{chen2021crossvit,zhang2021token}, last-timestep pooling summarizes the sequence into a fixed-length representation. This is passed through a decoder (L3) and softmax (L4) to produce the output distribution over workers.  
    Last-timestep pooling preserves the most recent temporal information, which is critical for decision-making in time-sensitive forecasting scenarios.
\end{itemize}

\subsubsection{Worker Model}
\label{sec:worker}
The selected worker predicts the future target sequence $\hat{Y}$ given the input sequence $X$ and covariates $C$, as illustrated in Figure~\ref{fig:arch}. \syss allows each worker to be instantiated with any time-series forecasting model, supporting both homogeneous and heterogeneous ensembles. In most experiments, we use multiple instances of the same base model to assess the effectiveness of the framework, while heterogeneous mixtures are examined in ablation studies.

We incorporate several established models as workers, including Informer~\cite{informer}, Autoformer~\cite{autoformer}, FedFormer~\cite{Fedformer}, TimesNet~\cite{TimesNet}, and i-Transformer~\cite{Liu2024iTransformerIT}. These models also serve as baselines for evaluation. Their base implementations are consistent with the official versions provided in the Time-Series-Library~\cite{thuml_tslib}, and we use the same hyper-parameter settings to ensure fair comparisons. For each base model, we implement a corresponding \syss variant within our framework. Users can specify the base models to use—either homogeneous or mixed—along with the number of workers of each type, through a simple configuration file.



\subsection{Training Mechanism}
\label{sec:train}

We introduce an iterative training algorithm to jointly train the manager and workers, shown as Figure~\ref{fig:training} and Algorithm \ref{alg:train}. The \textbf{manager} learns to assign each input to the most suitable worker, and the \textbf{workers} are trained based on the data routed from the manager, specializing in certain \mbox{context conditions.} 

\subsubsection{Main Training Process}
\begin{figure}[t]
    \centering
    \includegraphics[width=\linewidth]{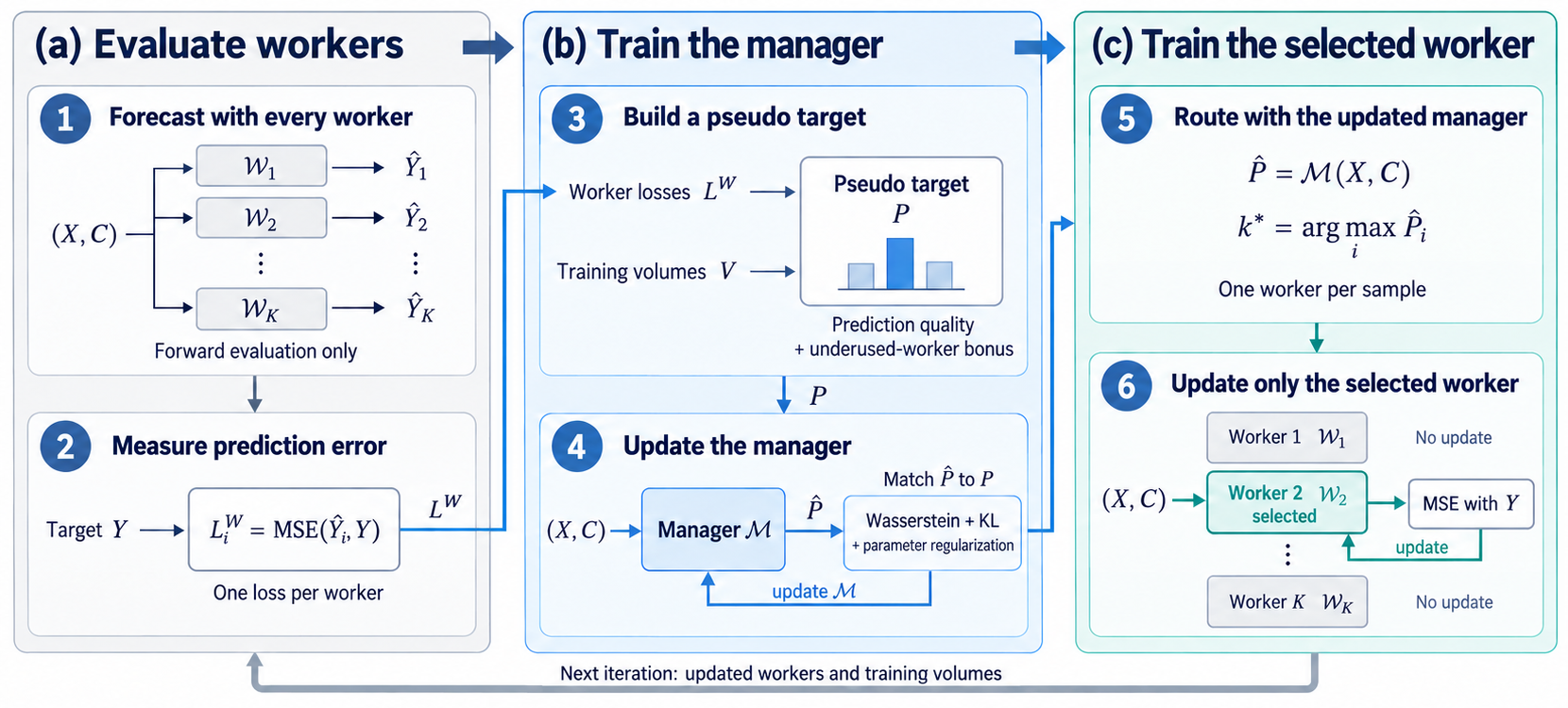}
    \caption{Iterative training of \sys, illustrated for one sample. (a) All workers first produce forecasts and losses (steps 1--2). (b) Their losses and training volumes define a pseudo target (step 3) that supervises the manager update (step 4). (c) The updated manager then selects one worker (step 5), and only that worker is trained on the sample (step 6).}
    \Description{Three-phase training diagram with six chronological steps. All workers first predict and are scored against the target. Worker losses and training volumes define a pseudo target that supervises the manager. The updated manager then selects one worker for the sample, and only that worker receives a forecasting update. A return arrow repeats the process with updated workers and training volumes.}
    \label{fig:training}
\end{figure}

Each training iteration consists of three phases (Figure~\ref{fig:training}). (a) Every worker $\mathcal{W}_i$ forecasts the target, $\hat{Y}_i = \mathcal{W}_i(X, C)$ (step~\circled{1}; line 4 in Algorithm~\ref{alg:train}), and we compute its loss $L^W_i$ (step~\circled{2}). This forward pass only evaluates the workers and does not update them. (b) The worker losses, together with the training volumes of the workers, define a pseudo ground truth $P$ for the manager output (step~\circled{3}; line 5), which supervises the update of the manager $\mathcal{M}$ (step~\circled{4}; line 6). (c) The updated manager predicts the routing probabilities $\hat{P} = \mathcal{M}(X, C)$ and selects the worker $\mathcal{W}_{k^*}$ with $k^* = \arg\max_i \hat{P}_i$ (step~\circled{5}; lines 7--8). Only this worker is trained on the sample (step~\circled{6}; lines 9--13), while the other workers are not updated.

This selective training strategy ensures that each worker is exposed to and learns from samples within specific contextual regimes. Over time, this encourages the emergence of functional specialization—each worker becomes proficient in forecasting under particular conditions. Section~\ref{sec:context} analyzes this specialization empirically. This approach is inspired by the principle of economic specialization~\cite{Hidalgo2021EconomicCT}, where agents maximize collective productivity by focusing on tasks they are best suited to. Similarly, \syss enables each worker to ``do what it does best,'' thereby improving overall forecasting accuracy and robustness in \mbox{heterogeneous environments.}

\subsubsection{Training Details}
The following details describe key components and design choices of the training process described above.

\parasum{Detail \#1: Training Order.}
We train the manager model first and then use it to select the appropriate worker for training. In the initial iteration, untrained workers make predictions on a set of samples, and their losses are used to generate a pseudo ground truth distribution for the manager. This allows the manager to learn how to assign tasks based on relative worker performance from the outset. Once trained, the manager routes future samples to the most suitable worker.

This process can be thought of as talent scouting before assigning roles: the manager first observes each worker's raw potential under various contexts—without prior bias—and learns who is naturally inclined to handle which kind of data. In contrast, if we train the workers first, the untrained manager must randomly assign tasks, akin to blindly assigning roles before knowing who's good at what. The randomly selected worker would then receive more training and, as a result, dominate in performance—not because of superior aptitude, but because of an early, unearned advantage. This skews the pseudo ground truth, leading the manager to reinforce this imbalance and reducing overall diversity and specialization.
Alternatively, training all workers on every sample from the beginning contradicts the core design of specialization, where workers are meant to focus on distinct contextual regimes.

By training the manager first—using the raw, untrained losses as an initial signal—we provide it with a fair and unbiased starting point for assessing each worker's relative strengths. While this is only the beginning of an iterative co-training process, it lays the foundation for a more principled and effective division of labor. This approach encourages the emergence of specialized expertise across workers, without introducing early random bias that could skew the learning dynamics.





\begin{algorithm}[!t]
\caption{Iterative Training Process of CATF}
\label{alg:train}
\begin{algorithmic}[1]
\REQUIRE Training data $\mathcal{D} = \{(X_i, C_i, Y_i)\}$, Manager $\mathcal{M}$, Workers $\{\mathcal{W}_1, \dots, \mathcal{W}_K\}$, epochs $E$, worker update steps $\alpha$
\STATE Initialize $\mathcal{M}$ and $\{\mathcal{W}_i\}_{i=1}^K$
\FOR{epoch $= 1$ to $E$}
    \FOR{each mini-batch $(X, C, Y) \in \mathcal{D}$}
        \STATE Compute all worker losses $L^W_i = \mathcal{L}(\mathcal{W}_i(X, C), Y)$
        \STATE Compute target weights $P$
        \STATE Train manager $\mathcal{M}$
        \STATE $\hat{P} = \mathcal{M}(X, C)$ \hfill \textcolor{gray}{\# predict routing probabilities}
        \STATE Select worker index $k^* = \arg\max \hat{P}$ for each sample
        \FOR{step $= 1$ to $\alpha$}
            \FOR{$i = 1$ to $K$}
                \STATE Select sub-batch $\{(x_j, y_j)\}$ where $k^*_j = i$
                \IF{selection not empty}
                    \STATE Train Worker $\mathcal{W}_i$ on selected sub-batch
                \ENDIF
            \ENDFOR
        \ENDFOR
    \ENDFOR
    \STATE Validate and save best checkpoint if improved
\ENDFOR
\end{algorithmic}
\end{algorithm}
        

\parasum{Detail \#2: Pseudo Ground Truth Calculation.}
A core challenge in training the manager is the absence of a ground-truth probability distribution $P$ that indicates which worker should handle each sample. To address this, we adopt the intuitive principle that a worker with lower prediction loss should be more likely to be selected. 
Concretely, each worker first makes a prediction for the given input (step~\circled{1} in Figure~\ref{fig:training}), and we compute its prediction loss $L^W_i$ (e.g., $MSE_i$) (step~\circled{2}). These losses are then used to construct a soft target distribution over workers. The simplest form of this pseudo ground truth is defined as:
\begin{equation}
    \label{eq:probOfLoss}
    \begin{aligned}
    MaxL &= \max_{1 \leq i \leq K}L^W_i,\\
    P &= Softmax\left(\frac{MaxL}{L^W}\right),
    \end{aligned}
\end{equation}
where lower losses result in higher scores. This distribution $P$ serves as a stand-in for the missing ground truth, guiding the training of the manager to produce probability outputs that align with worker performance (step~\circled{4} in Figure~\ref{fig:training}).


\parasum{Detail \#3: Training Regularization.}
A critical challenge in training \syss is maintaining a balanced learning process across all workers. Without proper regulation, a competitive imbalance may emerge: if one worker $W_i$ happens to perform better than others—even if not inherently more capable—the manager will increasingly assign it more samples in subsequent iterations. This feedback loop can result in $W_i$ becoming overly dominant, while other workers are starved of data and fail to specialize. Ultimately, this degrades model diversity and undermines the utility of contextual information.
This phenomenon resembles the classic exploration–exploitation dilemma in Monte-Carlo Tree Search (MCTS)~\cite{browne2012survey}, where the challenge lies in choosing between high-performing nodes and less-explored ones. Inspired by MCTS, we introduce a regularization term (step~\circled{3} in Figure~\ref{fig:training}, line 5 in Algorithm~\ref{alg:train}, Equation~\ref{eq:reg}) that considers the training volume $V_i$—the number of data samples allocated to each worker $W_i$ (line 13):
\begin{equation}
    \label{eq:reg}
    \begin{aligned}
        MaxV &= \max_{1 \leq i \leq K}V_i,\\
        P &= Softmax\left(e^{-\frac{L^W}{MaxL}} + \beta \cdot \frac{MaxV - V}{MaxV}\right),
    \end{aligned}
\end{equation}
where $e^{-L^W/MaxL}$ represents the new loss-based term (\texttt{TNewLoss}), and $(MaxV - V)/MaxV$ is the regularization term (\texttt{TReg}). This formulation differs from the original loss-based probability in Equation~\ref{eq:probOfLoss}, where $MaxL/L^W$ (\texttt{TOriginLoss}) may produce a much wider range and overshadow the contribution of the regularization term.

We design \texttt{TReg} to be large during early training—encouraging the manager to explore underutilized workers—and to diminish over time as training volumes equalize. This dynamic is supported by the fact that $MaxV$ generally grows faster than $MaxV - V$, naturally reducing \texttt{TReg}'s effect. The hyperparameter $\beta$ controls the strength of the regularization.

\texttt{TNewLoss} is preferred over \texttt{TOriginLoss} because its output range ($1/e$ to $1$) aligns better with \texttt{TReg} (which ranges from $0$ to $1$). This ensures that both terms meaningfully influence the resulting probability distribution $\hat{P}$ and that the manager balances performance with exploration throughout training.

\parasum{Detail \#4: Batch Training.}
For clarity, Figure~\ref{fig:training} depicts the training process on a single sample. However, our implementation is designed to support efficient batch-wise training. Specifically, in each training iteration (Algorithm~\ref{alg:train}, lines 3–13), we first evaluate all workers and update the manager on the current mini-batch (lines 4--6), and then compute the updated manager's output probabilities for all samples (line 7). Each sample is then assigned to the worker with the highest predicted probability via $\arg\max$ selection (line 8). Next, we partition the batch into $K$ sub-batches, one for each worker, based on the manager's assignments. Each worker $W_i$ is then trained only on its corresponding sub-batch—i.e., the samples for which it was selected—allowing it to focus on the specific types of inputs it is most suited for.

To improve training stability and accelerate convergence, we repeat the worker update step $\alpha$ times for each batch (lines 9–13 in Algorithm~\ref{alg:train}). 


\parasum{Detail \#5: Loss Function Selection.}
\label{sec:loss_selection}
For the \textbf{workers}, we adopt a standard loss metric for time-series prediction: the Mean Squared Error (MSE). This loss quantifies the average squared difference between the predicted target sequence $\hat{Y}$ and the ground truth $Y$, and is defined as:
\begin{align}
    \text{MSE}(Y, \hat{Y}) &= \frac{1}{L_Y} \sum_{i=1}^{L_Y} \left\| \hat{y}_{t + i\Delta t} - y_{t + i\Delta t} \right\|_2^2 \label{eq:mse}
\end{align}
We denote the MSE loss of worker $W_i$ as $L^W_i$, and use it as the primary feedback signal during worker training.

For the \textbf{manager}, we use a hybrid objective that combines the Wasserstein distance~\cite{arjovsky2017wasserstein} and the Kullback–Leibler (KL) divergence to measure the discrepancy between the manager's predicted probability distribution $\hat{P}$ and the pseudo ground truth $P$.

The Wasserstein distance is more continuous and geometry-aware than KL or Jensen–Shannon divergence~\cite{azari2025divergence}, offering smoother gradients and greater stability during optimization. It is defined as:
\begin{equation}
    \label{eq:wasDef}
    \mathbb{W}(\hat{P}, P) = \sup_{\parallel f\parallel_L \leq 1}\mathbb{E}_{x\sim \hat{P}}[f(x)] - \mathbb{E}_{x\sim P}[f(x)],
\end{equation}
where $f(x)$ is a 1-Lipschitz function satisfying $|f(x_1) - f(x_2)| \leq |x_1 - x_2|$. This constraint ensures smooth alignment between probability distributions.

While Wasserstein distance enables fine-grained distribution matching, using it alone often leads to over-concentration of samples on a single dominant worker, which undermines diversity. To address this, we incorporate the KL divergence:
\begin{equation}
\mathcal{L}_{\text{KL}}(P \| \hat{P}) = \sum_{i=1}^{K} P(i) \log\left( \frac{P(i)}{\hat{P}(i)} \right) \label{eq:kl}
\end{equation}
KL divergence penalizes mismatches where high-probability targets receive low attention, helping the manager distribute samples more evenly—particularly important during the early stages of training when $P$ is relatively uniform.

By combining both terms, we enable the manager to balance diversity and precision throughout training. The KL term encourages broad worker participation, while the Wasserstein term refines the accuracy of the assignment.
The resulting hybrid loss for the manager is as Equation~\ref{eq:manager_loss}:
\begin{align}
\mathcal{L}_{\text{manager}} = \ 
& \lambda_{\text{wass}} \cdot \text{Wasserstein}(\hat{P}, P) +
  \lambda_{\text{KL}} \cdot \text{KL}(P \| \hat{P}) \notag \\
& + \lambda_{\text{reg}} \cdot \sum\nolimits_{j} \| \theta_j \|_2
\label{eq:manager_loss}
\end{align}
, where $\theta_j$ denotes the $j$-th parameter tensor (the weight matrix or bias vector of a layer) of the manager model.
The KL divergence is much larger in magnitude than the Wasserstein distance. We therefore set $\lambda_{\text{wass}} = 10$ and $\lambda_{\text{KL}} = 0.01$, which bring the two terms to a comparable scale so that neither dominates the manager update, and $\lambda_{\text{reg}} = 1 \times 10^{-8}$. The last term is a group-lasso penalty~\cite{yuan2006model} on the manager's parameters, where $\theta_j$ denotes the $j$-th parameter tensor and $\lambda_{reg}=10^{-8}$. In our ablation, removing this term has no significant effect (Section~\ref{sec:ablation}).

\section{Experiments}


We implemented \syss and evaluated its performance through both comparative experiments and ablation studies. All models were trained using PyTorch~\cite{pytorch} with CUDA acceleration on a server equipped with an Nvidia RTX 6000 Ada GPU (48 GB). To ensure fair comparison, each experiment was repeated across \textbf{5 independent runs} per model, dataset, forecasting horizon, and normalization mode. The code is available at \url{https://github.com/via-cs/catf}.

\subsection{Setup and Baselines}

We compare \syss with five widely used multivariate time-series (MTS) forecasting models and a pretrained time-series foundation model on five real-world datasets, each evaluated at four forecasting horizons and under two normalization schemes. The experimental setup and baseline details are outlined below.


\subsubsection{Datasets}

We evaluate on five datasets in which a single target variable is driven by contextual variables of a different nature, such as weather and calendar variables for bike demand. We do not use standard multivariate benchmarks such as ETT or Electricity: they are designed for joint forecasting of variables of the same kind (e.g., the loads of different clients) and contain no such contextual variables. For each dataset, we chronologically partition the samples into 80\% for training, 10\% for validation, and 10\% for testing. The input length (lookback window) is fixed to $L=96$, and the output length is $L_y \in \{96, 192, 336, 720\}$, following the long-term forecasting protocol of prior work~\cite{Liu2024iTransformerIT, TimesNet, autoformer}. Dataset statistics are listed in Appendix Table~\ref{app:dataset}.


\begin{itemize}[leftmargin=10pt,nosep]
    \item \textbf{Weather~\cite{weather_dataset}} (21 variables): Meteorological measurements recorded every 10 minutes during 2020 at a weather station in Germany, including pressure, temperature, humidity, wind, and radiation. Following our code, the last column, CO$_2$ concentration (ppm), is used as the target and the other 20 variables as context. As in the original data, the target contains 50 missing-value codes ($-9999$) in the training period, which we keep unchanged.
    \item \textbf{PM2.5-LA and PM2.5-RV~\cite{epa_aqs,era5}} (15 variables each): Hourly PM2.5 concentrations from 2022 to 2025 at two air-quality monitoring sites in Southern California (Los Angeles and Riverside). The context consists of on-site meteorology (wind speed, temperature, relative humidity, solar and UV radiation, barometric pressure), ERA5 reanalysis variables (surface pressure, skin temperature, boundary-layer height, total precipitation), and co-pollutants (CO, SO$_2$, NO$_2$, O$_3$).
    \item \textbf{Bike Share~\cite{bike_sharing}} (12 variables): Hourly counts of rented bikes in the Capital Bikeshare system (Washington, D.C.) in 2011--2012. The context consists of calendar indicators (season, year, month, holiday, weekday, working day) and weather conditions (weather situation, temperature, feels-like temperature, humidity, wind speed).
    \item \textbf{Appliances~\cite{appliances_energy}} (26 variables): Energy use of appliances in a low-energy house, recorded every 10 minutes from January to May 2016. The context consists of lighting energy, temperature and humidity in nine rooms, and outdoor weather from a nearby station.
\end{itemize}


\subsubsection{Baselines}
We evaluate \syss against a diverse set of baseline models, including both widely adopted MTS forecasting models and a recent time-series foundation model.
This allows us to assess \sys's effectiveness across a broad range of model designs and forecasting settings.

\parasum{Worker Models and Their \sys-Enhanced Variants.}
\sys~is a flexible framework that allows users to freely select the architecture of each worker model. In our experiments, we incorporate five established forecasting models as workers: four Transformer-based models—Informer~\cite{informer}, Autoformer~\cite{autoformer}, FedFormer~\cite{Fedformer}, and i-Transformer~\cite{Liu2024iTransformerIT}—and one CNN-based model, TimesNet~\cite{TimesNet}. These models are chosen for their strong empirical performance and computational efficiency across a wide range of multivariate time-series forecasting tasks.

Each of these baseline models can be easily integrated as a worker within the \syss framework with minimum efforts. \syss supports both homogeneous (identical) and heterogeneous (mixed) worker configurations. In our main experiments, we instantiate \syss with three identical baseline models as workers (i.e., number of workers = 3). These variants are denoted as \texttt{\sys-<Baseline>} (such as, \texttt{\sys-TimesNet}, \texttt{\sys-iTransformer}).

\parasum{Time-Series Foundation Model.}
We additionally compare against \textbf{Time-MoE}~\cite{Shi2025TimeMoEBT}, a large-scale time-series foundation model based on a sparse Mixture-of-Experts (MoE) architecture that routes each token to a subset of experts through a learned gating network. Time-MoE is a natural reference point for \sys, as both rely on routing inputs to specialized sub-models, but Time-MoE uses soft token-level routing inside a single pretrained network whereas \syss uses hard sample-level routing across independently trained workers. We run the publicly released Time-MoE checkpoint (Time-MoE-50M) in a zero-shot setting on exactly the same test windows, horizons, and normalization schemes as all other models, so that its results are directly comparable. 



\subsubsection{Hyperparameters}
\label{sec:hyperparams}
All baselines use the default hyperparameters of their official implementations in the Time-Series Library~\cite{thuml_tslib} (Appendix Table~\ref{app:worker_models}). Each worker in \sys~uses exactly the same configuration as the corresponding baseline, so that differences in accuracy reflect the framework rather than model tuning.

\parasum{Training.} We train all models with Adam and a batch size of 32. Following the standard TSLib settings, baselines are trained for up to 10 epochs with a learning rate of $10^{-4}$ and early stopping on the validation MSE (patience 3). In \sys, the workers use the same learning rate as their baselines, and the manager uses $10^{-5}$ ($10^{-3}$ for \sys-FEDformer), both with a weight decay of $10^{-5}$.  Because each sample trains only one of the $K=3$ workers per epoch, a worker sees about one third of the training data in each epoch. We therefore train \sys~for up to 30 epochs with early stopping (patience 5), three times the epoch budget of the baselines. All settings are listed in Appendix Table~\ref{app:train_hparams}.



\parasum{Normalization and evaluation.}
Because the scale of the input variables differs by orders of magnitude across datasets (e.g., bike counts vs.\ PM2.5 concentration), we evaluate every model under two common normalization schemes: (i)~\emph{global} normalization, where each variable is standardized with the mean and standard deviation of the training split, and (ii)~\emph{sample} normalization, where each input window is standardized with its own per-variable mean and standard deviation and the target window is scaled with the statistics of the target variable in the input window.
For both schemes, predictions are mapped back to the original scale before computing the error, so all reported MSE values are in the original units of the target variable and are comparable across models within a dataset, but not across datasets.

\definecolor{bestbg}{HTML}{FDAE6B}  
\definecolor{secondbg}{HTML}{FEE6CE}  
\DeclareRobustCommand{\bestbox}[1]{{\setlength{\fboxsep}{1pt}\colorbox{bestbg}{#1}}}
\DeclareRobustCommand{\secondbox}[1]{{\setlength{\fboxsep}{1pt}\colorbox{secondbg}{#1}}}
\begin{table*}[t]
\caption{Forecasting results (MSE on the original scale, mean over 5 runs) of each baseline (Base) and its \sys-enhanced version (+\sys) under global and sample normalization; look-back length $L=96$. The better of each baseline and its \sys-enhanced version is in \textbf{bold}; the best result in each row is marked with a \bestbox{dark orange} background and the second best with a \secondbox{light orange} background. \sys~Win: number of the 20 settings (5 datasets $\times$ 4 horizons) in which \sys-$<$baseline$>$ has a lower mean MSE than $<$baseline$>$. Impr.: relative MSE reduction of \sys-$<$baseline$>$ over $<$baseline$>$, averaged over the 20 settings. The very large MSE of some baselines on Weather is caused by missing-value codes in the training target (Appendix~\ref{app:weather}). Mean $\pm$ SD for every setting is listed in Appendix Tables~\ref{tab:main_global} and~\ref{tab:main_sample}.}
\label{tab:main}
\centering
\resizebox{\textwidth}{!}{%
\setlength{\tabcolsep}{2.5pt}
\begin{tabular}{|c|c|cc|cc|cc|cc|cc|cc|cc|cc|cc|cc|}
\hline
\multirow{3}{*}{Dataset} & \multirow{3}{*}{$L_y$} & \multicolumn{4}{c|}{iTransformer (\citeyear{Liu2024iTransformerIT})} & \multicolumn{4}{c|}{TimesNet (\citeyear{TimesNet})} & \multicolumn{4}{c|}{FEDformer (\citeyear{Fedformer})} & \multicolumn{4}{c|}{Autoformer (\citeyear{autoformer})} & \multicolumn{4}{c|}{Informer (\citeyear{informer})} \\
\cline{3-22}
 & & \multicolumn{2}{c|}{Global} & \multicolumn{2}{c|}{Sample} & \multicolumn{2}{c|}{Global} & \multicolumn{2}{c|}{Sample} & \multicolumn{2}{c|}{Global} & \multicolumn{2}{c|}{Sample} & \multicolumn{2}{c|}{Global} & \multicolumn{2}{c|}{Sample} & \multicolumn{2}{c|}{Global} & \multicolumn{2}{c|}{Sample} \\
\cline{3-22}
 & & Base & +\sys & Base & +\sys & Base & +\sys & Base & +\sys & Base & +\sys & Base & +\sys & Base & +\sys & Base & +\sys & Base & +\sys & Base & +\sys \\
\hline
\multirow{4}{*}{\rotatebox{90}{Weather}} & 96 & 204.7 & \textbf{194.9} & 447.0 & \textbf{178.3} & 245.6 & \textbf{215.9} & 418.6 & \cellcolor{secondbg}\textbf{158.5} & 28{,}741 & \textbf{351.3} & 1{,}208{,}418 & \textbf{181.7} & 39{,}860 & \textbf{419.5} & 1{,}933{,}907 & \textbf{210.1} & 1{,}733 & \textbf{442.7} & 4{,}402 & \cellcolor{bestbg}\textbf{155.7} \\
 & 192 & 249.3 & \textbf{226.7} & 637.6 & \textbf{231.6} & 271.1 & \textbf{234.8} & 410.6 & \cellcolor{secondbg}\textbf{226.1} & 28{,}712 & \textbf{548.9} & 10{,}628{,}690 & \textbf{236.8} & 230{,}293 & \textbf{380.4} & 151{,}937 & \textbf{285.4} & 6{,}645 & \textbf{426.5} & 7{,}309 & \cellcolor{bestbg}\textbf{220.8} \\
 & 336 & 254.6 & \cellcolor{bestbg}\textbf{232.3} & 749.2 & \textbf{254.8} & 264.8 & \textbf{243.3} & 456.7 & \textbf{241.8} & 50{,}859 & \textbf{559.7} & 380{,}781 & \textbf{267.3} & 70{,}468 & \textbf{279.8} & 157{,}994 & \textbf{270.1} & 77{,}472 & \textbf{460.2} & 3{,}729 & \cellcolor{secondbg}\textbf{240.1} \\
 & 720 & 270.1 & \cellcolor{bestbg}\textbf{235.9} & 748.3 & \textbf{330.0} & 269.1 & \cellcolor{secondbg}\textbf{241.7} & 963.3 & \textbf{291.8} & 41{,}001 & \textbf{640.8} & 798{,}364 & \textbf{268.3} & 209{,}899 & \textbf{259.7} & 29{,}367 & \textbf{296.1} & 4{,}335 & \textbf{527.5} & 3{,}637 & \textbf{290.7} \\
\hline
\multirow{4}{*}{\rotatebox{90}{PM2.5-LA}} & 96 & 37.05 & \textbf{37.02} & 36.08 & \cellcolor{bestbg}\textbf{35.20} & \textbf{37.32} & 44.42 & 40.18 & \cellcolor{secondbg}\textbf{35.43} & 57.75 & \textbf{42.29} & 60.91 & \textbf{42.11} & 86.25 & \textbf{71.18} & \textbf{57.60} & 58.89 & 45.34 & \textbf{45.04} & 41.30 & \textbf{37.90} \\
 & 192 & \textbf{40.64} & 41.17 & 40.93 & \textbf{40.81} & \textbf{40.16} & 42.52 & 42.62 & \cellcolor{bestbg}\textbf{39.25} & 65.63 & \textbf{45.25} & 64.07 & \textbf{43.77} & 66.43 & \textbf{64.94} & \textbf{61.37} & 65.62 & 46.80 & \textbf{42.66} & 42.06 & \cellcolor{secondbg}\textbf{39.62} \\
 & 336 & \textbf{48.71} & 49.35 & \textbf{48.89} & 49.82 & \textbf{45.91} & 48.54 & 46.42 & \cellcolor{bestbg}\textbf{44.67} & 67.49 & \textbf{53.96} & 74.34 & \textbf{52.00} & 70.55 & \textbf{69.27} & \textbf{70.79} & 70.81 & 48.22 & \cellcolor{secondbg}\textbf{44.79} & 52.84 & \textbf{48.27} \\
 & 720 & \textbf{62.55} & 64.17 & \textbf{63.62} & 63.88 & \textbf{59.76} & 61.11 & 59.23 & \textbf{58.73} & 75.79 & \textbf{66.34} & 80.34 & \textbf{64.62} & 106.7 & \textbf{72.36} & 142.7 & \textbf{72.23} & \cellcolor{secondbg}52.81 & \cellcolor{bestbg}\textbf{51.37} & \textbf{62.88} & 63.56 \\
\hline
\multirow{4}{*}{\rotatebox{90}{PM2.5-RV}} & 96 & 83.31 & \textbf{81.07} & 78.83 & \textbf{78.43} & 80.60 & \cellcolor{secondbg}\textbf{76.56} & 78.84 & \cellcolor{bestbg}\textbf{73.83} & 105.6 & \textbf{87.11} & 105.0 & \textbf{87.67} & 140.4 & \textbf{132.5} & \textbf{125.5} & 130.1 & \textbf{82.47} & 83.02 & 86.27 & \textbf{81.69} \\
 & 192 & 92.20 & \textbf{90.23} & 90.67 & \textbf{89.73} & 87.78 & \cellcolor{bestbg}\textbf{83.44} & 99.05 & \cellcolor{secondbg}\textbf{84.37} & 131.5 & \textbf{98.74} & 146.2 & \textbf{96.72} & 142.6 & \textbf{133.5} & \textbf{158.9} & 163.9 & \textbf{94.80} & 97.51 & 101.8 & \textbf{91.40} \\
 & 336 & \textbf{95.85} & 96.27 & \textbf{94.49} & 95.08 & 93.31 & \cellcolor{bestbg}\textbf{91.94} & 98.14 & \cellcolor{secondbg}\textbf{92.40} & 128.1 & \textbf{103.8} & 129.9 & \textbf{109.9} & \textbf{136.3} & 136.9 & 166.5 & \textbf{137.6} & \textbf{100.1} & 100.6 & \textbf{102.6} & 102.7 \\
 & 720 & \textbf{87.58} & 87.89 & 87.28 & \textbf{86.70} & 87.86 & \textbf{87.73} & \cellcolor{bestbg}\textbf{85.81} & \cellcolor{secondbg}86.12 & 115.2 & \textbf{94.83} & 104.5 & \textbf{95.63} & 128.4 & \textbf{110.8} & 122.7 & \textbf{112.5} & 92.70 & \textbf{89.88} & \textbf{94.57} & 94.62 \\
\hline
\multirow{4}{*}{\rotatebox{90}{Bike Share}} & 96 & 15{,}889 & \cellcolor{secondbg}\textbf{14{,}815} & 15{,}134 & \cellcolor{bestbg}\textbf{14{,}258} & 16{,}967 & \textbf{16{,}450} & 15{,}418 & \textbf{14{,}886} & 18{,}171 & \textbf{15{,}597} & 27{,}592 & \textbf{14{,}944} & 24{,}641 & \textbf{21{,}516} & 25{,}573 & \textbf{24{,}799} & 18{,}653 & \textbf{18{,}214} & 16{,}080 & \textbf{15{,}038} \\
 & 192 & 21{,}640 & \textbf{20{,}864} & 21{,}140 & \cellcolor{secondbg}\textbf{20{,}356} & 22{,}654 & \textbf{22{,}035} & \textbf{20{,}500} & 20{,}508 & 25{,}283 & \cellcolor{bestbg}\textbf{19{,}930} & 37{,}106 & \textbf{21{,}570} & 32{,}384 & \textbf{29{,}645} & \textbf{33{,}966} & 34{,}204 & \textbf{22{,}758} & 23{,}480 & \textbf{22{,}524} & 22{,}916 \\
 & 336 & 27{,}939 & \textbf{27{,}025} & 27{,}477 & \cellcolor{secondbg}\textbf{26{,}907} & 28{,}267 & \textbf{27{,}661} & 27{,}702 & \textbf{27{,}672} & 28{,}721 & \cellcolor{bestbg}\textbf{23{,}498} & 43{,}632 & \textbf{27{,}528} & 39{,}925 & \textbf{37{,}735} & 48{,}489 & \textbf{41{,}074} & \textbf{28{,}818} & 32{,}018 & \textbf{29{,}075} & 33{,}134 \\
 & 720 & \textbf{37{,}411} & 37{,}451 & \textbf{36{,}714} & 37{,}845 & \textbf{36{,}838} & 38{,}760 & \textbf{35{,}968} & 38{,}228 & \cellcolor{secondbg}34{,}864 & \cellcolor{bestbg}\textbf{32{,}858} & 53{,}511 & \textbf{39{,}902} & 44{,}139 & \textbf{39{,}982} & 51{,}277 & \textbf{47{,}750} & \textbf{34{,}947} & 35{,}416 & 37{,}201 & \textbf{36{,}790} \\
\hline
\multirow{4}{*}{\rotatebox{90}{Appliances}} & 96 & 8{,}095 & \cellcolor{secondbg}\textbf{8{,}084} & 17{,}088 & \textbf{10{,}267} & 9{,}864 & \textbf{8{,}350} & 16{,}593 & \textbf{11{,}356} & 10{,}091 & \textbf{9{,}253} & 17{,}445 & \textbf{13{,}584} & 16{,}935 & \textbf{9{,}941} & 76{,}717 & \textbf{12{,}599} & 8{,}921 & \cellcolor{bestbg}\textbf{7{,}950} & 38{,}936 & \textbf{11{,}884} \\
 & 192 & \cellcolor{secondbg}\textbf{8{,}123} & 8{,}129 & 12{,}600 & \textbf{11{,}150} & 10{,}121 & \textbf{8{,}468} & 22{,}151 & \textbf{12{,}079} & 9{,}436 & \textbf{9{,}027} & 26{,}191 & \textbf{12{,}355} & 13{,}646 & \textbf{10{,}650} & 97{,}072 & \textbf{13{,}580} & 9{,}172 & \cellcolor{bestbg}\textbf{7{,}739} & 52{,}661 & \textbf{12{,}043} \\
 & 336 & 7{,}847 & \cellcolor{secondbg}\textbf{7{,}761} & 14{,}408 & \textbf{10{,}362} & 9{,}843 & \textbf{8{,}044} & 22{,}152 & \textbf{10{,}624} & 9{,}147 & \textbf{8{,}559} & 46{,}912 & \textbf{12{,}294} & 11{,}739 & \textbf{10{,}198} & 80{,}913 & \textbf{12{,}261} & 8{,}329 & \cellcolor{bestbg}\textbf{7{,}681} & 47{,}692 & \textbf{15{,}531} \\
 & 720 & \cellcolor{secondbg}\textbf{8{,}046} & 8{,}048 & 16{,}877 & \textbf{11{,}478} & 8{,}861 & \textbf{8{,}280} & 18{,}186 & \textbf{11{,}308} & 10{,}063 & \textbf{9{,}080} & 42{,}264 & \textbf{12{,}637} & 13{,}574 & \textbf{9{,}706} & 53{,}329 & \textbf{11{,}398} & 8{,}261 & \cellcolor{bestbg}\textbf{7{,}870} & 31{,}365 & \textbf{12{,}709} \\
\hline
\multicolumn{2}{|c|}{\sys~Win} & \multicolumn{2}{c|}{12/20} & \multicolumn{2}{c|}{16/20} & \multicolumn{2}{c|}{15/20} & \multicolumn{2}{c|}{17/20} & \multicolumn{2}{c|}{20/20} & \multicolumn{2}{c|}{20/20} & \multicolumn{2}{c|}{19/20} & \multicolumn{2}{c|}{14/20} & \multicolumn{2}{c|}{14/20} & \multicolumn{2}{c|}{15/20} \\
\hline
\multicolumn{2}{|c|}{Impr.\ (\%)} & \multicolumn{2}{c|}{+2.4} & \multicolumn{2}{c|}{+18.4} & \multicolumn{2}{c|}{+4.1} & \multicolumn{2}{c|}{+21.9} & \multicolumn{2}{c|}{+32.6} & \multicolumn{2}{c|}{+47.8} & \multicolumn{2}{c|}{+30.9} & \multicolumn{2}{c|}{+40.8} & \multicolumn{2}{c|}{+20.0} & \multicolumn{2}{c|}{+34.1} \\
\hline
\end{tabular}
}
\end{table*}

\definecolor{tbestbg}{HTML}{FEE6CE}  
\DeclareRobustCommand{\tbestbox}[1]{{\setlength{\fboxsep}{1pt}\colorbox{tbestbg}{#1}}}
\begin{table}[t]
\centering
\caption{Best standalone baseline vs.\ best \sys-enhanced model vs.\ zero-shot Time-MoE~\cite{Shi2025TimeMoEBT}. For each setting, the best baseline (best \sys-enhanced model) is the lowest mean MSE over the five baselines (\sys-enhanced models) and the two normalization schemes; for Time-MoE (single deterministic run), the better of the two normalization schemes is used. The superscript gives the normalization (G: global, S: sample). The best result in each row is in \textbf{bold} with a \tbestbox{light orange} background. \sys~Win: number of the 20 settings in which the best \sys-enhanced model has a lower MSE than the best baseline (Time-MoE); Impr.: relative MSE reduction of the best \sys-enhanced model over the best baseline (Time-MoE), averaged over the 20 settings.}
\label{tab:best}
\small
\begin{tabular}{|c|c|c|c|c|c|c|}
\hline
\multirow{2}{*}{Dataset} & \multirow{2}{*}{$L_y$} & \multicolumn{2}{c|}{Best baseline} & \multicolumn{2}{c|}{Best \sys-enhanced} & Time-MoE (\citeyear{Shi2025TimeMoEBT}) \\
\cline{3-7}
 & & Model & MSE & Model & MSE & MSE \\
\hline
\multirow{4}{*}{Weather} & 96 & iTransformer & 204.7$^{\mathrm{G}}$ & \sys-Informer & \cellcolor{tbestbg}\textbf{155.7$^{\mathrm{S}}$} & 164.9$^{\mathrm{S}}$ \\
 & 192 & iTransformer & 249.3$^{\mathrm{G}}$ & \sys-Informer & 220.8$^{\mathrm{S}}$ & \cellcolor{tbestbg}\textbf{215.0$^{\mathrm{S}}$} \\
 & 336 & iTransformer & 254.6$^{\mathrm{G}}$ & \sys-iTransformer & 232.3$^{\mathrm{G}}$ & \cellcolor{tbestbg}\textbf{225.6$^{\mathrm{S}}$} \\
 & 720 & TimesNet & 269.1$^{\mathrm{G}}$ & \sys-iTransformer & 235.9$^{\mathrm{G}}$ & \cellcolor{tbestbg}\textbf{233.5$^{\mathrm{S}}$} \\
\hline
\multirow{4}{*}{PM2.5-LA} & 96 & iTransformer & 36.08$^{\mathrm{S}}$ & \sys-iTransformer & \cellcolor{tbestbg}\textbf{35.20$^{\mathrm{S}}$} & 39.66$^{\mathrm{S}}$ \\
 & 192 & TimesNet & 40.16$^{\mathrm{G}}$ & \sys-TimesNet & \cellcolor{tbestbg}\textbf{39.25$^{\mathrm{S}}$} & 43.16$^{\mathrm{S}}$ \\
 & 336 & TimesNet & 45.91$^{\mathrm{G}}$ & \sys-TimesNet & \cellcolor{tbestbg}\textbf{44.67$^{\mathrm{S}}$} & 51.37$^{\mathrm{G}}$ \\
 & 720 & Informer & 52.81$^{\mathrm{G}}$ & \sys-Informer & \cellcolor{tbestbg}\textbf{51.37$^{\mathrm{G}}$} & 60.59$^{\mathrm{G}}$ \\
\hline
\multirow{4}{*}{PM2.5-RV} & 96 & iTransformer & 78.83$^{\mathrm{S}}$ & \sys-TimesNet & \cellcolor{tbestbg}\textbf{73.83$^{\mathrm{S}}$} & 78.68$^{\mathrm{G}}$ \\
 & 192 & TimesNet & 87.78$^{\mathrm{G}}$ & \sys-TimesNet & \cellcolor{tbestbg}\textbf{83.44$^{\mathrm{G}}$} & 87.28$^{\mathrm{G}}$ \\
 & 336 & TimesNet & 93.31$^{\mathrm{G}}$ & \sys-TimesNet & 91.94$^{\mathrm{G}}$ & \cellcolor{tbestbg}\textbf{91.45$^{\mathrm{G}}$} \\
 & 720 & TimesNet & \cellcolor{tbestbg}\textbf{85.81$^{\mathrm{S}}$} & \sys-TimesNet & 86.12$^{\mathrm{S}}$ & 87.41$^{\mathrm{G}}$ \\
\hline
\multirow{4}{*}{Bike Share} & 96 & iTransformer & 15{,}134$^{\mathrm{S}}$ & \sys-iTransformer & \cellcolor{tbestbg}\textbf{14{,}258$^{\mathrm{S}}$} & 15{,}802$^{\mathrm{S}}$ \\
 & 192 & TimesNet & 20{,}500$^{\mathrm{S}}$ & \sys-FEDformer & \cellcolor{tbestbg}\textbf{19{,}930$^{\mathrm{G}}$} & 20{,}044$^{\mathrm{G}}$ \\
 & 336 & iTransformer & 27{,}477$^{\mathrm{S}}$ & \sys-FEDformer & \cellcolor{tbestbg}\textbf{23{,}498$^{\mathrm{G}}$} & 23{,}512$^{\mathrm{G}}$ \\
 & 720 & FEDformer & 34{,}864$^{\mathrm{G}}$ & \sys-FEDformer & 32{,}858$^{\mathrm{G}}$ & \cellcolor{tbestbg}\textbf{30{,}793$^{\mathrm{G}}$} \\
\hline
\multirow{4}{*}{Appliances} & 96 & iTransformer & 8{,}095$^{\mathrm{G}}$ & \sys-Informer & \cellcolor{tbestbg}\textbf{7{,}950$^{\mathrm{G}}$} & 8{,}954$^{\mathrm{G}}$ \\
 & 192 & iTransformer & 8{,}123$^{\mathrm{G}}$ & \sys-Informer & \cellcolor{tbestbg}\textbf{7{,}739$^{\mathrm{G}}$} & 8{,}596$^{\mathrm{G}}$ \\
 & 336 & iTransformer & 7{,}847$^{\mathrm{G}}$ & \sys-Informer & \cellcolor{tbestbg}\textbf{7{,}681$^{\mathrm{G}}$} & 8{,}145$^{\mathrm{G}}$ \\
 & 720 & iTransformer & 8{,}046$^{\mathrm{G}}$ & \sys-Informer & \cellcolor{tbestbg}\textbf{7{,}870$^{\mathrm{G}}$} & 8{,}511$^{\mathrm{G}}$ \\
\hline
\multicolumn{2}{|c|}{\sys~Win} & \multicolumn{2}{c|}{19/20} & \multicolumn{2}{c|}{--} & 15/20 \\
\hline
\multicolumn{2}{|c|}{Impr.\ (\%)} & \multicolumn{2}{c|}{+5.9} & \multicolumn{2}{c|}{--} & +4.9 \\
\hline
\end{tabular}
\end{table}


\parasum{Number of workers.} We use $K=3$ workers for all datasets and models. The suitable number of workers is dataset dependent (Section~\ref{sec:ablation}): on PM2.5-RV, $K=6$ slightly reduces the MSE ($-1.4\%$) and $K=9$ increases it ($+2.6\%$), and on Bike Share, neither improves the MSE (Appendix Table~\ref{app:ablation_full}); none of these differences is statistically significant except for $K=9$ on Bike Share. Since each additional worker adds a forward pass to every training iteration, we fix $K=3$ as a trade-off between accuracy and training cost.

\subsection{Evaluation Results}
\label{sec:comp1}

Table~\ref{tab:main} reports the MSE of each baseline and its \sys-enhanced version for every dataset and horizon under both normalization schemes, averaged over 5 independent runs. Lower MSE values indicate better predictive accuracy. The full results, i.e., the mean and standard deviation (SD) over 5 runs for every setting, are given in Appendix Tables~\ref{tab:main_global} (global normalization) and~\ref{tab:main_sample} (sample normalization), and the per-setting MSE reduction of each \sys-enhanced model over its baseline in Appendix Table~\ref{app:reduction}. Table~\ref{tab:main} also reports the mean MSE reduction of each \sys-enhanced model over its baseline. Because the error of a collapsed baseline run can be several orders of magnitude larger than that of a stable one (e.g., FEDformer and Autoformer on Weather), this mean is dominated by a few settings; we therefore focus on win counts.


\subsubsection{Baseline vs \sys-Enhanced}
Table~\ref{tab:main} presents the performance of each baseline model and its corresponding \sys-enhanced variant side by side for direct comparison.

\parasum{\texttt{\sys-<baseline>} consistently improves over the baseline models.}
Across the 100 baseline--variant pairs per normalization scheme (5 baselines $\times$ 5 datasets $\times$ 4 horizons), \texttt{\sys-<baseline>} achieves a lower MSE than its baseline in 80 pairs under global normalization and 82 pairs under sample normalization. The gains are consistent across horizons: \syss wins in 23, 20, 19, and 18 of the 25 pairs at $L_y=96, 192, 336, 720$ under global normalization (23, 20, 20, and 19 under sample normalization), so the benefit only slightly decreases as the horizon grows. Moreover, when each model uses its better normalization scheme, the best \sys-enhanced model outperforms the best standalone baseline in 19 of the 20 settings (Table~\ref{tab:best}).

\parasum{\syss improves every worker type, with the largest gains for the less stable baselines.}
Of the 20 settings (5 datasets $\times$ 4 horizons), \sys-iTransformer achieves a lower MSE than iTransformer in 12 under global and 16 under sample normalization, and \sys-TimesNet than TimesNet in 15 and 17; for these strong and recent workers, the improvements are smaller but still consistent. The gains are largest for the older and less stable workers: \sys-FEDformer outperforms FEDformer in all 20 settings under both schemes, \sys-Autoformer outperforms Autoformer in 19 and 14, and \sys-Informer outperforms Informer in 14 and 15.

\parasum{\syss stabilizes training.}
A striking effect of \syss is the reduction of run-to-run variance. Several standalone baselines occasionally diverge on the original-scale targets, most visibly FEDformer, Autoformer, and Informer on Weather, whose MSE rises to $10^3$--$10^5$ under global and up to $10^7$ under sample normalization, with an SD of the same order. This instability can be traced to the missing-value codes in the training target of Weather (Appendix~\ref{app:weather}). Their \sys-enhanced counterparts remain within the range of the strong baselines (e.g., FEDformer on Weather at $L_y=96$: $28{,}741 \pm 16{,}160$ vs.\ $351.3 \pm 97.8$ with \sys). Because each worker only needs to fit the subset of contexts routed to it, the manager--worker decomposition makes the optimization of each worker considerably easier.

\parasum{The best \sys-enhanced model depends on the dataset.}
Across both normalization schemes, a \sys-enhanced model obtains the best result in 19 of the 20 settings of Table~\ref{tab:main}; the only exception is PM2.5-RV at $L_y=720$, where the standalone TimesNet is best. The best worker type depends on the dataset: \sys-Informer on Appliances, the short horizons of Weather, and PM2.5-LA at $L_y=720$; \sys-iTransformer on the long horizons of Weather and at $L_y=96$ on PM2.5-LA and Bike Share; \sys-FEDformer on Bike Share at $L_y \geq 192$; and \sys-TimesNet on PM2.5-RV and PM2.5-LA at $L_y \in \{192, 336\}$. Within each normalization scheme, a CATF-enhanced model is best in 18 of the 20 settings, both under global and under sample normalization (Appendix Tables~\ref{tab:main_global} and~\ref{tab:main_sample}).


\subsubsection{\sys{} vs.\ Time-MoE}
Table~\ref{tab:best} compares, for each setting, the best standalone baseline, the best \sys-enhanced model, and zero-shot Time-MoE, each under its better normalization scheme. Time-MoE is evaluated zero-shot and is deterministic, so it is reported from a single run. The best \sys-enhanced model achieves a lower MSE than the best baseline in 19 of the 20 settings and than Time-MoE in 15 of the 20 settings (mean MSE reduction 4.9\%). Time-MoE is better in the five settings, mostly at longer horizons (Weather for $L_y \geq 192$, PM2.5-RV for $L_y=336$, and Bike Share for $L_y=720$).
Overall, \syss allows comparatively small task-specific models to match or exceed a large pretrained foundation model on most context-rich forecasting tasks, while using a single worker per input at inference time.

\subsection{Ablation Study}
\label{sec:ablation}

We conduct the ablation study with \sys-iTransformer on PM2.5-RV under global normalization and $L=L_y=96$. The default configuration is identical to that of the main experiments: three homogeneous workers, a two-layer Transformer manager ($d_{model}=512$, $d_{ff}=2048$, 8 heads, dropout 0.05), and the combined KL and Wasserstein manager loss with group-lasso regularization. In each ablation, we change exactly one component and keep the others fixed. Each setting is repeated over 5 runs without fixed random seeds, and we test the difference from the default with Welch's $t$-test. The default reaches an MSE of 80.52, consistent with Table~\ref{tab:main} (81.07). Table~\ref{tab:ablation} summarizes the results; Appendix~\ref{app:ablation} reports the same ablation on Bike Share.

Overall, removing or changing any component of the manager increases the MSE by 0.5--1.7\%. Changing the number or mix of workers changes the MSE by $-1.4$\% to $+2.6$\%; $K{=}6$ and the mixture of the top-2 models are slightly better than the default. None of these differences is statistically significant ($p \geq 0.13$), so \syss is robust to these design choices. On Bike Share, the results are similar, except that nine workers and mixing workers with different architectures significantly degrade the MSE (Appendix Table~\ref{app:ablation_full}).

\begin{table}[t]
\caption{Ablation study of \sys-iTransformer on PM2.5-RV (global normalization, $L=L_y=96$). MSE is mean $\pm$ SD over 5 runs without fixed random seeds; the default setting is an independent re-run of the configuration in Table~\ref{tab:main} (MSE $81.07 \pm 1.18$ there). $\Delta$ is the relative change in mean MSE w.r.t.\ the default (positive is worse); $p$ is from Welch's $t$-test against the default (bold: $p<0.05$). The results on Bike Share are reported in Appendix Table~\ref{app:ablation_full}.}
\label{tab:ablation}
\centering
\small
\setlength{\tabcolsep}{3pt}
\begin{tabular}{llrrr}
\toprule
\multicolumn{2}{l}{\textbf{Setting}} & \textbf{MSE} & $\Delta$ (\%) & $p$ \\
\midrule
\multicolumn{2}{l}{Default} & 80.52 $\pm$ 1.02 & -- & -- \\
\midrule
\multirow{2}{*}{Loss} & KL only & 81.87 $\pm$ 2.75 & $+1.7$ & 0.347 \\
 & Wasserstein only & 81.56 $\pm$ 1.10 & $+1.3$ & 0.157 \\
\midrule
\multirow{1}{*}{Reg.} & No reg. & 80.94 $\pm$ 1.19 & $+0.5$ & 0.559 \\
\midrule
\multirow{2}{*}{Manager} & Small ($d_{model}{=}128$) & 81.23 $\pm$ 0.78 & $+0.9$ & 0.252 \\
 & Large ($d_{model}{=}768$) & 81.05 $\pm$ 0.81 & $+0.7$ & 0.386 \\
\midrule
\multirow{2}{*}{Workers} & $K{=}6$ & 79.37 $\pm$ 1.14 & $-1.4$ & 0.132 \\
 & $K{=}9$ & 82.61 $\pm$ 2.99 & $+2.6$ & 0.200 \\
\midrule
\multirow{2}{*}{Mix} & Top-3 models$^{a}$ ($K{=}3$) & 81.63 $\pm$ 2.13 & $+1.4$ & 0.334 \\
 & Top-2 models$^{a}$ $\times$2 ($K{=}4$) & 80.04 $\pm$ 2.70 & $-0.6$ & 0.728 \\
\bottomrule
\end{tabular}
\\[2pt]
{\footnotesize $^{a}$Top-performing models on PM2.5-RV in Appendix Table~\ref{tab:main_global}: top-3 = TimesNet + iTransformer + Informer, top-2 = TimesNet + iTransformer.}
\end{table}


\subsubsection{Manager Model Loss Function}
The manager loss (Equation~\ref{eq:manager_loss}) consists of a divergence term, which combines the KL divergence and the Wasserstein distance, and a group-lasso regularization term on the manager's parameters.

\parasum{Divergence Term.} Removing either divergence term increases the error. With the KL divergence alone, the MSE increases by 1.7\% ($p=0.35$), and its run-to-run SD is almost three times that of the default (2.75 vs.\ 1.02). With the Wasserstein distance alone, the MSE increases by 1.3\% ($p=0.16$). On Bike Share, the Wasserstein distance alone also over-concentrates the routing: the most used worker receives 73\% of the test samples on average, compared with 49\% for the default. The combined loss therefore achieves the lowest mean error, although the differences are not statistically significant, and we use it as the default.

\parasum{Regularization.} Removing the regularization term slightly increases the MSE ($+0.5\%$, $p=0.56$). We keep it in the default configuration.

\subsubsection{Manager Model Architecture}
We test a smaller manager ($d_{model}=128$, $d_{ff}=512$, 4 heads) and a larger one ($d_{model}=768$, $d_{ff}=3076$, 12 heads). Both slightly degrade the MSE, by 0.9\% and 0.7\%, although the differences are not statistically significant ($p \geq 0.25$). The larger manager doubles the total number of parameters of \sys-iTransformer (15.5M vs.\ 7.3M) without improving the accuracy, whereas the smaller one reduces it to 1.1M at a small cost in accuracy. The default manager capacity therefore gives the best accuracy among the tested sizes, and \syss is not highly sensitive.

\subsubsection{Number of Workers} \label{ab:worker_number}
Increasing the number of workers from $K=3$ to $K=6$ reduces the MSE by 1.4\% ($p=0.13$), whereas $K=9$ increases it by 2.6\% ($p=0.20$) and leads to the largest run-to-run variation (SD 2.99). PM2.5-RV has 14 context variables from on-site meteorology, reanalysis data, and co-pollutants, which may benefit from a moderately larger number of workers. On Bike Share, whose context consists of a small number of calendar and weather variables, neither $K=6$ nor $K=9$ improves the MSE. The suitable $K$ is therefore dataset dependent, and we use $K=3$ in the main experiments as a trade-off between accuracy and cost. \begin{yqc} \end{yqc} A promising direction for future work is to determine the number of workers adaptively, e.g., by estimating the number of latent context regimes in the data.

\subsubsection{Mixture of Workers}
We also explore heterogeneous worker configurations by combining the top-performing models on PM2.5-RV, i.e., 1$\times$TimesNet + 1$\times$iTransformer + 1$\times$Informer or 2$\times$TimesNet + 2$\times$iTransformer. Both mixtures perform on par with the default ($+1.4\%$ and $-0.6\%$, $p \geq 0.33$), with a larger run-to-run variation (SD 2.13 and 2.70). On Bike Share, the corresponding 3-model mixture significantly degrades the MSE by 13.1\% ($p<0.001$). Thus, heterogeneous mixtures do not consistently improve forecasting accuracy, and we use homogeneous workers in the main experiments. A promising direction for future work is to explore targeted regularization or adaptive routing strategies to better leverage the complementary strengths of diverse worker models.

\section{Case Study: Interpreting the Routing}
\label{sec:context}
\label{sec:case_bike}

Because each prediction of \sys~is produced by a single Worker, which Worker handles which input can be observed directly, without the approximations of post-hoc explanation methods such as attention weights~\cite{lim2021temporal} or feature attributions~\cite{crabbe2021explaining, queen2023encoding}. We examine whether these routing decisions reflect the context on Bike Share, where the effect of the context is well known: bike demand has two commuting peaks on working days and a single midday peak on weekends. We analyze \sys-iTransformer ($L=L_y=96$, global normalization, $K=3$, five runs) with three questions: \textbf{(i) does the learned routing improve the forecasts}, \textbf{(ii) which input variables drive the routing}, and \textbf{(iii) what has each Worker learned}. As a sanity check~\cite{adebayo2018sanity}, we repeat each analysis with an untrained Manager of the same architecture; a pattern that also appears for the untrained Manager reflects the input data rather than what the Manager has learned. On synthetic data with known regimes, the same analysis recovers the true regimes, the variable that determines them, and the specialization of the Workers (Appendix~\ref{app:context_syn}). Workers are labeled by test usage ($W_1$ is the most used); details of the analysis are given in Appendix~\ref{app:context_methods}.

\begin{figure*}[tbp]
\centering
\includegraphics[width=\textwidth]{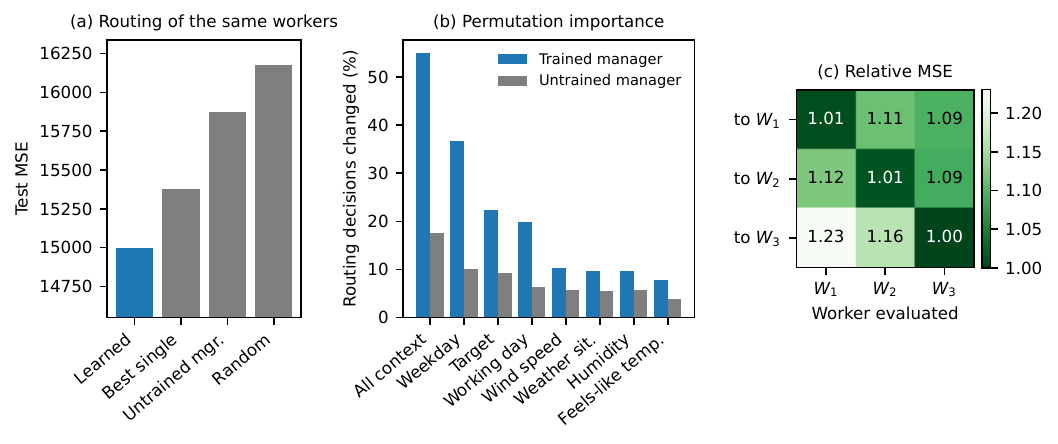}
\caption{Routing of \sys-iTransformer on Bike Share (mean over 5 runs). (a) Test MSE of the same trained workers under the learned routing, the best single worker for all samples, an untrained manager, and random routing with the same worker shares. (b) Share of routing decisions that change when one variable is shuffled across test samples in the manager input only (top 8 of 12 inputs; ``All context'' shuffles all context variables at once), for the trained and an untrained manager. (c) MSE of each worker on the samples routed to each worker, divided by the lowest MSE in the row and averaged over runs.}
\label{app:bike_routing}
\end{figure*}

\definecolor{wone}{HTML}{1F77B4}\definecolor{wtwo}{HTML}{FF7F0E}\definecolor{wthree}{HTML}{2CA02C}  
\begin{figure*}[tbp]
\centering
\includegraphics[width=\textwidth]{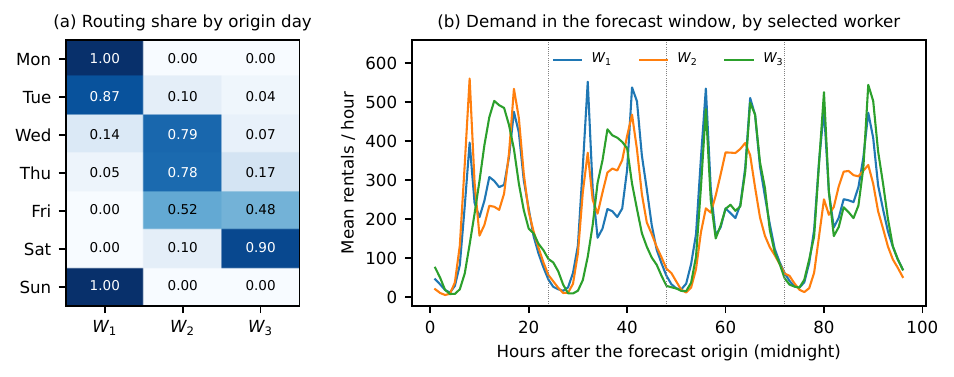}
\caption{What each worker of \sys-iTransformer handles on Bike Share (one run). (a) Share of the forecasts made on each day of the week that are routed to each worker. (b) Mean bike rentals in the 96-hour forecast window for the forecasts made at midnight, grouped by the selected worker. \textcolor{wone}{$W_1$} handles the forecasts made on Sunday to Tuesday, whose window consists mostly of working days (29 forecasts, on average 3.4 working days in the window); \textcolor{wtwo}{$W_2$} handles the forecasts made on Wednesday to Friday, whose window reaches into the weekend (29 forecasts, 2.3 working days); \textcolor{wthree}{$W_3$} handles the forecasts made on Friday and Saturday, whose window starts with the weekend (11 forecasts, 1.9 working days).}
\label{app:bike_profiles}
\end{figure*}

\parasum{Does the learned routing improve the forecasts?} During inference, we keep the trained Workers unchanged and replace only the Manager's routing with alternative assignments, so that any change in accuracy is due to the routing alone. The learned routing (MSE $15{,}000$) outperforms random routing with the same Worker shares ($16{,}175$) and the untrained Manager ($15{,}876$) in all five runs, and the best single Worker for all samples ($15{,}376$) in three of five runs (Figure~\ref{app:bike_routing}a). The gain comes from the specialization of the Workers. For the samples that the Manager routes to a given Worker, Figure~\ref{app:bike_routing}c compares the MSE of all three Workers: the selected Worker is the most accurate, and the other two Workers have a 9--23\% higher MSE. Each Worker is thus better than the others on its own samples, so the routing must send each sample to the right Worker; random routing, which ignores the context, loses this advantage.

\parasum{Which input variables drive the routing?} We shuffle one input variable at a time across the test samples in the Manager input only and record the share of routing decisions that change (permutation importance~\cite{breiman2001random}). The day of the week is by far the most important variable (37\% of the decisions change, vs.\ 10\% for the untrained Manager), followed by the target history (22\% vs.\ 9\%) and the working-day indicator (20\% vs.\ 6\%), whereas each weather variable changes only 8--10\% (Figure~\ref{app:bike_routing}b). A counterfactual analysis shows that these calendar variables, i.e., the day of the week and the working-day indicator, cause the routing: when the day of the week of the input window is shifted by $d \in \{1, \dots, 6\}$ days, 55\% of the decisions change, and 80\% of the new decisions match the typical Worker of the shifted day; when instead the target history is replaced with that of $d$ days later and the calendar is kept, only 11\% change (Appendix Figure~\ref{app:bike_counterfactual}).

\parasum{What has each Worker learned?} As shown above, the routing is driven by the calendar variables, i.e., the day of the week and the working-day indicator, which together determine how many working days and weekend days the forecast window contains. We therefore first examine how the Workers divide the week. The Manager splits the week by the days in the 96-hour forecast window (Figure~\ref{app:bike_profiles}a): forecasts made on Sunday to Tuesday, whose window consists mostly of working days, go to $W_1$; forecasts made on Wednesday to Friday, whose window reaches into the weekend, mostly to $W_2$; and forecasts made on Friday and Saturday, whose window starts with the weekend, mostly to $W_3$. Each Worker therefore forecasts a different demand trajectory (Figure~\ref{app:bike_profiles}b): demand recovering from the weekend, demand declining toward the weekend, and the weekend itself. The other context variables are only weakly related to the routing: the NMI between the selected Worker and the day of the week is 0.44, compared with at most 0.05 for the holiday indicator, the month, each weather variable, and the recent demand (Appendix Figure~\ref{app:bike_context_heatmap}). The context of the samples handled by each Worker shows the same picture (Appendix Figure~\ref{app:bike_feature_profiles}): each Worker has learned the patterns of a different context, and the differences between the Workers are largest for the context variables that drive the routing. Over the input and forecast windows, the Workers differ most in the working-day share, by 1.2 standard deviations between the Worker means, and accordingly in the demand trajectory, so each Worker is trained on and specializes in a different weekly pattern. For the weather variables, the differences are much smaller (0.2--0.4 standard deviations) and largely reflect the days on which the weather occurred; the weather has low importance because the Manager sees only the past weather, which says little about the forecast window. All five runs partition the samples by the day of the week, although the boundaries between the Workers differ between runs (Appendix~\ref{app:context_bike}).

In summary, the routing of \sys~on Bike Share is consistent with domain knowledge: the Manager infers the upcoming working-day/weekend pattern from the calendar, and each phase of the week is handled by a Worker that specializes in it.

\FloatBarrier

\section{Discussion}
\label{sec:discussion}

\parasum{When does \sys~help?} The routing is most useful when the forecasting regime is determined by the context. On Bike Share, the calendar determines whether the forecast window contains working days or the weekend, and the learned routing clearly outperforms random routing (Section~\ref{sec:context}); on synthetic data with context-driven regimes, the Manager recovers the regimes almost perfectly (Appendix~\ref{app:context_syn}). On PM2.5-RV, where the concentration varies with continuously changing meteorological conditions, the Workers barely specialize and the learned routing is on par with the best single Worker (Appendix~\ref{app:context_rv}). The routing analysis therefore also indicates whether a dataset contains context-driven regimes. Independently of the routing, splitting the data among several Workers stabilizes the training of baselines that diverge on Weather, whose training target contains missing-value codes (Section~\ref{sec:comp1}).

\parasum{Interpretability.} The routing decisions are an exact description of which Worker produced a forecast, and they can be related to the context with simple tools such as permutation importance and counterfactuals. They explain, however, only which Worker is used, not how this Worker arrives at its forecast. Moreover, the partition learned by the Manager can differ between runs, e.g., in where the boundaries between the Workers lie or in whether holidays are treated as weekends (Appendix~\ref{app:context_bike}); interpretations should therefore be checked across several runs.

\parasum{Limitations and future work.} \sys~trains the Manager and all Workers jointly and therefore needs considerably more training time and memory than a single model (Appendix Table~\ref{app:cost}); at inference time, only the Manager and the selected Worker are executed. The suitable number of Workers is dataset dependent and is currently set by hand (Section~\ref{sec:ablation}), and mixtures of different worker architectures did not consistently improve the accuracy. Promising directions are to determine the number of Workers adaptively, e.g., by estimating the number of latent context regimes, to better exploit heterogeneous Workers, and to apply the Manager--Worker design to other context-dependent prediction tasks, such as trajectory forecasting or human behavior modeling.

\section{Conclusion}

We introduced \sys, a context-aware forecasting framework in which a Transformer-based Manager routes each input to one of several Worker models, trained jointly so that the Workers specialize without manual data partitioning. Experiments on five context-rich datasets, four forecasting horizons, and two normalization schemes show that \sys~consistently improves its standalone counterparts, reduces their run-to-run variance, and is competitive with a pretrained foundation model, and a case study on bike-sharing data shows that its routing decisions follow the context in an interpretable way.

\bibliographystyle{ACM-Reference-Format}
\bibliography{reference}

\clearpage
\appendix
\newcommand{\apptocsec}[2]{\par\medskip\noindent\textbf{\makebox[1.8em][l]{\ref{#1}}#2}\dotfill\textbf{\pageref{#1}}\par}
\newcommand{\apptocsub}[2]{\par\noindent\hspace*{1.8em}\makebox[2.4em][l]{\ref{#1}}#2\dotfill\pageref{#1}\par}
\section*{Appendix Contents}
{\normalsize\linespread{1.3}\selectfont
\apptocsec{app:ex}{Experiment Details}
\apptocsub{app:ex_data}{Dataset}
\apptocsub{app:ex_hparams}{Hyper-parameters}
\apptocsec{app:results}{Complete Results of Forecasting Tasks}
\apptocsub{app:res_full}{Full Results}
\apptocsub{app:res_reduction}{Reduction Percentage}
\apptocsub{app:res_mae}{MAE Results}
\apptocsub{app:res_norm}{Global vs.\ Sample Normalization}
\apptocsub{app:res_bestnorm}{Best Normalization per Model}
\apptocsub{app:weather}{Instability of the Baselines on Weather}
\apptocsub{app:res_cost}{Computational Cost}
\apptocsec{app:ablation}{Ablation Study}
\apptocsec{app:context}{Details of the Case Study}
\apptocsub{app:context_methods}{Analysis Methods}
\apptocsub{app:context_syn}{Synthetic Data: Recovering Known Regimes}
\apptocsub{app:context_bike}{Bike Share: Additional Results}
\apptocsub{app:context_rv}{PM2.5-RV}
}
\clearpage


\section{Experiment Details}
\label{app:ex}

\subsection{Dataset}
\label{app:ex_data}

We conducted our comparative experiments on 5 datasets: Weather, PM2.5-LA, PM2.5-RV, Bike Share, and Appliances. For every dataset, the timestamp column is dropped, the last column is used as the forecasting target, and all remaining columns are used as context. The details of the datasets are presented in Table~\ref{app:dataset}.

\begin{table*}[h]
\centering
\caption{Statistics of the datasets used in the forecasting tasks. \#Variables counts all numeric columns, including the target. Dataset Size is the number of time steps.}
\label{app:dataset}
\resizebox{\textwidth}{!}{%
\begin{tabular}{lccccccc}
\toprule
\textbf{Dataset} & \textbf{\#Variables} & \textbf{Target} & \textbf{Input Length ($L$)} & \textbf{Output Length ($L_y$)} & \textbf{Frequency} & \textbf{Dataset Size} & \textbf{Domain} \\
\midrule
Weather    & 21 & CO$_2$ (ppm)      & 96 & \{96, 192, 336, 720\} & 10 minutes & 52,696 & Meteorology \\
PM2.5-LA   & 15 & PM2.5 ($\mu$g/m$^3$) & 96 & \{96, 192, 336, 720\} & 1 hour     & 35,063 & Air quality \\
PM2.5-RV   & 15 & PM2.5 ($\mu$g/m$^3$) & 96 & \{96, 192, 336, 720\} & 1 hour     & 35,063 & Air quality \\
Bike Share & 12 & Rental count      & 96 & \{96, 192, 336, 720\} & 1 hour     & 17,379 & Transportation \\
Appliances & 26 & Energy use (Wh)   & 96 & \{96, 192, 336, 720\} & 10 minutes & 19,735 & Energy \\
\bottomrule
\end{tabular}
}
\end{table*}

\subsection{Hyper-parameters}
\label{app:ex_hparams}
To ensure a much fairer comparison, we adopted baseline hyperparameters from scripts in Time-Series Library \cite{thuml_tslib}, which are the same as worker models, presented in Table~\ref{app:worker_models}. For the manager, we used the same default setting of hyperparameters (Table~\ref{app:manager_models}), except for fine-tuning the manager model learning rates for some experiments (Table~\ref{app:manager_lr}).

\subsubsection{Worker Models}
We adopted the same worker model hyperparameter as the baseline. All models have the same hyperparameter settings across all datasets, as follows.



\begin{table}[h]
\centering
\caption{Summary of worker models and baselines' hyperparameters used in our CATF framework.}
\label{app:worker_models}
\resizebox{\textwidth}{!}{%
\begin{tabular}{lccccccc}
\toprule
\textbf{Model} & \textbf{\#Layers} & \textbf{\#Model Dimension} & \textbf{\#Feedforward Dimension} & \textbf{\#Attention Heads} & \textbf{\#Dropout } & \textbf{\#Attention Factor} & \textbf{\#Learning Rate} \\
\midrule
i-Transformer     & 2 & 128 & 128 & 8 & 0.1   & 3  & 0.0001  \\
TimesNet      & 2 & 32 & 32 & 8 & 0.1  & 3 & 0.0001 \\
FEDformer       & 2 & 512 & 2048 & 8 & 0.1  & 3 & 0.0001 \\
Autoformer        & 2 & 512 & 2048 & 8 & 0.1  & 3 & 0.0001  \\
Informer        & 2 & 512 & 2048 & 8 & 0.1  & 3 & 0.0001 \\
\bottomrule
\end{tabular}
}
\end{table}

\subsubsection{Manager Models}

For all the training of CATF-<baseline> models in comparative experiments, we use the same hyperparameters of the manager model architecture, except we fine-tuned some manager models' learning rates for some datasets.

Here are the default settings of the hyperparameters for the manager model architecture in Table~\ref{app:manager_models}.

\begin{table}[h]
\centering
\caption{Default setting of the manager model hyperparameter that is used in our CATF framework.}
\label{app:manager_models}
\resizebox{\textwidth}{!}{%
\vspace{-2em} 
\begin{tabular}{lccccc}
\toprule
\textbf{Model Type} & \textbf{\#Layers} & \textbf{\#Model Dimension} & \textbf{\#Feedforward Dimension} & \textbf{\#Attention Heads} & \textbf{\#Dropout } \\
\midrule
Transformer     & 2 & 512 & 2048 & 8 & 0.05     \\
\bottomrule
\end{tabular}
}
\end{table}

Table~\ref{app:manager_lr} summarizes the learning rates and optimizer settings used for the manager and the workers. The same values are used for all datasets, horizons, and normalization schemes.

\begin{table}[H]
\centering
\caption{Learning rates of the manager and worker models in \sys-enhanced models (identical for all datasets). Both are optimized with Adam and a weight decay of $10^{-5}$.}
\label{app:manager_lr}
\begin{tabular}{lcc}
\toprule
\textbf{Model} & \textbf{Manager LR} & \textbf{Worker LR} \\
\midrule
CATF-iTransformer & 0.00001 & 0.0001 \\
CATF-TimesNet     & 0.00001 & 0.0001 \\
CATF-FEDformer    & 0.001   & 0.0001 \\
CATF-Autoformer   & 0.00001 & 0.0001 \\
CATF-Informer     & 0.00001 & 0.0001 \\
\bottomrule
\end{tabular}
\end{table}

\subsubsection{Training}

In Table \ref{app:train_hparams}, we present the shared and CATF-specific training settings, including split sizes, epochs, etc.

\begin{table}[H]
\centering
\caption{Training hyperparameters for CATF and Baseline models.}
\label{app:train_hparams}
\begin{tabular}{lcc}
\toprule
\textbf{Hyperparameter} & \textbf{Baseline} & \textbf{CATF Models} \\
\midrule
\multicolumn{3}{l}{\textit{Shared Settings}} \\
\midrule
Batch Size                  & 32 & 32 \\
Training Size & 0.8 & 0.8 \\
Validation Size & 0.1 & 0.1\\
Test Size & 0.1 & 0.1\\
Epochs                      & 10 & 30 \\
Early Stopping Patience     & 3 & 5 \\
Optimizer & Adam & Adam (weight decay $10^{-5}$) \\
Runs per setting & 5 & 5 \\
\midrule
\multicolumn{3}{l}{\textit{CATF-Specific Settings}} \\
\midrule
Number of Workers                & - & 3\\
Gradient Clipping Value             & - & 1\\
Worker Update Steps                & - & 2\\
\bottomrule
\end{tabular}

\end{table}

\FloatBarrier

\section{Complete Results of Forecasting Tasks}
\label{app:results}

\subsection{Full Results}
\label{app:res_full}
Tables~\ref{tab:main_global} and~\ref{tab:main_sample} report the MSE of every baseline and its \sys-enhanced version for each dataset and horizon under global and sample normalization (mean $\pm$ SD over 5 runs). They complement Table~\ref{tab:main}, which reports the mean MSE only.

\begingroup
\renewcommand{\arraystretch}{1.25}
\setlength\extrarowheight{1pt}
\begin{table*}[t]
    \caption{Long-term forecasting results under \textbf{Global} normalization. MSE is computed on the original (de-normalized) scale and reported as MEAN $\pm$ SD over 5 runs. The look-back length is 96 for all settings. The better of each baseline and its \sys-enhanced version is in \textbf{bold}; the best result in each row is marked with a \bestbox{dark orange} background and the second best with a \secondbox{light orange} background. \sys~Win: number of the 20 settings (5 datasets $\times$ 4 horizons) in which \sys-$<$baseline$>$ has a lower mean MSE than $<$baseline$>$. Impr.: relative MSE reduction of \sys-$<$baseline$>$ over $<$baseline$>$, averaged over the 20 settings.}
    \label{tab:main_global}
    \centering
    \resizebox{\textwidth}{!}{%
    \begin{tabular}{|l|c|cc|cc|cc|cc|cc|}
        \hline
        Dataset & $L_y$ & \shortstack{ iTransformer \\ (\citeyear{Liu2024iTransformerIT}) } & \shortstack{ \sys-\\iTransformer } & \shortstack{ TimesNet \\ (\citeyear{TimesNet}) } & \shortstack{ \sys-\\TimesNet } & \shortstack{ FEDformer \\ (\citeyear{Fedformer}) } & \shortstack{ \sys-\\FEDformer } & \shortstack{ Autoformer \\ (\citeyear{autoformer}) } & \shortstack{ \sys-\\Autoformer } & \shortstack{ Informer \\ (\citeyear{informer}) } & \shortstack{ \sys-\\Informer } \\
        \hline
        \multirow{4}{*}{Weather} & 96 & \cellcolor{secondbg}204.7 $\pm$ 2.29 & \cellcolor{bestbg}\textbf{194.9 $\pm$ 5.19} & 245.6 $\pm$ 10.69 & \textbf{215.9 $\pm$ 21.53} & 28,741 $\pm$ 16,160 & \textbf{351.3 $\pm$ 97.78} & 39,860 $\pm$ 35,693 & \textbf{419.5 $\pm$ 104.1} & 1,733 $\pm$ 830.7 & \textbf{442.7 $\pm$ 35.52} \\
          & 192 & 249.3 $\pm$ 3.11 & \cellcolor{bestbg}\textbf{226.7 $\pm$ 8.12} & 271.1 $\pm$ 13.23 & \cellcolor{secondbg}\textbf{234.8 $\pm$ 8.91} & 28,712 $\pm$ 19,726 & \textbf{548.9 $\pm$ 19.67} & 230,293 $\pm$ 340,357 & \textbf{380.4 $\pm$ 53.86} & 6,645 $\pm$ 10,171 & \textbf{426.5 $\pm$ 44.44} \\
          & 336 & 254.6 $\pm$ 2.15 & \cellcolor{bestbg}\textbf{232.3 $\pm$ 3.39} & 264.8 $\pm$ 3.40 & \cellcolor{secondbg}\textbf{243.3 $\pm$ 6.20} & 50,859 $\pm$ 39,591 & \textbf{559.7 $\pm$ 79.00} & 70,468 $\pm$ 97,935 & \textbf{279.8 $\pm$ 13.24} & 77,472 $\pm$ 170,210 & \textbf{460.2 $\pm$ 41.99} \\
          & 720 & 270.1 $\pm$ 2.85 & \cellcolor{bestbg}\textbf{235.9 $\pm$ 6.54} & 269.1 $\pm$ 5.97 & \cellcolor{secondbg}\textbf{241.7 $\pm$ 4.23} & 41,001 $\pm$ 22,172 & \textbf{640.8 $\pm$ 66.33} & 209,899 $\pm$ 267,336 & \textbf{259.7 $\pm$ 12.00} & 4,335 $\pm$ 4,652 & \textbf{527.5 $\pm$ 59.48} \\
        \hline
        \multirow{4}{*}{PM2.5-LA} & 96 & \cellcolor{secondbg}37.05 $\pm$ 0.39 & \cellcolor{bestbg}\textbf{37.02 $\pm$ 0.28} & \textbf{37.32 $\pm$ 1.68} & 44.42 $\pm$ 3.64 & 57.75 $\pm$ 4.31 & \textbf{42.29 $\pm$ 1.29} & 86.25 $\pm$ 24.26 & \textbf{71.18 $\pm$ 6.13} & 45.34 $\pm$ 4.37 & \textbf{45.04 $\pm$ 6.22} \\
          & 192 & \cellcolor{secondbg}\textbf{40.64 $\pm$ 0.30} & 41.17 $\pm$ 0.24 & \cellcolor{bestbg}\textbf{40.16 $\pm$ 1.94} & 42.52 $\pm$ 2.36 & 65.63 $\pm$ 1.73 & \textbf{45.25 $\pm$ 0.86} & 66.43 $\pm$ 8.03 & \textbf{64.94 $\pm$ 4.65} & 46.80 $\pm$ 3.08 & \textbf{42.66 $\pm$ 0.25} \\
          & 336 & \textbf{48.71 $\pm$ 0.14} & 49.35 $\pm$ 0.76 & \cellcolor{secondbg}\textbf{45.91 $\pm$ 2.01} & 48.54 $\pm$ 1.97 & 67.49 $\pm$ 2.75 & \textbf{53.96 $\pm$ 2.78} & 70.55 $\pm$ 5.09 & \textbf{69.27 $\pm$ 2.78} & 48.22 $\pm$ 2.55 & \cellcolor{bestbg}\textbf{44.79 $\pm$ 0.57} \\
          & 720 & \textbf{62.55 $\pm$ 0.37} & 64.17 $\pm$ 0.42 & \textbf{59.76 $\pm$ 1.06} & 61.11 $\pm$ 2.93 & 75.79 $\pm$ 5.62 & \textbf{66.34 $\pm$ 1.85} & 106.7 $\pm$ 42.71 & \textbf{72.36 $\pm$ 3.80} & \cellcolor{secondbg}52.81 $\pm$ 0.80 & \cellcolor{bestbg}\textbf{51.37 $\pm$ 0.79} \\
        \hline
        \multirow{4}{*}{PM2.5-RV} & 96 & 83.31 $\pm$ 0.63 & \textbf{81.07 $\pm$ 1.18} & \cellcolor{secondbg}80.60 $\pm$ 4.73 & \cellcolor{bestbg}\textbf{76.56 $\pm$ 1.93} & 105.6 $\pm$ 5.02 & \textbf{87.11 $\pm$ 2.39} & 140.4 $\pm$ 27.32 & \textbf{132.5 $\pm$ 9.88} & \textbf{82.47 $\pm$ 6.03} & 83.02 $\pm$ 0.82 \\
          & 192 & 92.20 $\pm$ 0.71 & \textbf{90.23 $\pm$ 0.58} & \cellcolor{secondbg}87.78 $\pm$ 4.06 & \cellcolor{bestbg}\textbf{83.44 $\pm$ 1.35} & 131.5 $\pm$ 8.30 & \textbf{98.74 $\pm$ 9.20} & 142.6 $\pm$ 26.11 & \textbf{133.5 $\pm$ 14.51} & \textbf{94.80 $\pm$ 5.01} & 97.51 $\pm$ 1.59 \\
          & 336 & \textbf{95.85 $\pm$ 1.57} & 96.27 $\pm$ 1.15 & \cellcolor{secondbg}93.31 $\pm$ 2.49 & \cellcolor{bestbg}\textbf{91.94 $\pm$ 1.95} & 128.1 $\pm$ 6.07 & \textbf{103.8 $\pm$ 6.66} & \textbf{136.3 $\pm$ 11.15} & 136.9 $\pm$ 9.93 & \textbf{100.1 $\pm$ 3.35} & 100.6 $\pm$ 0.74 \\
          & 720 & \cellcolor{bestbg}\textbf{87.58 $\pm$ 0.66} & 87.89 $\pm$ 0.23 & 87.86 $\pm$ 1.28 & \cellcolor{secondbg}\textbf{87.73 $\pm$ 1.56} & 115.2 $\pm$ 16.02 & \textbf{94.83 $\pm$ 1.54} & 128.4 $\pm$ 45.85 & \textbf{110.8 $\pm$ 4.55} & 92.70 $\pm$ 3.41 & \textbf{89.88 $\pm$ 0.59} \\
        \hline
        \multirow{4}{*}{Bike Share} & 96 & 15,889 $\pm$ 189.8 & \cellcolor{bestbg}\textbf{14,815 $\pm$ 78.46} & 16,967 $\pm$ 583.8 & \textbf{16,450 $\pm$ 1,166} & 18,171 $\pm$ 612.9 & \cellcolor{secondbg}\textbf{15,597 $\pm$ 1,344} & 24,641 $\pm$ 3,579 & \textbf{21,516 $\pm$ 2,918} & 18,653 $\pm$ 1,798 & \textbf{18,214 $\pm$ 515.1} \\
          & 192 & 21,640 $\pm$ 188.7 & \cellcolor{secondbg}\textbf{20,864 $\pm$ 139.4} & 22,654 $\pm$ 455.7 & \textbf{22,035 $\pm$ 731.2} & 25,283 $\pm$ 1,524 & \cellcolor{bestbg}\textbf{19,930 $\pm$ 655.7} & 32,384 $\pm$ 1,689 & \textbf{29,645 $\pm$ 1,023} & \textbf{22,758 $\pm$ 1,585} & 23,480 $\pm$ 732.0 \\
          & 336 & 27,939 $\pm$ 175.9 & \cellcolor{secondbg}\textbf{27,025 $\pm$ 76.15} & 28,267 $\pm$ 354.7 & \textbf{27,661 $\pm$ 424.5} & 28,721 $\pm$ 2,802 & \cellcolor{bestbg}\textbf{23,498 $\pm$ 770.5} & 39,925 $\pm$ 3,912 & \textbf{37,735 $\pm$ 4,623} & \textbf{28,818 $\pm$ 670.0} & 32,018 $\pm$ 1,878 \\
          & 720 & \textbf{37,411 $\pm$ 193.5} & 37,451 $\pm$ 257.1 & \textbf{36,838 $\pm$ 960.6} & 38,760 $\pm$ 373.4 & \cellcolor{secondbg}34,864 $\pm$ 2,454 & \cellcolor{bestbg}\textbf{32,858 $\pm$ 2,245} & 44,139 $\pm$ 1,902 & \textbf{39,982 $\pm$ 2,296} & \textbf{34,947 $\pm$ 510.6} & 35,416 $\pm$ 309.4 \\
        \hline
        \multirow{4}{*}{Appliances} & 96 & 8,095 $\pm$ 194.7 & \cellcolor{secondbg}\textbf{8,084 $\pm$ 96.98} & 9,864 $\pm$ 703.8 & \textbf{8,350 $\pm$ 152.6} & 10,091 $\pm$ 737.9 & \textbf{9,253 $\pm$ 152.4} & 16,935 $\pm$ 1,463 & \textbf{9,941 $\pm$ 287.8} & 8,921 $\pm$ 830.9 & \cellcolor{bestbg}\textbf{7,950 $\pm$ 321.7} \\
          & 192 & \cellcolor{secondbg}\textbf{8,123 $\pm$ 95.14} & 8,129 $\pm$ 38.71 & 10,121 $\pm$ 918.2 & \textbf{8,468 $\pm$ 187.6} & 9,436 $\pm$ 176.2 & \textbf{9,027 $\pm$ 178.2} & 13,646 $\pm$ 2,756 & \textbf{10,650 $\pm$ 449.0} & 9,172 $\pm$ 488.4 & \cellcolor{bestbg}\textbf{7,739 $\pm$ 332.7} \\
          & 336 & 7,847 $\pm$ 88.18 & \cellcolor{secondbg}\textbf{7,761 $\pm$ 36.02} & 9,843 $\pm$ 1,338 & \textbf{8,044 $\pm$ 155.7} & 9,147 $\pm$ 520.2 & \textbf{8,559 $\pm$ 93.00} & 11,739 $\pm$ 849.4 & \textbf{10,198 $\pm$ 434.5} & 8,329 $\pm$ 365.6 & \cellcolor{bestbg}\textbf{7,681 $\pm$ 207.4} \\
          & 720 & \cellcolor{secondbg}\textbf{8,046 $\pm$ 38.00} & 8,048 $\pm$ 63.82 & 8,861 $\pm$ 654.1 & \textbf{8,280 $\pm$ 87.46} & 10,063 $\pm$ 779.3 & \textbf{9,080 $\pm$ 118.1} & 13,574 $\pm$ 3,183 & \textbf{9,706 $\pm$ 124.9} & 8,261 $\pm$ 225.2 & \cellcolor{bestbg}\textbf{7,870 $\pm$ 175.5} \\
        \hline
        \textbf{\sys~Win} & & \multicolumn{2}{c|}{\textbf{12}} & \multicolumn{2}{c|}{\textbf{15}} & \multicolumn{2}{c|}{\textbf{20}} & \multicolumn{2}{c|}{\textbf{19}} & \multicolumn{2}{c|}{\textbf{14}} \\
        \hline
        \textbf{Impr.\ \% (mean)} & & \multicolumn{2}{c|}{\textbf{+2.4}} & \multicolumn{2}{c|}{\textbf{+4.1}} & \multicolumn{2}{c|}{\textbf{+32.6}} & \multicolumn{2}{c|}{\textbf{+30.9}} & \multicolumn{2}{c|}{\textbf{+20.0}} \\
        \hline
    \end{tabular}
    }
\end{table*}
\endgroup

\begingroup
\renewcommand{\arraystretch}{1.25}
\setlength\extrarowheight{1pt}
\begin{table*}[t]
    \caption{Long-term forecasting results under \textbf{Sample} normalization. MSE is computed on the original (de-normalized) scale and reported as MEAN $\pm$ SD over 5 runs. The look-back length is 96 for all settings. The better of each baseline and its \sys-enhanced version is in \textbf{bold}; the best result in each row is marked with a \bestbox{dark orange} background and the second best with a \secondbox{light orange} background. \sys~Win: number of the 20 settings (5 datasets $\times$ 4 horizons) in which \sys-$<$baseline$>$ has a lower mean MSE than $<$baseline$>$. Impr.: relative MSE reduction of \sys-$<$baseline$>$ over $<$baseline$>$, averaged over the 20 settings.}
    \label{tab:main_sample}
    \centering
    \resizebox{\textwidth}{!}{%
    \begin{tabular}{|l|c|cc|cc|cc|cc|cc|}
        \hline
        Dataset & $L_y$ & \shortstack{ iTransformer \\ (\citeyear{Liu2024iTransformerIT}) } & \shortstack{ \sys-\\iTransformer } & \shortstack{ TimesNet \\ (\citeyear{TimesNet}) } & \shortstack{ \sys-\\TimesNet } & \shortstack{ FEDformer \\ (\citeyear{Fedformer}) } & \shortstack{ \sys-\\FEDformer } & \shortstack{ Autoformer \\ (\citeyear{autoformer}) } & \shortstack{ \sys-\\Autoformer } & \shortstack{ Informer \\ (\citeyear{informer}) } & \shortstack{ \sys-\\Informer } \\
        \hline
        \multirow{4}{*}{Weather} & 96 & 447.0 $\pm$ 47.15 & \textbf{178.3 $\pm$ 9.00} & 418.6 $\pm$ 150.0 & \cellcolor{secondbg}\textbf{158.5 $\pm$ 10.52} & 1,208,418 $\pm$ 1,398,675 & \textbf{181.7 $\pm$ 6.57} & 1,933,907 $\pm$ 1,800,987 & \textbf{210.1 $\pm$ 24.66} & 4,402 $\pm$ 3,262 & \cellcolor{bestbg}\textbf{155.7 $\pm$ 6.54} \\
          & 192 & 637.6 $\pm$ 71.19 & \textbf{231.6 $\pm$ 4.39} & 410.6 $\pm$ 120.8 & \cellcolor{secondbg}\textbf{226.1 $\pm$ 6.49} & 10,628,690 $\pm$ 15,360,229 & \textbf{236.8 $\pm$ 9.04} & 151,937 $\pm$ 325,720 & \textbf{285.4 $\pm$ 31.47} & 7,309 $\pm$ 2,449 & \cellcolor{bestbg}\textbf{220.8 $\pm$ 15.75} \\
          & 336 & 749.2 $\pm$ 64.09 & \textbf{254.8 $\pm$ 8.98} & 456.7 $\pm$ 167.3 & \cellcolor{secondbg}\textbf{241.8 $\pm$ 6.77} & 380,781 $\pm$ 420,438 & \textbf{267.3 $\pm$ 16.56} & 157,994 $\pm$ 251,704 & \textbf{270.1 $\pm$ 12.01} & 3,729 $\pm$ 1,411 & \cellcolor{bestbg}\textbf{240.1 $\pm$ 5.54} \\
          & 720 & 748.3 $\pm$ 46.15 & \textbf{330.0 $\pm$ 35.88} & 963.3 $\pm$ 360.3 & \textbf{291.8 $\pm$ 17.00} & 798,364 $\pm$ 652,673 & \cellcolor{bestbg}\textbf{268.3 $\pm$ 18.87} & 29,367 $\pm$ 28,419 & \textbf{296.1 $\pm$ 33.97} & 3,637 $\pm$ 1,223 & \cellcolor{secondbg}\textbf{290.7 $\pm$ 16.44} \\
        \hline
        \multirow{4}{*}{PM2.5-LA} & 96 & 36.08 $\pm$ 0.96 & \cellcolor{bestbg}\textbf{35.20 $\pm$ 0.96} & 40.18 $\pm$ 4.20 & \cellcolor{secondbg}\textbf{35.43 $\pm$ 1.30} & 60.91 $\pm$ 9.88 & \textbf{42.11 $\pm$ 2.02} & \textbf{57.60 $\pm$ 6.91} & 58.89 $\pm$ 3.52 & 41.30 $\pm$ 2.24 & \textbf{37.90 $\pm$ 1.28} \\
          & 192 & 40.93 $\pm$ 0.48 & \textbf{40.81 $\pm$ 0.46} & 42.62 $\pm$ 3.67 & \cellcolor{bestbg}\textbf{39.25 $\pm$ 1.73} & 64.07 $\pm$ 2.94 & \textbf{43.77 $\pm$ 1.81} & \textbf{61.37 $\pm$ 14.40} & 65.62 $\pm$ 8.96 & 42.06 $\pm$ 1.10 & \cellcolor{secondbg}\textbf{39.62 $\pm$ 0.90} \\
          & 336 & \textbf{48.89 $\pm$ 0.38} & 49.82 $\pm$ 0.69 & \cellcolor{secondbg}46.42 $\pm$ 0.96 & \cellcolor{bestbg}\textbf{44.67 $\pm$ 2.05} & 74.34 $\pm$ 9.01 & \textbf{52.00 $\pm$ 2.87} & \textbf{70.79 $\pm$ 23.11} & 70.81 $\pm$ 3.65 & 52.84 $\pm$ 2.26 & \textbf{48.27 $\pm$ 3.12} \\
          & 720 & \textbf{63.62 $\pm$ 1.45} & 63.88 $\pm$ 1.15 & \cellcolor{secondbg}59.23 $\pm$ 0.75 & \cellcolor{bestbg}\textbf{58.73 $\pm$ 1.94} & 80.34 $\pm$ 13.98 & \textbf{64.62 $\pm$ 1.35} & 142.7 $\pm$ 82.92 & \textbf{72.23 $\pm$ 2.72} & \textbf{62.88 $\pm$ 2.04} & 63.56 $\pm$ 2.81 \\
        \hline
        \multirow{4}{*}{PM2.5-RV} & 96 & 78.83 $\pm$ 0.70 & \cellcolor{secondbg}\textbf{78.43 $\pm$ 0.42} & 78.84 $\pm$ 4.16 & \cellcolor{bestbg}\textbf{73.83 $\pm$ 1.74} & 105.0 $\pm$ 10.35 & \textbf{87.67 $\pm$ 2.04} & \textbf{125.5 $\pm$ 11.71} & 130.1 $\pm$ 13.80 & 86.27 $\pm$ 6.23 & \textbf{81.69 $\pm$ 3.16} \\
          & 192 & 90.67 $\pm$ 1.47 & \cellcolor{secondbg}\textbf{89.73 $\pm$ 0.52} & 99.05 $\pm$ 6.72 & \cellcolor{bestbg}\textbf{84.37 $\pm$ 4.47} & 146.2 $\pm$ 3.44 & \textbf{96.72 $\pm$ 4.31} & \textbf{158.9 $\pm$ 35.83} & 163.9 $\pm$ 38.35 & 101.8 $\pm$ 5.41 & \textbf{91.40 $\pm$ 1.67} \\
          & 336 & \cellcolor{secondbg}\textbf{94.49 $\pm$ 1.71} & 95.08 $\pm$ 1.85 & 98.14 $\pm$ 3.71 & \cellcolor{bestbg}\textbf{92.40 $\pm$ 1.92} & 129.9 $\pm$ 10.81 & \textbf{109.9 $\pm$ 3.75} & 166.5 $\pm$ 39.61 & \textbf{137.6 $\pm$ 10.42} & \textbf{102.6 $\pm$ 14.28} & 102.7 $\pm$ 2.36 \\
          & 720 & 87.28 $\pm$ 0.80 & \textbf{86.70 $\pm$ 0.84} & \cellcolor{bestbg}\textbf{85.81 $\pm$ 1.57} & \cellcolor{secondbg}86.12 $\pm$ 1.11 & 104.5 $\pm$ 9.92 & \textbf{95.63 $\pm$ 2.94} & 122.7 $\pm$ 15.69 & \textbf{112.5 $\pm$ 1.01} & \textbf{94.57 $\pm$ 5.98} & 94.62 $\pm$ 4.06 \\
        \hline
        \multirow{4}{*}{Bike Share} & 96 & 15,134 $\pm$ 148.1 & \cellcolor{bestbg}\textbf{14,258 $\pm$ 39.33} & 15,418 $\pm$ 861.6 & \cellcolor{secondbg}\textbf{14,886 $\pm$ 1,192} & 27,592 $\pm$ 2,130 & \textbf{14,944 $\pm$ 675.7} & 25,573 $\pm$ 3,915 & \textbf{24,799 $\pm$ 4,503} & 16,080 $\pm$ 446.5 & \textbf{15,038 $\pm$ 871.0} \\
          & 192 & 21,140 $\pm$ 109.6 & \cellcolor{bestbg}\textbf{20,356 $\pm$ 177.1} & \cellcolor{secondbg}\textbf{20,500 $\pm$ 527.8} & 20,508 $\pm$ 713.8 & 37,106 $\pm$ 3,519 & \textbf{21,570 $\pm$ 226.8} & \textbf{33,966 $\pm$ 2,009} & 34,204 $\pm$ 2,555 & \textbf{22,524 $\pm$ 578.7} & 22,916 $\pm$ 969.5 \\
          & 336 & \cellcolor{secondbg}27,477 $\pm$ 214.3 & \cellcolor{bestbg}\textbf{26,907 $\pm$ 369.9} & 27,702 $\pm$ 622.4 & \textbf{27,672 $\pm$ 273.7} & 43,632 $\pm$ 3,068 & \textbf{27,528 $\pm$ 715.9} & 48,489 $\pm$ 3,288 & \textbf{41,074 $\pm$ 1,975} & \textbf{29,075 $\pm$ 675.6} & 33,134 $\pm$ 1,690 \\
          & 720 & \cellcolor{secondbg}\textbf{36,714 $\pm$ 288.4} & 37,845 $\pm$ 447.8 & \cellcolor{bestbg}\textbf{35,968 $\pm$ 824.9} & 38,228 $\pm$ 1,150 & 53,511 $\pm$ 2,412 & \textbf{39,902 $\pm$ 934.9} & 51,277 $\pm$ 4,942 & \textbf{47,750 $\pm$ 2,465} & 37,201 $\pm$ 1,044 & \textbf{36,790 $\pm$ 237.4} \\
        \hline
        \multirow{4}{*}{Appliances} & 96 & 17,088 $\pm$ 2,518 & \cellcolor{bestbg}\textbf{10,267 $\pm$ 727.9} & 16,593 $\pm$ 1,273 & \cellcolor{secondbg}\textbf{11,356 $\pm$ 1,938} & 17,445 $\pm$ 2,985 & \textbf{13,584 $\pm$ 2,197} & 76,717 $\pm$ 96,329 & \textbf{12,599 $\pm$ 1,494} & 38,936 $\pm$ 15,594 & \textbf{11,884 $\pm$ 2,970} \\
          & 192 & 12,600 $\pm$ 906.8 & \cellcolor{bestbg}\textbf{11,150 $\pm$ 1,708} & 22,151 $\pm$ 5,401 & \textbf{12,079 $\pm$ 883.9} & 26,191 $\pm$ 9,761 & \textbf{12,355 $\pm$ 2,335} & 97,072 $\pm$ 27,511 & \textbf{13,580 $\pm$ 1,081} & 52,661 $\pm$ 22,981 & \cellcolor{secondbg}\textbf{12,043 $\pm$ 3,531} \\
          & 336 & 14,408 $\pm$ 1,905 & \cellcolor{bestbg}\textbf{10,362 $\pm$ 1,076} & 22,152 $\pm$ 7,005 & \cellcolor{secondbg}\textbf{10,624 $\pm$ 1,113} & 46,912 $\pm$ 43,302 & \textbf{12,294 $\pm$ 1,322} & 80,913 $\pm$ 63,971 & \textbf{12,261 $\pm$ 1,005} & 47,692 $\pm$ 24,051 & \textbf{15,531 $\pm$ 10,359} \\
          & 720 & 16,877 $\pm$ 2,276 & \textbf{11,478 $\pm$ 1,603} & 18,186 $\pm$ 4,418 & \cellcolor{bestbg}\textbf{11,308 $\pm$ 393.4} & 42,264 $\pm$ 20,699 & \textbf{12,637 $\pm$ 2,454} & 53,329 $\pm$ 58,727 & \cellcolor{secondbg}\textbf{11,398 $\pm$ 1,104} & 31,365 $\pm$ 7,205 & \textbf{12,709 $\pm$ 2,409} \\
        \hline
        \textbf{\sys~Win} & & \multicolumn{2}{c|}{\textbf{16}} & \multicolumn{2}{c|}{\textbf{17}} & \multicolumn{2}{c|}{\textbf{20}} & \multicolumn{2}{c|}{\textbf{14}} & \multicolumn{2}{c|}{\textbf{15}} \\
        \hline
        \textbf{Impr.\ \% (mean)} & & \multicolumn{2}{c|}{\textbf{+18.4}} & \multicolumn{2}{c|}{\textbf{+21.9}} & \multicolumn{2}{c|}{\textbf{+47.8}} & \multicolumn{2}{c|}{\textbf{+40.8}} & \multicolumn{2}{c|}{\textbf{+34.1}} \\
        \hline
    \end{tabular}
    }
\end{table*}
\endgroup

\subsection{Reduction Percentage}
\label{app:res_reduction}
Table \ref{app:reduction} presents the percentage reduction in the MSE mean achieved by the CATF-\textless baseline\textgreater{} compared to the \textless baseline\textgreater{}. A lower MSE indicates better performance.
To assess whether the improvements are statistically significant, we apply a two-sided Welch's $t$-test to the 5 runs of each \sys-enhanced model and its standalone counterpart in every setting. Of the 200 comparisons, \syss is significantly better in 91 ($p<0.05$) and significantly worse in 9, and the difference is not significant in the remaining 100. The share of significant improvements is highest for \sys-FEDformer (29 of 40) and lowest for \sys-Autoformer (6 of 40), whose baseline has a very large run-to-run variance, so that even large mean improvements are often not significant with 5 runs.
\begin{table*}[h]
\centering
\caption{MSE reduction (\%) of \sys-$<$baseline$>$ over $<$baseline$>$, computed on the mean MSE over 5 runs as (baseline $-$ \sys)/baseline. Positive values mean \sys{} is better. G = global normalization, S = sample normalization. Significance is tested with a two-sided Welch's $t$-test over the 5 runs of each model: \textbf{bold} = \sys{} is significantly better ($p<0.05$), \underline{underlined} = \sys{} is significantly worse ($p<0.05$).}
\label{app:reduction}
\small
\setlength{\tabcolsep}{5pt}
\renewcommand{\arraystretch}{0.9}
\begin{tabular}{llrrrrrrrrrr}
\toprule
\textbf{Dataset} & $L_y$ & \multicolumn{2}{c}{\textbf{iTransformer}} & \multicolumn{2}{c}{\textbf{TimesNet}} & \multicolumn{2}{c}{\textbf{FEDformer}} & \multicolumn{2}{c}{\textbf{Autoformer}} & \multicolumn{2}{c}{\textbf{Informer}} \\
 & & G & S & G & S & G & S & G & S & G & S \\
\midrule
\multirow{4}{*}{Weather} & 96 & \textbf{+4.8} & \textbf{+60.1} & \textbf{+12.1} & \textbf{+62.1} & \textbf{+98.8} & +100.0 & +98.9 & +100.0 & \textbf{+74.5} & \textbf{+96.5} \\
 & 192 & \textbf{+9.1} & \textbf{+63.7} & \textbf{+13.4} & \textbf{+44.9} & \textbf{+98.1} & +100.0 & +99.8 & +99.8 & +93.6 & \textbf{+97.0} \\
 & 336 & \textbf{+8.7} & \textbf{+66.0} & \textbf{+8.1} & \textbf{+47.1} & \textbf{+98.9} & +99.9 & +99.6 & +99.8 & +99.4 & \textbf{+93.6} \\
 & 720 & \textbf{+12.7} & \textbf{+55.9} & \textbf{+10.2} & \textbf{+69.7} & \textbf{+98.4} & +100.0 & +99.9 & +99.0 & +87.8 & \textbf{+92.0} \\
\midrule
\multirow{4}{*}{PM2.5-LA} & 96 & +0.1 & +2.4 & \underline{-19.0} & +11.8 & \textbf{+26.8} & \textbf{+30.9} & +17.5 & -2.3 & +0.7 & \textbf{+8.2} \\
 & 192 & \underline{-1.3} & +0.3 & -5.9 & +7.9 & \textbf{+31.0} & \textbf{+31.7} & +2.2 & -6.9 & \textbf{+8.8} & \textbf{+5.8} \\
 & 336 & -1.3 & \underline{-1.9} & -5.7 & +3.8 & \textbf{+20.0} & \textbf{+30.1} & +1.8 & -0.0 & \textbf{+7.1} & \textbf{+8.6} \\
 & 720 & \underline{-2.6} & -0.4 & -2.3 & +0.8 & \textbf{+12.5} & +19.6 & +32.2 & +49.4 & \textbf{+2.7} & -1.1 \\
\midrule
\multirow{4}{*}{PM2.5-RV} & 96 & \textbf{+2.7} & +0.5 & +5.0 & +6.4 & \textbf{+17.5} & \textbf{+16.5} & +5.6 & -3.6 & -0.7 & +5.3 \\
 & 192 & \textbf{+2.1} & +1.0 & +4.9 & \textbf{+14.8} & \textbf{+24.9} & \textbf{+33.8} & +6.4 & -3.1 & -2.9 & \textbf{+10.2} \\
 & 336 & -0.4 & -0.6 & +1.5 & \textbf{+5.8} & \textbf{+19.0} & \textbf{+15.4} & -0.5 & +17.4 & -0.4 & -0.0 \\
 & 720 & -0.4 & +0.7 & +0.1 & -0.4 & \textbf{+17.7} & +8.5 & +13.7 & +8.4 & +3.0 & -0.1 \\
\midrule
\multirow{4}{*}{Bike Share} & 96 & \textbf{+6.8} & \textbf{+5.8} & +3.0 & +3.5 & \textbf{+14.2} & \textbf{+45.8} & +12.7 & +3.0 & +2.4 & +6.5 \\
 & 192 & \textbf{+3.6} & \textbf{+3.7} & +2.7 & -0.0 & \textbf{+21.2} & \textbf{+41.9} & \textbf{+8.5} & -0.7 & -3.2 & -1.7 \\
 & 336 & \textbf{+3.3} & \textbf{+2.1} & \textbf{+2.1} & +0.1 & \textbf{+18.2} & \textbf{+36.9} & +5.5 & \textbf{+15.3} & \underline{-11.1} & \underline{-14.0} \\
 & 720 & -0.1 & \underline{-3.1} & \underline{-5.2} & \underline{-6.3} & +5.8 & \textbf{+25.4} & \textbf{+9.4} & +6.9 & -1.3 & +1.1 \\
\midrule
\multirow{4}{*}{Appliances} & 96 & +0.1 & \textbf{+39.9} & \textbf{+15.3} & \textbf{+31.6} & +8.3 & +22.1 & \textbf{+41.3} & +83.6 & +10.9 & \textbf{+69.5} \\
 & 192 & -0.1 & +11.5 & \textbf{+16.3} & \textbf{+45.5} & \textbf{+4.3} & \textbf{+52.8} & +22.0 & \textbf{+86.0} & \textbf{+15.6} & \textbf{+77.1} \\
 & 336 & +1.1 & \textbf{+28.1} & \textbf{+18.3} & \textbf{+52.0} & +6.4 & +73.8 & \textbf{+13.1} & +84.8 & \textbf{+7.8} & \textbf{+67.4} \\
 & 720 & -0.0 & \textbf{+32.0} & +6.6 & \textbf{+37.8} & \textbf{+9.8} & \textbf{+70.1} & +28.5 & +78.6 & \textbf{+4.7} & \textbf{+59.5} \\
\midrule
\multicolumn{2}{l}{\#Wins} & 12/20 & 16/20 & 15/20 & 17/20 & 20/20 & 20/20 & 19/20 & 14/20 & 14/20 & 15/20 \\
\multicolumn{2}{l}{\#Sig.\ better} & 9/20 & 10/20 & 8/20 & 10/20 & 17/20 & 12/20 & 4/20 & 2/20 & 7/20 & 12/20 \\
\multicolumn{2}{l}{\#Sig.\ worse} & 2/20 & 2/20 & 2/20 & 1/20 & 0/20 & 0/20 & 0/20 & 0/20 & 1/20 & 1/20 \\
\bottomrule
\end{tabular}
\end{table*}


\subsection{MAE Results}
\label{app:res_mae}
Tables~\ref{app:mae_global} and~\ref{app:mae_sample} report the MAE of all models under global and sample normalization, in the same format as Tables~\ref{tab:main_global} and~\ref{tab:main_sample}. The MAE results are consistent with the MSE results: \sys-enhanced models have a lower mean MAE than their standalone counterparts in 162 of the 200 comparisons (84 under global and 78 under sample normalization).

\begingroup
\renewcommand{\arraystretch}{1.25}
\setlength\extrarowheight{1pt}
\begin{table*}[h]
    \caption{Long-term forecasting results under \textbf{Global} normalization. MAE is computed on the original (de-normalized) scale and reported as MEAN $\pm$ SD over 5 runs (TimeMoE: single zero-shot run). The look-back length is 96 for all settings. The better of each baseline and its \sys-enhanced version is in \textbf{bold}; the best result in each row is marked with a \bestbox{dark orange} background and the second best with a \secondbox{light orange} background. \sys~Win: number of the 20 settings (5 datasets $\times$ 4 horizons) in which \sys-\textless baseline\textgreater{} has a lower mean MAE than \textless baseline\textgreater{}. Impr.: relative MAE reduction of \sys-\textless baseline\textgreater{} over \textless baseline\textgreater{}, averaged over the 20 settings.}
    \label{app:mae_global}
    \centering
    \resizebox{\textwidth}{!}{%
    \begin{tabular}{|l|c|cc|cc|cc|cc|cc|c|}
        \hline
        Dataset & $L_y$ & \shortstack{ \cite{Liu2024iTransformerIT} \\ iTransformer } & \shortstack{ \sys-\\iTransformer } & \shortstack{ \cite{TimesNet} \\ TimesNet } & \shortstack{ \sys-\\TimesNet } & \shortstack{ \cite{Fedformer} \\ FEDformer } & \shortstack{ \sys-\\FEDformer } & \shortstack{ \cite{autoformer} \\ Autoformer } & \shortstack{ \sys-\\Autoformer } & \shortstack{ \cite{informer} \\ Informer } & \shortstack{ \sys-\\Informer } & \shortstack{TimeMoE\\(zero-shot)} \\
        \hline
        \multirow{4}{*}{Weather} & 96 & 10.31 $\pm$ 0.07 & \cellcolor{bestbg}\textbf{10.05 $\pm$ 0.16} & 11.41 $\pm$ 0.25 & \cellcolor{secondbg}\textbf{10.23 $\pm$ 0.49} & 133.1 $\pm$ 42.16 & \textbf{14.26 $\pm$ 1.53} & 154.9 $\pm$ 88.36 & \textbf{15.97 $\pm$ 2.15} & 32.48 $\pm$ 8.31 & \textbf{16.52 $\pm$ 0.83} & 21.61 \\
         & 192 & 11.64 $\pm$ 0.07 & \cellcolor{bestbg}\textbf{11.04 $\pm$ 0.17} & 12.16 $\pm$ 0.27 & \cellcolor{secondbg}\textbf{11.30 $\pm$ 0.17} & 133.1 $\pm$ 64.82 & \textbf{18.50 $\pm$ 0.29} & 241.3 $\pm$ 233.3 & \textbf{14.93 $\pm$ 1.06} & 32.27 $\pm$ 12.88 & \textbf{15.93 $\pm$ 1.12} & 24.60 \\
         & 336 & 12.03 $\pm$ 0.05 & \cellcolor{bestbg}\textbf{11.43 $\pm$ 0.11} & 12.25 $\pm$ 0.08 & \cellcolor{secondbg}\textbf{11.72 $\pm$ 0.10} & 195.6 $\pm$ 75.97 & \textbf{18.37 $\pm$ 1.31} & 150.0 $\pm$ 121.0 & \textbf{12.79 $\pm$ 0.34} & 46.55 $\pm$ 63.01 & \textbf{16.70 $\pm$ 0.97} & 26.44 \\
         & 720 & 12.71 $\pm$ 0.08 & \cellcolor{bestbg}\textbf{11.74 $\pm$ 0.20} & 12.64 $\pm$ 0.17 & \cellcolor{secondbg}\textbf{11.92 $\pm$ 0.11} & 174.9 $\pm$ 58.48 & \textbf{20.08 $\pm$ 1.08} & 274.6 $\pm$ 233.7 & \textbf{12.26 $\pm$ 0.26} & 34.14 $\pm$ 6.19 & \textbf{18.16 $\pm$ 1.20} & 28.63 \\
        \hline
        \multirow{4}{*}{PM2.5-LA} & 96 & \cellcolor{secondbg}\textbf{3.91 $\pm$ 0.01} & 3.93 $\pm$ 0.00 & \cellcolor{bestbg}\textbf{3.91 $\pm$ 0.06} & 4.13 $\pm$ 0.19 & 5.77 $\pm$ 0.24 & \textbf{4.72 $\pm$ 0.07} & 6.80 $\pm$ 1.00 & \textbf{6.27 $\pm$ 0.26} & 5.07 $\pm$ 0.26 & \textbf{4.97 $\pm$ 0.22} & 4.41 \\
         & 192 & \cellcolor{secondbg}\textbf{4.20 $\pm$ 0.02} & 4.24 $\pm$ 0.02 & \cellcolor{bestbg}\textbf{4.18 $\pm$ 0.09} & 4.24 $\pm$ 0.09 & 6.19 $\pm$ 0.09 & \textbf{4.88 $\pm$ 0.05} & 6.02 $\pm$ 0.39 & \textbf{5.99 $\pm$ 0.22} & 5.21 $\pm$ 0.14 & \textbf{4.96 $\pm$ 0.05} & 4.92 \\
         & 336 & \textbf{4.71 $\pm$ 0.01} & 4.74 $\pm$ 0.04 & \cellcolor{bestbg}\textbf{4.55 $\pm$ 0.11} & \cellcolor{secondbg}4.66 $\pm$ 0.09 & 6.18 $\pm$ 0.16 & \textbf{5.34 $\pm$ 0.17} & 6.22 $\pm$ 0.28 & \textbf{6.20 $\pm$ 0.10} & 5.29 $\pm$ 0.12 & \textbf{5.12 $\pm$ 0.06} & 5.42 \\
         & 720 & \textbf{5.76 $\pm$ 0.02} & 5.87 $\pm$ 0.03 & \cellcolor{bestbg}\textbf{5.58 $\pm$ 0.03} & 5.66 $\pm$ 0.14 & 6.54 $\pm$ 0.26 & \textbf{6.04 $\pm$ 0.10} & 8.03 $\pm$ 1.81 & \textbf{6.42 $\pm$ 0.18} & 5.66 $\pm$ 0.06 & \cellcolor{secondbg}\textbf{5.62 $\pm$ 0.07} & 5.93 \\
        \hline
        \multirow{4}{*}{PM2.5-RV} & 96 & 6.25 $\pm$ 0.04 & \textbf{6.16 $\pm$ 0.06} & 6.08 $\pm$ 0.17 & \cellcolor{secondbg}\textbf{6.00 $\pm$ 0.03} & 7.53 $\pm$ 0.23 & \textbf{6.62 $\pm$ 0.05} & 8.36 $\pm$ 0.74 & \textbf{8.13 $\pm$ 0.24} & 6.10 $\pm$ 0.19 & \textbf{6.01 $\pm$ 0.07} & \cellcolor{bestbg}5.97 \\
         & 192 & 6.51 $\pm$ 0.05 & \textbf{6.44 $\pm$ 0.03} & 6.28 $\pm$ 0.14 & \cellcolor{bestbg}\textbf{6.20 $\pm$ 0.06} & 8.48 $\pm$ 0.28 & \textbf{7.11 $\pm$ 0.27} & 8.31 $\pm$ 0.57 & \textbf{8.29 $\pm$ 0.38} & \textbf{6.47 $\pm$ 0.16} & 6.48 $\pm$ 0.02 & \cellcolor{secondbg}6.26 \\
         & 336 & \textbf{6.61 $\pm$ 0.05} & 6.63 $\pm$ 0.02 & 6.57 $\pm$ 0.04 & \cellcolor{secondbg}\textbf{6.42 $\pm$ 0.06} & 8.33 $\pm$ 0.34 & \textbf{7.22 $\pm$ 0.32} & 8.41 $\pm$ 0.11 & \textbf{8.35 $\pm$ 0.24} & 6.62 $\pm$ 0.07 & \textbf{6.55 $\pm$ 0.04} & \cellcolor{bestbg}6.40 \\
         & 720 & \textbf{6.49 $\pm$ 0.04} & 6.50 $\pm$ 0.03 & \textbf{6.43 $\pm$ 0.03} & 6.47 $\pm$ 0.09 & 7.80 $\pm$ 0.50 & \textbf{6.99 $\pm$ 0.14} & 8.32 $\pm$ 1.72 & \textbf{7.78 $\pm$ 0.22} & 6.54 $\pm$ 0.26 & \cellcolor{bestbg}\textbf{6.29 $\pm$ 0.04} & \cellcolor{secondbg}6.36 \\
        \hline
        \multirow{4}{*}{Bike Share} & 96 & 83.38 $\pm$ 0.60 & \cellcolor{bestbg}\textbf{78.50 $\pm$ 0.30} & 81.95 $\pm$ 1.47 & \textbf{80.21 $\pm$ 3.90} & 93.44 $\pm$ 2.26 & \textbf{84.74 $\pm$ 2.74} & 116.0 $\pm$ 11.46 & \textbf{105.4 $\pm$ 9.40} & 91.06 $\pm$ 5.20 & \textbf{88.51 $\pm$ 1.52} & \cellcolor{secondbg}78.52 \\
         & 192 & 97.71 $\pm$ 0.59 & \cellcolor{secondbg}\textbf{93.95 $\pm$ 0.39} & 96.91 $\pm$ 1.97 & \textbf{94.42 $\pm$ 2.02} & 107.0 $\pm$ 3.25 & \textbf{95.92 $\pm$ 1.13} & 135.1 $\pm$ 5.61 & \textbf{127.2 $\pm$ 2.66} & \textbf{100.9 $\pm$ 3.61} & 102.9 $\pm$ 2.54 & \cellcolor{bestbg}87.81 \\
         & 336 & 112.8 $\pm$ 0.36 & \textbf{108.7 $\pm$ 0.22} & 108.6 $\pm$ 1.48 & \textbf{106.6 $\pm$ 1.29} & 114.1 $\pm$ 4.96 & \cellcolor{secondbg}\textbf{106.1 $\pm$ 3.09} & 154.9 $\pm$ 7.83 & \textbf{146.9 $\pm$ 14.25} & \textbf{119.0 $\pm$ 2.82} & 138.5 $\pm$ 10.43 & \cellcolor{bestbg}97.52 \\
         & 720 & 134.9 $\pm$ 0.39 & \textbf{133.1 $\pm$ 0.65} & \cellcolor{secondbg}\textbf{126.8 $\pm$ 2.72} & 129.9 $\pm$ 1.29 & 129.1 $\pm$ 4.01 & \textbf{127.0 $\pm$ 3.49} & 167.4 $\pm$ 3.80 & \textbf{156.7 $\pm$ 5.79} & \textbf{138.1 $\pm$ 1.73} & 150.3 $\pm$ 0.64 & \cellcolor{bestbg}118.8 \\
        \hline
        \multirow{4}{*}{Appliances} & 96 & \cellcolor{secondbg}47.50 $\pm$ 1.34 & \cellcolor{bestbg}\textbf{46.53 $\pm$ 1.02} & 55.55 $\pm$ 2.75 & \textbf{47.56 $\pm$ 1.28} & 64.61 $\pm$ 4.50 & \textbf{60.66 $\pm$ 1.47} & 92.33 $\pm$ 5.10 & \textbf{61.17 $\pm$ 2.73} & 53.78 $\pm$ 3.13 & \textbf{48.72 $\pm$ 3.82} & 51.78 \\
         & 192 & 47.72 $\pm$ 0.88 & \cellcolor{bestbg}\textbf{46.13 $\pm$ 0.76} & 56.96 $\pm$ 3.29 & \textbf{48.88 $\pm$ 2.91} & 58.98 $\pm$ 0.92 & \textbf{56.28 $\pm$ 1.67} & 79.03 $\pm$ 11.01 & \textbf{61.86 $\pm$ 2.15} & 55.81 $\pm$ 3.22 & \cellcolor{secondbg}\textbf{47.32 $\pm$ 1.67} & 51.50 \\
         & 336 & \cellcolor{secondbg}46.35 $\pm$ 0.66 & \cellcolor{bestbg}\textbf{45.16 $\pm$ 0.60} & 56.96 $\pm$ 4.05 & \textbf{48.48 $\pm$ 2.11} & 60.44 $\pm$ 4.10 & \textbf{54.25 $\pm$ 0.91} & 69.15 $\pm$ 4.92 & \textbf{63.83 $\pm$ 5.33} & 52.47 $\pm$ 1.49 & \textbf{48.81 $\pm$ 2.05} & 52.54 \\
         & 720 & \cellcolor{secondbg}46.41 $\pm$ 0.42 & \cellcolor{bestbg}\textbf{45.68 $\pm$ 0.28} & 53.16 $\pm$ 3.00 & \textbf{47.81 $\pm$ 1.60} & 62.84 $\pm$ 6.14 & \textbf{56.70 $\pm$ 0.74} & 85.42 $\pm$ 23.30 & \textbf{61.40 $\pm$ 3.46} & 53.19 $\pm$ 1.30 & \textbf{49.15 $\pm$ 1.73} & 55.05 \\
        \hline
        \textbf{\sys~Win} & & \multicolumn{2}{c|}{\textbf{14}} & \multicolumn{2}{c|}{\textbf{14}} & \multicolumn{2}{c|}{\textbf{20}} & \multicolumn{2}{c|}{\textbf{20}} & \multicolumn{2}{c|}{\textbf{16}} & -- \\
        \hline
        \textbf{Impr.\ \% (mean)} & & \multicolumn{2}{c|}{\textbf{+2.1}} & \multicolumn{2}{c|}{\textbf{+3.9}} & \multicolumn{2}{c|}{\textbf{+26.3}} & \multicolumn{2}{c|}{\textbf{+26.4}} & \multicolumn{2}{c|}{\textbf{+12.1}} & -- \\
        \hline
    \end{tabular}
    }
\end{table*}
\endgroup

\begingroup
\renewcommand{\arraystretch}{1.25}
\setlength\extrarowheight{1pt}
\begin{table*}[h]
    \caption{Long-term forecasting results under \textbf{Sample} normalization. MAE is computed on the original (de-normalized) scale and reported as MEAN $\pm$ SD over 5 runs (TimeMoE: single zero-shot run). The look-back length is 96 for all settings. The better of each baseline and its \sys-enhanced version is in \textbf{bold}; the best result in each row is marked with a \bestbox{dark orange} background and the second best with a \secondbox{light orange} background. \sys~Win: number of the 20 settings (5 datasets $\times$ 4 horizons) in which \sys-\textless baseline\textgreater{} has a lower mean MAE than \textless baseline\textgreater{}. Impr.: relative MAE reduction of \sys-\textless baseline\textgreater{} over \textless baseline\textgreater{}, averaged over the 20 settings.}
    \label{app:mae_sample}
    \centering
    \resizebox{\textwidth}{!}{%
    \begin{tabular}{|l|c|cc|cc|cc|cc|cc|c|}
        \hline
        Dataset & $L_y$ & \shortstack{ \cite{Liu2024iTransformerIT} \\ iTransformer } & \shortstack{ \sys-\\iTransformer } & \shortstack{ \cite{TimesNet} \\ TimesNet } & \shortstack{ \sys-\\TimesNet } & \shortstack{ \cite{Fedformer} \\ FEDformer } & \shortstack{ \sys-\\FEDformer } & \shortstack{ \cite{autoformer} \\ Autoformer } & \shortstack{ \sys-\\Autoformer } & \shortstack{ \cite{informer} \\ Informer } & \shortstack{ \sys-\\Informer } & \shortstack{TimeMoE\\(zero-shot)} \\
        \hline
        \multirow{4}{*}{Weather} & 96 & 14.13 $\pm$ 0.67 & \textbf{9.28 $\pm$ 0.20} & 11.84 $\pm$ 0.57 & \cellcolor{secondbg}\textbf{8.61 $\pm$ 0.25} & 725.8 $\pm$ 476.1 & \textbf{9.45 $\pm$ 0.19} & 983.5 $\pm$ 593.3 & \textbf{10.04 $\pm$ 0.43} & 28.87 $\pm$ 8.79 & \cellcolor{bestbg}\textbf{8.53 $\pm$ 0.20} & 9.05 \\
         & 192 & 17.43 $\pm$ 1.09 & \textbf{10.89 $\pm$ 0.12} & 12.80 $\pm$ 0.83 & \textbf{10.71 $\pm$ 0.10} & 1,691 $\pm$ 2,008 & \textbf{11.22 $\pm$ 0.19} & 185.3 $\pm$ 305.8 & \textbf{12.06 $\pm$ 0.46} & 32.24 $\pm$ 6.16 & \cellcolor{secondbg}\textbf{10.65 $\pm$ 0.25} & \cellcolor{bestbg}10.63 \\
         & 336 & 19.16 $\pm$ 0.73 & \textbf{11.76 $\pm$ 0.17} & 13.89 $\pm$ 1.76 & \cellcolor{secondbg}\textbf{11.44 $\pm$ 0.19} & 380.7 $\pm$ 239.9 & \textbf{12.32 $\pm$ 0.37} & 194.7 $\pm$ 123.2 & \textbf{12.12 $\pm$ 0.22} & 29.58 $\pm$ 7.54 & \textbf{11.47 $\pm$ 0.10} & \cellcolor{bestbg}11.14 \\
         & 720 & 19.89 $\pm$ 0.54 & \textbf{13.34 $\pm$ 0.57} & 17.53 $\pm$ 2.17 & \cellcolor{secondbg}\textbf{12.36 $\pm$ 0.35} & 588.6 $\pm$ 274.4 & \textbf{12.52 $\pm$ 0.49} & 118.6 $\pm$ 83.34 & \textbf{13.18 $\pm$ 0.89} & 28.66 $\pm$ 4.30 & \textbf{12.62 $\pm$ 0.35} & \cellcolor{bestbg}11.55 \\
        \hline
        \multirow{4}{*}{PM2.5-LA} & 96 & 3.93 $\pm$ 0.06 & \cellcolor{secondbg}\textbf{3.87 $\pm$ 0.07} & 4.02 $\pm$ 0.12 & \cellcolor{bestbg}\textbf{3.85 $\pm$ 0.06} & 5.21 $\pm$ 0.46 & \textbf{4.24 $\pm$ 0.12} & \textbf{4.89 $\pm$ 0.24} & 5.07 $\pm$ 0.20 & 4.33 $\pm$ 0.18 & \textbf{4.12 $\pm$ 0.10} & 4.10 \\
         & 192 & \cellcolor{secondbg}\textbf{4.25 $\pm$ 0.04} & 4.26 $\pm$ 0.03 & 4.26 $\pm$ 0.15 & \cellcolor{bestbg}\textbf{4.14 $\pm$ 0.06} & 5.45 $\pm$ 0.10 & \textbf{4.39 $\pm$ 0.12} & \textbf{5.13 $\pm$ 0.48} & 5.37 $\pm$ 0.31 & 4.40 $\pm$ 0.07 & \textbf{4.26 $\pm$ 0.07} & 4.41 \\
         & 336 & \textbf{4.77 $\pm$ 0.03} & 4.83 $\pm$ 0.04 & \cellcolor{secondbg}4.52 $\pm$ 0.03 & \cellcolor{bestbg}\textbf{4.49 $\pm$ 0.10} & 6.00 $\pm$ 0.43 & \textbf{4.94 $\pm$ 0.18} & \textbf{5.79 $\pm$ 1.05} & 5.85 $\pm$ 0.15 & 5.09 $\pm$ 0.11 & \textbf{4.82 $\pm$ 0.19} & 4.98 \\
         & 720 & \textbf{5.78 $\pm$ 0.08} & 5.80 $\pm$ 0.06 & \cellcolor{bestbg}\textbf{5.59 $\pm$ 0.08} & \cellcolor{secondbg}5.62 $\pm$ 0.13 & 6.55 $\pm$ 0.65 & \textbf{5.84 $\pm$ 0.08} & 8.94 $\pm$ 3.09 & \textbf{6.34 $\pm$ 0.15} & \textbf{5.85 $\pm$ 0.11} & 5.86 $\pm$ 0.15 & 6.08 \\
        \hline
        \multirow{4}{*}{PM2.5-RV} & 96 & \textbf{6.08 $\pm$ 0.05} & 6.14 $\pm$ 0.05 & \cellcolor{bestbg}\textbf{6.04 $\pm$ 0.09} & \cellcolor{secondbg}6.05 $\pm$ 0.09 & 7.26 $\pm$ 0.33 & \textbf{6.57 $\pm$ 0.12} & \textbf{7.90 $\pm$ 0.29} & 7.98 $\pm$ 0.54 & 6.25 $\pm$ 0.11 & \textbf{6.11 $\pm$ 0.13} & 6.60 \\
         & 192 & \cellcolor{secondbg}\textbf{6.47 $\pm$ 0.03} & 6.50 $\pm$ 0.06 & 6.59 $\pm$ 0.22 & \cellcolor{bestbg}\textbf{6.26 $\pm$ 0.15} & 8.65 $\pm$ 0.11 & \textbf{6.83 $\pm$ 0.18} & \textbf{8.57 $\pm$ 0.58} & 8.67 $\pm$ 0.65 & 6.72 $\pm$ 0.09 & \textbf{6.52 $\pm$ 0.12} & 6.85 \\
         & 336 & 6.69 $\pm$ 0.08 & \cellcolor{secondbg}\textbf{6.64 $\pm$ 0.06} & 6.77 $\pm$ 0.11 & \cellcolor{bestbg}\textbf{6.52 $\pm$ 0.10} & 8.14 $\pm$ 0.45 & \textbf{7.33 $\pm$ 0.17} & 8.65 $\pm$ 0.51 & \textbf{8.28 $\pm$ 0.38} & 7.07 $\pm$ 0.20 & \textbf{7.05 $\pm$ 0.24} & 7.03 \\
         & 720 & \textbf{6.53 $\pm$ 0.04} & 6.53 $\pm$ 0.06 & \cellcolor{bestbg}\textbf{6.44 $\pm$ 0.04} & \cellcolor{secondbg}6.46 $\pm$ 0.08 & 7.20 $\pm$ 0.42 & \textbf{6.94 $\pm$ 0.19} & 8.16 $\pm$ 0.85 & \textbf{7.64 $\pm$ 0.10} & \textbf{6.90 $\pm$ 0.08} & 6.98 $\pm$ 0.28 & 6.97 \\
        \hline
        \multirow{4}{*}{Bike Share} & 96 & 83.57 $\pm$ 0.47 & \textbf{79.44 $\pm$ 0.16} & 81.15 $\pm$ 2.75 & \cellcolor{bestbg}\textbf{78.20 $\pm$ 4.61} & 116.5 $\pm$ 7.39 & \textbf{81.16 $\pm$ 2.43} & 116.8 $\pm$ 9.46 & \textbf{112.2 $\pm$ 9.07} & 83.96 $\pm$ 1.69 & \textbf{80.78 $\pm$ 2.91} & \cellcolor{secondbg}78.94 \\
         & 192 & 100.0 $\pm$ 0.56 & \textbf{95.97 $\pm$ 0.83} & 95.55 $\pm$ 1.81 & \cellcolor{secondbg}\textbf{93.79 $\pm$ 1.01} & 135.9 $\pm$ 6.82 & \textbf{97.54 $\pm$ 1.40} & \textbf{132.4 $\pm$ 5.62} & 134.8 $\pm$ 7.27 & \textbf{103.3 $\pm$ 2.98} & 103.5 $\pm$ 2.60 & \cellcolor{bestbg}91.42 \\
         & 336 & 115.4 $\pm$ 0.40 & \cellcolor{secondbg}\textbf{112.0 $\pm$ 0.80} & 112.3 $\pm$ 1.14 & \textbf{112.0 $\pm$ 1.42} & 139.2 $\pm$ 4.26 & \textbf{113.6 $\pm$ 1.89} & 161.7 $\pm$ 9.95 & \textbf{143.1 $\pm$ 4.66} & \textbf{122.0 $\pm$ 2.51} & 143.6 $\pm$ 8.16 & \cellcolor{bestbg}106.2 \\
         & 720 & \textbf{136.4 $\pm$ 0.59} & 137.1 $\pm$ 0.79 & \cellcolor{bestbg}\textbf{131.2 $\pm$ 3.55} & \cellcolor{secondbg}132.6 $\pm$ 1.95 & 157.4 $\pm$ 5.82 & \textbf{137.4 $\pm$ 1.74} & 177.8 $\pm$ 8.56 & \textbf{170.8 $\pm$ 3.69} & \textbf{153.9 $\pm$ 2.76} & 156.8 $\pm$ 1.93 & 133.1 \\
        \hline
        \multirow{4}{*}{Appliances} & 96 & 74.04 $\pm$ 5.14 & \cellcolor{secondbg}\textbf{60.15 $\pm$ 2.18} & 70.47 $\pm$ 3.88 & \textbf{62.76 $\pm$ 6.36} & 80.90 $\pm$ 4.74 & \textbf{72.17 $\pm$ 6.41} & 159.6 $\pm$ 116.5 & \textbf{69.60 $\pm$ 5.08} & 94.24 $\pm$ 14.76 & \textbf{63.80 $\pm$ 4.27} & \cellcolor{bestbg}54.71 \\
         & 192 & 67.30 $\pm$ 3.81 & \cellcolor{secondbg}\textbf{62.45 $\pm$ 4.11} & 71.11 $\pm$ 3.60 & \textbf{65.18 $\pm$ 1.96} & 93.68 $\pm$ 8.12 & \textbf{68.63 $\pm$ 9.13} & 160.5 $\pm$ 34.00 & \textbf{76.08 $\pm$ 5.17} & 95.50 $\pm$ 12.24 & \textbf{63.60 $\pm$ 5.92} & \cellcolor{bestbg}54.24 \\
         & 336 & 72.78 $\pm$ 5.10 & \textbf{62.01 $\pm$ 2.70} & 72.69 $\pm$ 6.57 & \cellcolor{secondbg}\textbf{59.66 $\pm$ 5.79} & 117.1 $\pm$ 49.33 & \textbf{67.62 $\pm$ 5.90} & 151.8 $\pm$ 65.44 & \textbf{72.94 $\pm$ 4.62} & 91.48 $\pm$ 15.24 & \textbf{64.74 $\pm$ 8.34} & \cellcolor{bestbg}54.76 \\
         & 720 & 74.95 $\pm$ 1.90 & \cellcolor{secondbg}\textbf{62.68 $\pm$ 4.62} & 68.41 $\pm$ 6.09 & \textbf{64.19 $\pm$ 2.82} & 118.9 $\pm$ 29.06 & \textbf{68.76 $\pm$ 9.08} & 148.9 $\pm$ 102.5 & \textbf{67.81 $\pm$ 7.98} & 95.17 $\pm$ 7.05 & \textbf{66.78 $\pm$ 3.87} & \cellcolor{bestbg}57.15 \\
        \hline
        \textbf{\sys~Win} & & \multicolumn{2}{c|}{\textbf{13}} & \multicolumn{2}{c|}{\textbf{16}} & \multicolumn{2}{c|}{\textbf{20}} & \multicolumn{2}{c|}{\textbf{14}} & \multicolumn{2}{c|}{\textbf{15}} & -- \\
        \hline
        \textbf{Impr.\ \% (mean)} & & \multicolumn{2}{c|}{\textbf{+10.6}} & \multicolumn{2}{c|}{\textbf{+7.7}} & \multicolumn{2}{c|}{\textbf{+35.7}} & \multicolumn{2}{c|}{\textbf{+31.8}} & \multicolumn{2}{c|}{\textbf{+19.0}} & -- \\
        \hline
    \end{tabular}
    }
\end{table*}
\endgroup

\subsection{Global vs.\ Sample Normalization}
\label{app:res_norm}
Table~\ref{app:norm} compares the two normalization schemes for every model. The standalone baselines are clearly sensitive to this choice: global normalization is better in 11--17 of the 20 settings, and sample normalization is drastically worse on Weather and Appliances for most baselines (e.g., $+2{,}976\%$ for FEDformer on Weather). With \sys, this gap shrinks considerably: the median difference over all settings lies between $-1.6\%$ and $+5.5\%$ for the \sys-enhanced models, compared with $+0.5\%$ to $+52\%$ for the baselines. Appliances is the only dataset on which global normalization is better for all models, including \sys. Time-MoE, which was pretrained with per-window normalization, is the only model that is much better with sample normalization on Weather ($-76\%$).

\begin{table*}[h]
\centering
\caption{Global vs.\ sample normalization. For each model and dataset, the cell is the median over the four horizons of the relative difference in mean MSE, (sample $-$ global)/global (\%); positive values mean that global normalization is better. The last two columns give the number of the 20 settings (5 datasets $\times$ 4 horizons) in which global normalization is better, and the median relative difference over all 20 settings.}
\label{app:norm}
\resizebox{\textwidth}{!}{%
\begin{tabular}{lrrrrrrr}
\toprule
\textbf{Model} & Weather & PM2.5-LA & PM2.5-RV & Bike Share & Appliances & \#Global better & Median (all) \\
\midrule
iTransformer & $+166$ & $+0.5$ & $-1.5$ & $-2.1$ & $+97$ & 11/20 & $+0.5$ \\
TimesNet & $+71$ & $+3.6$ & $+1.5$ & $-5.7$ & $+112$ & 13/20 & $+6.9$ \\
FEDformer & $+2{,}976$ & $+5.7$ & $+0.4$ & $+52$ & $+249$ & 17/20 & $+52$ \\
Autoformer & $+45$ & $-3.6$ & $+3.5$ & $+11$ & $+471$ & 14/20 & $+14$ \\
Informer & $-3.1$ & $+0.3$ & $+3.5$ & $-0.1$ & $+405$ & 14/20 & $+5.5$ \\
\midrule
\sys-iTransformer & $+5.9$ & $-0.7$ & $-1.3$ & $-1.4$ & $+35$ & 9/20 & $-0.4$ \\
\sys-TimesNet & $-2.1$ & $-7.8$ & $-0.7$ & $-4.2$ & $+36$ & 8/20 & $-1.6$ \\
\sys-FEDformer & $-55$ & $-2.9$ & $+0.7$ & $+13$ & $+41$ & 10/20 & $+0.1$ \\
\sys-Autoformer & $-14$ & $+0.4$ & $+1.0$ & $+15$ & $+23$ & 14/20 & $+5.5$ \\
\sys-Informer & $-48$ & $+0.3$ & $+0.2$ & $+0.5$ & $+59$ & 10/20 & $+0.2$ \\
\midrule
Time-MoE (zero-shot) & $-76$ & $+1.4$ & $+13$ & $+12$ & $+12$ & 13/20 & $+10$ \\
\bottomrule
\end{tabular}
}
\end{table*}

\subsection{Best Normalization per Model}
\label{app:res_bestnorm}
Since the better normalization scheme differs across models and datasets, Table~\ref{app:best_norm} compares each baseline and its \sys-enhanced version under their respective best normalization, i.e., the lower mean MSE of global and sample normalization. Under this comparison, \syss improves over its standalone counterpart in 82 of the 100 comparisons (iTransformer 13/20, TimesNet 17/20, FEDformer 20/20, Autoformer 16/20, Informer 16/20).

\begin{table*}[h]
\centering
\caption{Best normalization per model. For each model, we take the lower mean MSE of global (G) and sample (S) normalization (Tables~\ref{tab:main_global} and~\ref{tab:main_sample}), separately for the standalone baseline (Base) and its \sys-enhanced version; the superscript gives the selected normalization. Bold marks the better of each baseline and its \sys-enhanced version. \sys~Win: number of the 20 settings in which \sys~improves over its baseline; Impr.: relative MSE reduction (\%), averaged over the 20 settings.}
\label{app:best_norm}
\resizebox{\textwidth}{!}{%
\begin{tabular}{llrrrrrrrrrr}
\toprule
 & & \multicolumn{2}{c}{iTransformer} & \multicolumn{2}{c}{TimesNet} & \multicolumn{2}{c}{FEDformer} & \multicolumn{2}{c}{Autoformer} & \multicolumn{2}{c}{Informer} \\
\cmidrule(lr){3-4}\cmidrule(lr){5-6}\cmidrule(lr){7-8}\cmidrule(lr){9-10}\cmidrule(lr){11-12}
Dataset & $L_y$ & Base & \sys & Base & \sys & Base & \sys & Base & \sys & Base & \sys \\
\midrule
\multirow{4}{*}{Weather} & 96 & 204.7$^{\mathrm{G}}$ & \textbf{178.3$^{\mathrm{S}}$} & 245.6$^{\mathrm{G}}$ & \textbf{158.5$^{\mathrm{S}}$} & 28{,}741$^{\mathrm{G}}$ & \textbf{181.7$^{\mathrm{S}}$} & 39{,}860$^{\mathrm{G}}$ & \textbf{210.1$^{\mathrm{S}}$} & 1{,}733$^{\mathrm{G}}$ & \textbf{155.7$^{\mathrm{S}}$} \\
 & 192 & 249.3$^{\mathrm{G}}$ & \textbf{226.7$^{\mathrm{G}}$} & 271.1$^{\mathrm{G}}$ & \textbf{226.1$^{\mathrm{S}}$} & 28{,}712$^{\mathrm{G}}$ & \textbf{236.8$^{\mathrm{S}}$} & 151{,}937$^{\mathrm{S}}$ & \textbf{285.4$^{\mathrm{S}}$} & 6{,}645$^{\mathrm{G}}$ & \textbf{220.8$^{\mathrm{S}}$} \\
 & 336 & 254.6$^{\mathrm{G}}$ & \textbf{232.3$^{\mathrm{G}}$} & 264.8$^{\mathrm{G}}$ & \textbf{241.8$^{\mathrm{S}}$} & 50{,}859$^{\mathrm{G}}$ & \textbf{267.3$^{\mathrm{S}}$} & 70{,}468$^{\mathrm{G}}$ & \textbf{270.1$^{\mathrm{S}}$} & 3{,}729$^{\mathrm{S}}$ & \textbf{240.1$^{\mathrm{S}}$} \\
 & 720 & 270.1$^{\mathrm{G}}$ & \textbf{235.9$^{\mathrm{G}}$} & 269.1$^{\mathrm{G}}$ & \textbf{241.7$^{\mathrm{G}}$} & 41{,}001$^{\mathrm{G}}$ & \textbf{268.3$^{\mathrm{S}}$} & 29{,}367$^{\mathrm{S}}$ & \textbf{259.7$^{\mathrm{G}}$} & 3{,}637$^{\mathrm{S}}$ & \textbf{290.7$^{\mathrm{S}}$} \\
\midrule
\multirow{4}{*}{PM2.5-LA} & 96 & 36.08$^{\mathrm{S}}$ & \textbf{35.20$^{\mathrm{S}}$} & 37.32$^{\mathrm{G}}$ & \textbf{35.43$^{\mathrm{S}}$} & 57.75$^{\mathrm{G}}$ & \textbf{42.11$^{\mathrm{S}}$} & \textbf{57.60$^{\mathrm{S}}$} & 58.89$^{\mathrm{S}}$ & 41.30$^{\mathrm{S}}$ & \textbf{37.90$^{\mathrm{S}}$} \\
 & 192 & \textbf{40.64$^{\mathrm{G}}$} & 40.81$^{\mathrm{S}}$ & 40.16$^{\mathrm{G}}$ & \textbf{39.25$^{\mathrm{S}}$} & 64.07$^{\mathrm{S}}$ & \textbf{43.77$^{\mathrm{S}}$} & \textbf{61.37$^{\mathrm{S}}$} & 64.94$^{\mathrm{G}}$ & 42.06$^{\mathrm{S}}$ & \textbf{39.62$^{\mathrm{S}}$} \\
 & 336 & \textbf{48.71$^{\mathrm{G}}$} & 49.35$^{\mathrm{G}}$ & 45.91$^{\mathrm{G}}$ & \textbf{44.67$^{\mathrm{S}}$} & 67.49$^{\mathrm{G}}$ & \textbf{52.00$^{\mathrm{S}}$} & 70.55$^{\mathrm{G}}$ & \textbf{69.27$^{\mathrm{G}}$} & 48.22$^{\mathrm{G}}$ & \textbf{44.79$^{\mathrm{G}}$} \\
 & 720 & \textbf{62.55$^{\mathrm{G}}$} & 63.88$^{\mathrm{S}}$ & 59.23$^{\mathrm{S}}$ & \textbf{58.73$^{\mathrm{S}}$} & 75.79$^{\mathrm{G}}$ & \textbf{64.62$^{\mathrm{S}}$} & 106.7$^{\mathrm{G}}$ & \textbf{72.23$^{\mathrm{S}}$} & 52.81$^{\mathrm{G}}$ & \textbf{51.37$^{\mathrm{G}}$} \\
\midrule
\multirow{4}{*}{PM2.5-RV} & 96 & 78.83$^{\mathrm{S}}$ & \textbf{78.43$^{\mathrm{S}}$} & 78.84$^{\mathrm{S}}$ & \textbf{73.83$^{\mathrm{S}}$} & 105.0$^{\mathrm{S}}$ & \textbf{87.11$^{\mathrm{G}}$} & \textbf{125.5$^{\mathrm{S}}$} & 130.1$^{\mathrm{S}}$ & 82.47$^{\mathrm{G}}$ & \textbf{81.69$^{\mathrm{S}}$} \\
 & 192 & 90.67$^{\mathrm{S}}$ & \textbf{89.73$^{\mathrm{S}}$} & 87.78$^{\mathrm{G}}$ & \textbf{83.44$^{\mathrm{G}}$} & 131.5$^{\mathrm{G}}$ & \textbf{96.72$^{\mathrm{S}}$} & 142.6$^{\mathrm{G}}$ & \textbf{133.5$^{\mathrm{G}}$} & 94.80$^{\mathrm{G}}$ & \textbf{91.40$^{\mathrm{S}}$} \\
 & 336 & \textbf{94.49$^{\mathrm{S}}$} & 95.08$^{\mathrm{S}}$ & 93.31$^{\mathrm{G}}$ & \textbf{91.94$^{\mathrm{G}}$} & 128.1$^{\mathrm{G}}$ & \textbf{103.8$^{\mathrm{G}}$} & \textbf{136.3$^{\mathrm{G}}$} & 136.9$^{\mathrm{G}}$ & \textbf{100.1$^{\mathrm{G}}$} & 100.6$^{\mathrm{G}}$ \\
 & 720 & 87.28$^{\mathrm{S}}$ & \textbf{86.70$^{\mathrm{S}}$} & \textbf{85.81$^{\mathrm{S}}$} & 86.12$^{\mathrm{S}}$ & 104.5$^{\mathrm{S}}$ & \textbf{94.83$^{\mathrm{G}}$} & 122.7$^{\mathrm{S}}$ & \textbf{110.8$^{\mathrm{G}}$} & 92.70$^{\mathrm{G}}$ & \textbf{89.88$^{\mathrm{G}}$} \\
\midrule
\multirow{4}{*}{Bike Share} & 96 & 15{,}134$^{\mathrm{S}}$ & \textbf{14{,}258$^{\mathrm{S}}$} & 15{,}418$^{\mathrm{S}}$ & \textbf{14{,}886$^{\mathrm{S}}$} & 18{,}171$^{\mathrm{G}}$ & \textbf{14{,}944$^{\mathrm{S}}$} & 24{,}641$^{\mathrm{G}}$ & \textbf{21{,}516$^{\mathrm{G}}$} & 16{,}080$^{\mathrm{S}}$ & \textbf{15{,}038$^{\mathrm{S}}$} \\
 & 192 & 21{,}140$^{\mathrm{S}}$ & \textbf{20{,}356$^{\mathrm{S}}$} & \textbf{20{,}500$^{\mathrm{S}}$} & 20{,}508$^{\mathrm{S}}$ & 25{,}283$^{\mathrm{G}}$ & \textbf{19{,}930$^{\mathrm{G}}$} & 32{,}384$^{\mathrm{G}}$ & \textbf{29{,}645$^{\mathrm{G}}$} & \textbf{22{,}524$^{\mathrm{S}}$} & 22{,}916$^{\mathrm{S}}$ \\
 & 336 & 27{,}477$^{\mathrm{S}}$ & \textbf{26{,}907$^{\mathrm{S}}$} & 27{,}702$^{\mathrm{S}}$ & \textbf{27{,}661$^{\mathrm{G}}$} & 28{,}721$^{\mathrm{G}}$ & \textbf{23{,}498$^{\mathrm{G}}$} & 39{,}925$^{\mathrm{G}}$ & \textbf{37{,}735$^{\mathrm{G}}$} & \textbf{28{,}818$^{\mathrm{G}}$} & 32{,}018$^{\mathrm{G}}$ \\
 & 720 & \textbf{36{,}714$^{\mathrm{S}}$} & 37{,}451$^{\mathrm{G}}$ & \textbf{35{,}968$^{\mathrm{S}}$} & 38{,}228$^{\mathrm{S}}$ & 34{,}864$^{\mathrm{G}}$ & \textbf{32{,}858$^{\mathrm{G}}$} & 44{,}139$^{\mathrm{G}}$ & \textbf{39{,}982$^{\mathrm{G}}$} & \textbf{34{,}947$^{\mathrm{G}}$} & 35{,}416$^{\mathrm{G}}$ \\
\midrule
\multirow{4}{*}{Appliances} & 96 & 8{,}095$^{\mathrm{G}}$ & \textbf{8{,}084$^{\mathrm{G}}$} & 9{,}864$^{\mathrm{G}}$ & \textbf{8{,}350$^{\mathrm{G}}$} & 10{,}091$^{\mathrm{G}}$ & \textbf{9{,}253$^{\mathrm{G}}$} & 16{,}935$^{\mathrm{G}}$ & \textbf{9{,}941$^{\mathrm{G}}$} & 8{,}921$^{\mathrm{G}}$ & \textbf{7{,}950$^{\mathrm{G}}$} \\
 & 192 & \textbf{8{,}123$^{\mathrm{G}}$} & 8{,}129$^{\mathrm{G}}$ & 10{,}121$^{\mathrm{G}}$ & \textbf{8{,}468$^{\mathrm{G}}$} & 9{,}436$^{\mathrm{G}}$ & \textbf{9{,}027$^{\mathrm{G}}$} & 13{,}646$^{\mathrm{G}}$ & \textbf{10{,}650$^{\mathrm{G}}$} & 9{,}172$^{\mathrm{G}}$ & \textbf{7{,}739$^{\mathrm{G}}$} \\
 & 336 & 7{,}847$^{\mathrm{G}}$ & \textbf{7{,}761$^{\mathrm{G}}$} & 9{,}843$^{\mathrm{G}}$ & \textbf{8{,}044$^{\mathrm{G}}$} & 9{,}147$^{\mathrm{G}}$ & \textbf{8{,}559$^{\mathrm{G}}$} & 11{,}739$^{\mathrm{G}}$ & \textbf{10{,}198$^{\mathrm{G}}$} & 8{,}329$^{\mathrm{G}}$ & \textbf{7{,}681$^{\mathrm{G}}$} \\
 & 720 & \textbf{8{,}046$^{\mathrm{G}}$} & 8{,}048$^{\mathrm{G}}$ & 8{,}861$^{\mathrm{G}}$ & \textbf{8{,}280$^{\mathrm{G}}$} & 10{,}063$^{\mathrm{G}}$ & \textbf{9{,}080$^{\mathrm{G}}$} & 13{,}574$^{\mathrm{G}}$ & \textbf{9{,}706$^{\mathrm{G}}$} & 8{,}261$^{\mathrm{G}}$ & \textbf{7{,}870$^{\mathrm{G}}$} \\
\midrule
\sys~Win & & \multicolumn{2}{c}{13/20} & \multicolumn{2}{c}{17/20} & \multicolumn{2}{c}{20/20} & \multicolumn{2}{c}{16/20} & \multicolumn{2}{c}{16/20} \\
Impr.\ (\%) & & \multicolumn{2}{c}{+2.7} & \multicolumn{2}{c}{+7.4} & \multicolumn{2}{c}{+32.9} & \multicolumn{2}{c}{+28.8} & \multicolumn{2}{c}{+21.8} \\
\bottomrule
\end{tabular}
}
\end{table*}


\subsection{Instability of the Baselines on Weather}
\label{app:weather}
The target of Weather, the CO$_2$ concentration, contains 50 missing-value codes ($-9999$\,ppm), which we keep as in the original data. All of them lie in the training split (February to August 2020), while the validation and test splits contain none. The regular measurements have a mean of 425\,ppm and a standard deviation of 18.5\,ppm in the training split; with the missing-value codes, the standard deviation of the training target increases to 359\,ppm. Under global normalization, the regular measurements therefore occupy a very narrow range of the normalized space, and a small error in this space becomes a large error in ppm: an error of 0.5 in the normalized space corresponds to about 180\,ppm, i.e., an MSE of about $3 \times 10^4$, the order of magnitude observed for FEDformer and Autoformer (Table~\ref{tab:main_global}). Under sample normalization, the input windows that contain such a value are dominated by it. A model that is pulled toward these extreme training targets therefore produces large errors on the clean test data, which explains the very large MSE and SD of FEDformer, Autoformer, and Informer on Weather. Their \sys-enhanced counterparts are trained on the same data but remain within the range of the strong baselines (Section~\ref{sec:comp1}).

\subsection{Computational Cost}
\label{app:res_cost}
Table~\ref{app:cost} reports the training time per epoch, the peak GPU memory, and the number of trained epochs of the baselines and the \sys-enhanced models. In each training iteration, \syss runs a forward pass of all $K$ workers to compute the pseudo ground truth, trains the manager, and updates the selected workers $\alpha=2$ times; in addition, the parameters and optimizer states of all $K$ workers and the manager are kept in memory. As a result, a \sys-enhanced model requires a median of $5.8\times$ the training time per epoch and $3.8\times$ the peak memory of its standalone counterpart across all models and horizons, and it is trained for about 27 epochs instead of about 7. The relative overhead is largest for the lightweight iTransformer and smallest for the heavier FEDformer, Autoformer, and Informer. At inference time, only the manager and the selected worker are executed for each input.

\begin{table*}[h]
\centering
\caption{Computational cost of the baselines and \sys-enhanced models (3 workers): training time per epoch (s), peak GPU memory during training (GB), and the average number of trained epochs, for each forecasting horizon $L_y$. Values are averaged over the 5 datasets, 2 normalization schemes, and 5 runs. Several runs shared one GPU, so the times are indicative. For the zero-shot Time-MoE (50M), we report the total inference time on the test split and the peak memory.}
\label{app:cost}
\resizebox{\textwidth}{!}{%
\begin{tabular}{llrrrrrrrrr}
\toprule
 & & \multicolumn{2}{c}{$L_y=96$} & \multicolumn{2}{c}{$L_y=192$} & \multicolumn{2}{c}{$L_y=336$} & \multicolumn{2}{c}{$L_y=720$} & \\
\cmidrule(lr){3-4}\cmidrule(lr){5-6}\cmidrule(lr){7-8}\cmidrule(lr){9-10}
\textbf{Model} & & Time (s/ep.) & Mem. (GB) & Time (s/ep.) & Mem. (GB) & Time (s/ep.) & Mem. (GB) & Time (s/ep.) & Mem. (GB) & Epochs \\
\midrule
iTransformer & & 17 & 0.04 & 19 & 0.04 & 33 & 0.09 & 23 & 0.05 & 7.1 \\
\sys-iTransformer & & 283 & 0.46 & 239 & 0.46 & 398 & 0.48 & 250 & 0.48 & 26.7 \\
\midrule
TimesNet & & 76 & 0.13 & 95 & 0.15 & 728 & 0.20 & 205 & 0.29 & 7.0 \\
\sys-TimesNet & & 725 & 0.57 & 559 & 0.65 & 2{,}227 & 0.79 & 922 & 1.00 & 26.1 \\
\midrule
FEDformer & & 301 & 1.62 & 329 & 1.78 & 1{,}443 & 2.07 & 499 & 2.68 & 6.7 \\
\sys-FEDformer & & 1{,}688 & 5.96 & 1{,}888 & 6.57 & 2{,}226 & 7.43 & 1{,}460 & 9.78 & 26.5 \\
\midrule
Autoformer & & 172 & 2.45 & 193 & 3.83 & 1{,}912 & 5.63 & 485 & 11.07 & 6.9 \\
\sys-Autoformer & & 1{,}571 & 9.32 & 1{,}859 & 14.73 & 2{,}998 & 21.69 & 1{,}358 & 43.17 & 26.9 \\
\midrule
Informer & & 100 & 0.78 & 167 & 0.94 & 1{,}234 & 1.23 & 300 & 1.85 & 7.1 \\
\sys-Informer & & 884 & 2.60 & 1{,}413 & 3.22 & 2{,}240 & 4.07 & 1{,}149 & 6.36 & 27.1 \\
\midrule
Time-MoE (zero-shot) & & 994$^\dagger$ & 0.23 & 1{,}165$^\dagger$ & 0.23 & 767$^\dagger$ & 0.24 & 1{,}241$^\dagger$ & 0.26 & -- \\
\bottomrule
\end{tabular}
}
\\[2pt]{\footnotesize $^\dagger$Total inference time (s) on the test split.}
\end{table*}

\FloatBarrier

\section{Ablation Study}
\label{app:ablation}

Table~\ref{app:ablation_full} reports the full results of the ablation study in Section~\ref{sec:ablation} on PM2.5-RV, together with the same ablation on Bike Share and the number of parameters. On Bike Share, the default setting is again an independent re-run of the configuration in Table~\ref{tab:main}. For the mixture of workers, we combine the top-performing \sys-enhanced models of each dataset.

\begin{table*}[h]
\centering
\caption{Full ablation results of \sys-iTransformer on Bike Share and PM2.5-RV (global normalization, $L=L_y=96$). MSE and MAE are on the original scale, mean $\pm$ SD over 5 runs without fixed random seeds. $\Delta$ is the relative change in mean MSE w.r.t.\ the default setting of the same dataset (positive is worse); $p$ is from Welch's $t$-test against the default (bold: $p<0.05$). \#Params counts the manager and all workers on Bike Share.}
\label{app:ablation_full}
\resizebox{\textwidth}{!}{%
\begin{tabular}{llrrrrrrrrr}
\toprule
 & & \multicolumn{4}{c}{Bike Share} & \multicolumn{4}{c}{PM2.5-RV} & \\
\cmidrule(lr){3-6}\cmidrule(lr){7-10}
\multicolumn{2}{l}{\textbf{Setting}} & \textbf{MSE} & \textbf{MAE} & $\Delta$ (\%) & $p$ & \textbf{MSE} & \textbf{MAE} & $\Delta$ (\%) & $p$ & \#Params (M) \\
\multicolumn{2}{l}{Default (KL + Wass., reg., $d_{model}{=}512$, $K{=}3$)} & 15{,}097 $\pm$ 241 & 78.56 $\pm$ 0.67 & -- & -- & 80.52 $\pm$ 1.02 & 6.18 $\pm$ 0.03 & -- & -- & 7.3 \\
\midrule
\multirow{2}{*}{Manager loss} & KL only & 15{,}452 $\pm$ 282 & 79.36 $\pm$ 0.71 & $+2.4$ & 0.065 & 81.87 $\pm$ 2.75 & 6.15 $\pm$ 0.03 & $+1.7$ & 0.347 & 7.3 \\
 & Wasserstein only & 15{,}407 $\pm$ 372 & 80.48 $\pm$ 0.83 & $+2.1$ & 0.163 & 81.56 $\pm$ 1.10 & 6.21 $\pm$ 0.04 & $+1.3$ & 0.157 & 7.3 \\
\midrule
\multirow{1}{*}{Regularization} & No reg. & 15{,}169 $\pm$ 184 & 79.01 $\pm$ 0.40 & $+0.5$ & 0.609 & 80.94 $\pm$ 1.19 & 6.14 $\pm$ 0.03 & $+0.5$ & 0.559 & 7.3 \\
\midrule
\multirow{2}{*}{Manager arch.} & Small ($d_{model}{=}128$, $d_{ff}{=}512$, 4 heads) & 15{,}155 $\pm$ 298 & 79.32 $\pm$ 0.77 & $+0.4$ & 0.743 & 81.23 $\pm$ 0.78 & 6.19 $\pm$ 0.04 & $+0.9$ & 0.252 & 1.1 \\
 & Large ($d_{model}{=}768$, $d_{ff}{=}3076$, 12 heads) & 15{,}042 $\pm$ 167 & 78.21 $\pm$ 0.60 & $-0.4$ & 0.691 & 81.05 $\pm$ 0.81 & 6.17 $\pm$ 0.05 & $+0.7$ & 0.386 & 15.5 \\
\midrule
\multirow{2}{*}{Worker number} & $K{=}6$ & 15{,}321 $\pm$ 302 & 79.26 $\pm$ 1.09 & $+1.5$ & 0.232 & 79.37 $\pm$ 1.14 & 6.08 $\pm$ 0.10 & $-1.4$ & 0.132 & 7.9 \\
 & $K{=}9$ & 15{,}467 $\pm$ 172 & 79.93 $\pm$ 0.75 & $+2.5$ & \textbf{0.026} & 82.61 $\pm$ 2.99 & 6.11 $\pm$ 0.07 & $+2.6$ & 0.200 & 8.6 \\
\midrule
\multirow{2}{*}{Worker mix} & Top-3 models$^{a}$ ($K{=}3$) & 17{,}067 $\pm$ 579 & 82.88 $\pm$ 1.15 & $+13.1$ & \textbf{$<$0.001} & 81.63 $\pm$ 2.13 & 6.13 $\pm$ 0.04 & $+1.4$ & 0.334 & 24.9 \\
 & Top-2 models$^{a}$ $\times$2 ($K{=}4$) & 14{,}728 $\pm$ 837 & 79.98 $\pm$ 1.52 & $-2.4$ & 0.390 & 80.04 $\pm$ 2.70 & 6.07 $\pm$ 0.06 & $-0.6$ & 0.728 & 40.8 \\
\bottomrule
\end{tabular}
}
\\[2pt]
\parbox{\textwidth}{\footnotesize $^{a}$Top-performing models of each dataset in Table~\ref{tab:main_global}: Bike Share: top-3 = iTransformer + FEDformer + TimesNet, top-2 = iTransformer + FEDformer; PM2.5-RV: top-3 = TimesNet + iTransformer + Informer, top-2 = TimesNet + iTransformer.}
\end{table*}

\FloatBarrier

\section{Details of the Case Study}
\label{app:context}

This section describes the analysis methods (Appendix~\ref{app:context_methods}), validates them on synthetic data with known regimes (Appendix~\ref{app:context_syn}), gives further results of the Bike Share case study in Section~\ref{sec:context} (Appendix~\ref{app:context_bike}), and repeats the analysis on PM2.5-RV (Appendix~\ref{app:context_rv}).

\subsection{Analysis Methods}
\label{app:context_methods}
All analyses use \sys-iTransformer ($L=L_y=96$, global normalization, $K=3$). The figures show a single trained model; quantitative results are averaged over all runs (3 on the synthetic data, 5 on Bike Share).

\parasum{Routing by context.} We compare the Worker selected for each sample with a property of its context, e.g., the true regime or the day of the week, and quantify the agreement with the normalized mutual information (NMI) between the two labelings, $\mathrm{NMI}(A,B) = 2\,I(A;B)/(H(A)+H(B))$, where $I$ is the mutual information and $H$ the entropy. The NMI is 1 if the two labelings match one-to-one and 0 if the routing is independent of the property.

\parasum{Permutation importance~\cite{breiman2001random}.} For one variable at a time, we replace its input window in each test sample with that of a randomly chosen other test sample, while all other variables stay unchanged. This is done in the Manager input only; the Workers still receive the original input. We record the share of test samples whose selected Worker changes, averaged over three random shuffles. ``All context'' shuffles all context variables at once.

\parasum{Sanity check.} Following the model-randomization test of \citet{adebayo2018sanity}, we repeat the analyses with an untrained Manager of the same architecture. A pattern that also appears for the untrained Manager reflects the input data rather than what the Manager has learned.

\parasum{Intervention.} This analysis is performed during inference. We keep the trained Workers unchanged (they are not retrained) and, when forecasting the test set, replace only the learned routing with (a) a random assignment with the same Worker shares, (b) the best single Worker for all samples, and (c) the untrained Manager. If the routing is meaningful, each replacement should increase the MSE. We also record how often the Manager selects the Worker with the lowest error on a sample, and, for the samples routed to each Worker, the MSE of every Worker relative to the lowest one, which shows whether the Workers specialize.

\parasum{Counterfactual (Bike Share).} We change selected context variables in the Manager input and keep all other inputs fixed. The calendar variables (the day of the week and the working-day indicator) are changed in their original units and then normalized as in training. When the day of the week of the input window is shifted by $d \in \{1, \dots, 6\}$ days, the working-day indicator is recomputed accordingly. The typical Worker of a day is the majority Worker of the real forecasts made on that day at the same hour. Unlike the two previous analyses, this tests whether a variable causes the routing decision rather than whether it is associated with it.

\subsection{Synthetic Data: Recovering Known Regimes}
\label{app:context_syn}
\label{sec:case_syn}
On real data, the correct assignment of samples to Workers is unknown, so an interpretation can only be checked indirectly, e.g., against domain knowledge. We therefore construct a synthetic dataset in which the regime of the target is determined by the context, so that the interpretation can be compared with the ground truth: if \sys~is context-aware, the routing should follow the regimes, the Manager should rely on the variable that announces the regime, and each Worker should handle one regime.

\parasum{Data.} The dataset contains $T=20{,}000$ hourly steps $t = 0, \dots, T-1$ and is generated in three steps.

\emph{(1) Regime sequence.} We first draw a regime $r_t \in \{1, 2, 3\}$ for every step: the time axis is split into consecutive segments whose lengths are drawn uniformly from 72 to 240 hours (3 to 10 days), and each segment is assigned a regime drawn uniformly from the two regimes that differ from the previous segment. The regime sequence is random and does not depend on any variable; each regime covers about one third of the steps.

\emph{(2) Target.} The target is generated from the regime of the current step:
\begin{equation*}
y_t = \mu_t + \epsilon_t, \quad \epsilon_t \sim \mathcal{N}(0, 1), \qquad
\mu_t = \begin{cases}
20 + 10 \sin(2\pi t / 24) & \text{if } r_t = 1 \text{ (daily cycle)},\\
20 + 10 \sin(2\pi t / 12 + \pi/3) & \text{if } r_t = 2 \text{ (half-day cycle)},\\
20 + a_t & \text{if } r_t = 3 \text{ (drift)},
\end{cases}
\end{equation*}
where $a_t = 0.97\, a_{t-1} + \eta_t$, $\eta_t \sim \mathcal{N}(0, 2^2)$, $a_0 = 0$, is a first-order autoregressive (AR(1)) process, which drifts slowly around zero without any periodicity. The three regimes thus have clearly different dynamics: one peak per day, two peaks per day, and no cycle.

\emph{(3) Context.} There are six context variables. Only one of them, the \emph{regime signal}, carries information about the regime: it equals the regime 48 hours ahead plus small Gaussian noise,
\begin{equation*}
s_t = r_{t+48} + \nu_t, \quad \nu_t \sim \mathcal{N}(0, 0.1^2),
\end{equation*}
similar to a forecast of an upcoming condition. It does not enter the generation of the target; the target and the regime signal are only linked through the regime sequence, the target through the current regime $r_t$ and the regime signal through the regime 48 hours ahead. The other five context variables are uninformative about the regime: four independent random walks $z^{(k)}_t = z^{(k)}_{t-1} + \xi^{(k)}_t$, $\xi^{(k)}_t \sim \mathcal{N}(0, 1)$, $k = 1, \dots, 4$, and the hour of the day encoded as $h_t = \sin(2\pi (t \bmod 24)/24)$. The hour of the day can help to predict the values of the two periodic regimes, but it does not indicate which regime is active.

\parasum{Task and ground truth.} As in the main experiments, the model receives the past $L=96$ hours of the target and the six context variables and forecasts the next $L_y=96$ hours of the target. The true regime of a sample is the regime that covers most of its forecast window. The target history reveals the current regime, but not an upcoming switch; the regime signal announces a switch 48 hours in advance. A context-aware Manager should therefore route the samples by regime and rely on the regime signal. We use the same splits (80/10/10), global normalization, and hyperparameters as for \sys-iTransformer in the main experiments, and run each model 3 times.

\begin{figure*}[tbp]
\centering
\includegraphics[width=\textwidth]{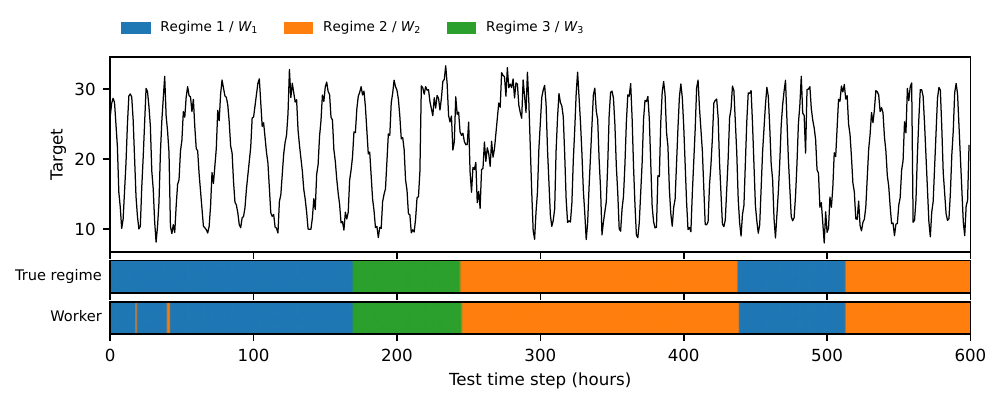}
\caption{Routing on synthetic data over the first 600 test steps (one run). Top: target series. Middle: true regime of the forecast window. Bottom: worker selected by the manager. The selected worker follows the regime switches; the agreement is 99.4\% over the whole test set.}
\label{app:syn_timeline}
\end{figure*}

\begin{figure*}[tbp]
\centering
\includegraphics[width=\textwidth]{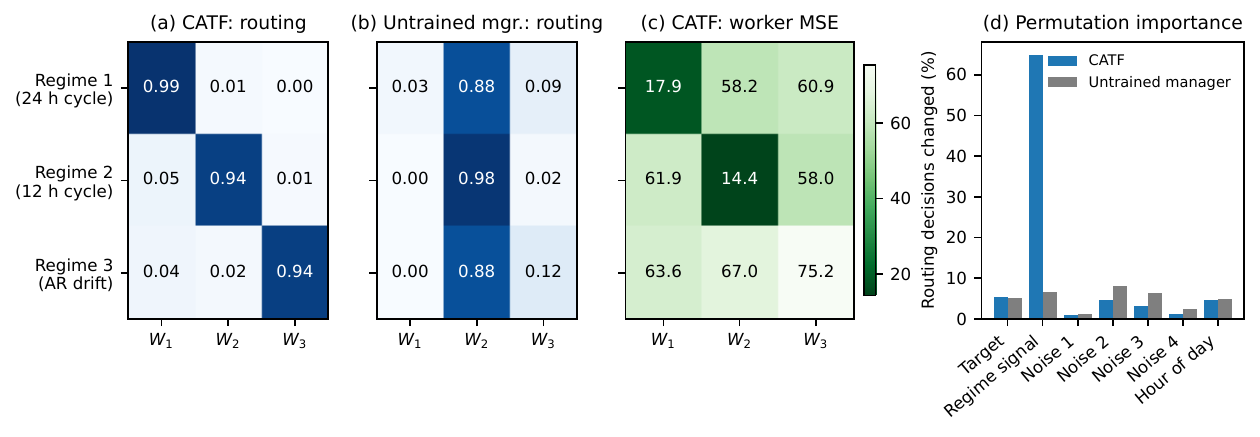}
\caption{Routing on synthetic data, averaged over 3 runs. (a) Share of the samples of each true regime that the trained manager routes to each worker. (b) The same for an untrained manager with the same architecture and the same trained workers. (c) MSE of each worker on each regime. (d) Permutation importance: share of routing decisions that change when one input variable is shuffled across test samples in the manager input only (workers receive the original input), for the trained and the untrained manager. Workers are relabeled in each run so that $W_k$ is the worker that receives regime $k$ most often.}
\label{app:syn_routing}
\end{figure*}

\parasum{Does the routing recover the regimes?} Figure~\ref{app:syn_timeline} shows the test series together with the true regime and the selected Worker: the selected Worker switches whenever the regime switches, and the routing agrees with the regime for 99.4\% of the test samples. Over all runs, 99\%, 94\%, and 94\% of the samples of the three regimes are sent to one Worker each (Figure~\ref{app:syn_routing}a), and the NMI between routing and regime is $0.85$. An untrained Manager with the same input reaches an NMI of only $0.04$ (Figure~\ref{app:syn_routing}b).

\parasum{Does the Manager use the right variable?} Shuffling the regime signal in the Manager input changes 65\% of the routing decisions, while shuffling any other variable, including the target history, changes at most 5.3\% on average (Figure~\ref{app:syn_routing}d). For the untrained Manager, the regime signal is no more important than the distractors. Inspecting the Manager alone therefore reveals which context variable drives the regimes, without access to the data-generating process.

\parasum{What has each Worker learned?} $W_1$ and $W_2$ each specialize in one of the two periodic regimes, and $W_3$ absorbs the drift regime: on the daily cycle, the MSE of $W_1$ is 17.9 while that of the other Workers is 58--61; on the half-day cycle, the MSE of $W_2$ is 14.4 while that of the other Workers is 58--62 (Figure~\ref{app:syn_routing}c). The drifting regime is not predictable 96 hours ahead, and all Workers have a high MSE on it (64--75); routing it to a separate Worker keeps it out of the training data of $W_1$ and $W_2$.

\parasum{Does the learned routing improve the forecasts?} With the same trained Workers, the learned routing (MSE 33.4) is much better than random routing (50.7), the best single Worker (43.8), and an untrained Manager (58.1), and \sys-iTransformer reduces the MSE of the standalone iTransformer by 7.2\% (33.4 vs.\ 36.0; Table~\ref{app:synthetic}, which also gives the standard deviations over the 3 runs).

In summary, when the regime depends on the context, the routing analysis recovers the true structure of the data: which samples belong together (the regimes), which variable determines them (the regime signal), and what each Worker has learned (one periodic regime each for $W_1$ and $W_2$, the drift regime for $W_3$).

\begin{table}[htbp]
\centering
\caption{Synthetic data ($L=L_y=96$, global normalization, 3 runs, mean $\pm$ SD). The routing variants reuse the trained workers and change only the routing. NMI: between the routing and the true regime.}
\label{app:synthetic}
\small
\setlength{\tabcolsep}{4pt}
\renewcommand{\arraystretch}{0.95}
\begin{tabular}{lrr}
\toprule
\textbf{Model / routing} & \textbf{MSE} & \textbf{NMI} \\
\midrule
iTransformer (baseline) & 35.98 $\pm$ 0.34 & -- \\
\sys-iTransformer & \textbf{33.39 $\pm$ 0.51} & 0.85 $\pm$ 0.15 \\
\midrule
\multicolumn{3}{l}{\emph{Same trained workers, different routing}} \\
\quad Random routing (same worker shares) & 50.69 $\pm$ 4.18 & -- \\
\quad Best single worker for all samples & 43.77 $\pm$ 1.76 & -- \\
\quad Untrained manager & 58.13 $\pm$ 6.06 & 0.04 $\pm$ 0.04 \\
\bottomrule
\end{tabular}
\end{table}

\subsection{Bike Share: Additional Results}
\label{app:context_bike}

\parasum{Routing accuracy.} The Manager selects the most accurate Worker for 54\% of the test samples, compared with 36\% for the untrained Manager and 33\% by chance. On the samples routed to a Worker, that Worker has the lowest MSE in 80\% of the cases.

\begin{figure}[tbp]
\centering
\includegraphics[width=\linewidth]{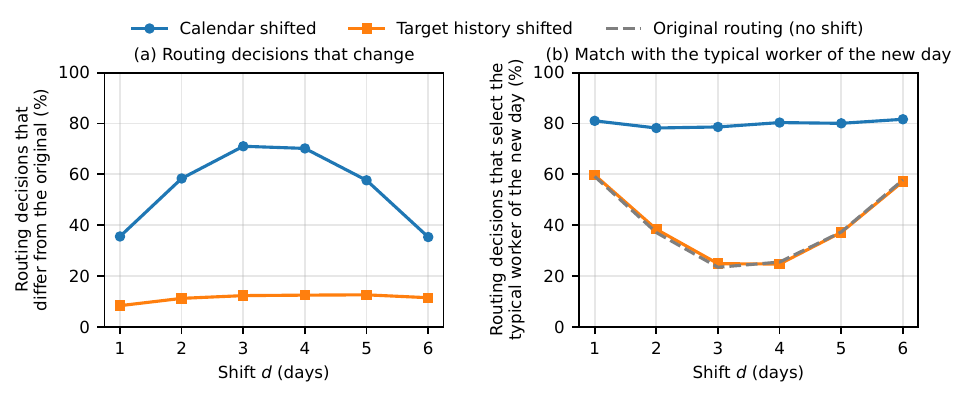}
\caption{Counterfactual analysis of the routing of \sys-iTransformer on Bike Share (mean over 5 runs). For each shift $d$, either the calendar of the input window (day of the week and working-day indicator) is shifted by $d$ days (blue) or the target history is replaced with that of the forecast made $d$ days later (orange), with all other inputs unchanged. (a) Percentage of test forecasts whose selected worker differs from the one selected for the original input. (b) Percentage of test forecasts whose selected worker is the typical worker of the new day, i.e., the majority worker of the real forecasts made $d$ days later at the same hour; the dashed line is this percentage for the original input before any change.}
\label{app:bike_counterfactual}
\end{figure}

\parasum{Counterfactual analysis.} Figure~\ref{app:bike_counterfactual} shows the counterfactual analysis of Section~\ref{sec:context} for each shift $d$. Shifting the calendar changes 35\% of the routing decisions for $d=1$ or $6$ and about 70\% for $d=3$ or $4$: a shift by one day often stays within the days handled by the same Worker, whereas a shift by half a week moves most forecasts to the days of another Worker. For every $d$, 78--82\% of the decisions after the shift match the typical Worker of the new day, compared with 23--60\% for the original routing. Replacing the target history instead changes only 8--13\% of the decisions, and the routing does not follow the day from which the history is taken: its match with the typical Worker of that day is the same as for the original routing. The Manager therefore routes by the calendar rather than by the shape of the recent demand.

\parasum{Consistency across runs.} All five runs partition the samples by the day of the week (NMI between the selected Worker and the day of the week 0.13--0.48, vs.\ at most 0.10 for the hour of the day), although the boundaries between the Workers differ between runs. In the counterfactual analysis, the share of new decisions that match the typical Worker of the shifted day is 64--87\% across runs (40\% before the shift), and replacing the target history changes 5--23\% of the decisions; both results hold in all five runs.

\begin{figure*}[tbp]
\centering
\includegraphics[width=\textwidth]{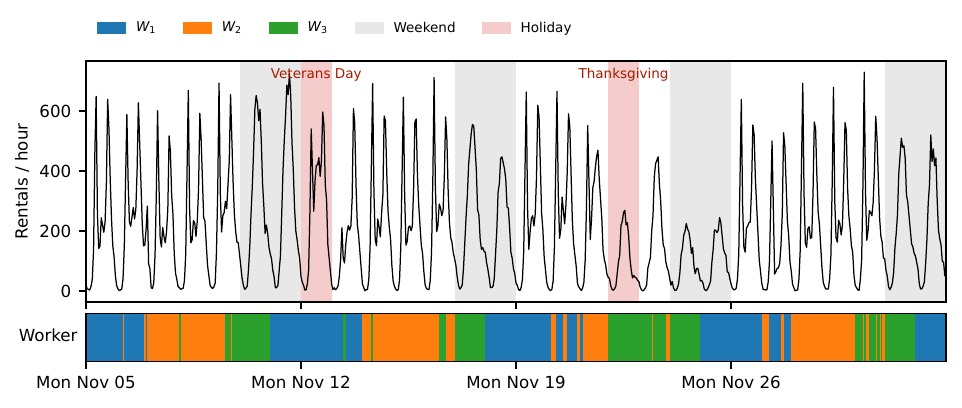}
\caption{Routing of \sys-iTransformer on Bike Share over four weeks of the test period (one run). Top: hourly bike rentals, with weekends in gray and holidays in red. Bottom: the worker selected for the forecast that starts at each hour. The routing follows the weekly cycle: $W_1$ for forecasts made on Sunday to Tuesday, $W_2$ for Wednesday to Friday, and $W_3$ for Friday and Saturday. Thanksgiving (Thursday, Nov 22) is routed to $W_3$, like a Saturday.}
\label{app:bike_timeline}
\end{figure*}

\parasum{Routing over time and holidays.} Figure~\ref{app:bike_timeline} shows the routing of one run over four weeks of the test period. The routing follows the weekly cycle and does not depend on the hour of the day (NMI 0.00). Thanksgiving (Thursday, November 22, 2012), on which demand drops to a weekend level, is routed entirely to $W_3$, the Worker for Saturdays, whereas 86\% of the forecasts on regular Thursdays are routed to $W_2$. The counterfactual analysis confirms that this decision is caused by the holiday indicators: setting the holiday and working-day indicators of Thanksgiving to those of a regular Thursday moves 79\% of its forecasts to $W_2$, and marking regular Thursdays as holidays moves 64\% of their forecasts to $W_3$ (from 8\%). This behavior, however, is specific to this run: in three of the other four runs, Thanksgiving is routed mostly like a regular Thursday. With only a handful of holidays in the training data, the Manager does not consistently learn a holiday regime, in contrast to the weekly cycle.

\begin{figure*}[tbp]
\centering
\includegraphics[width=\textwidth]{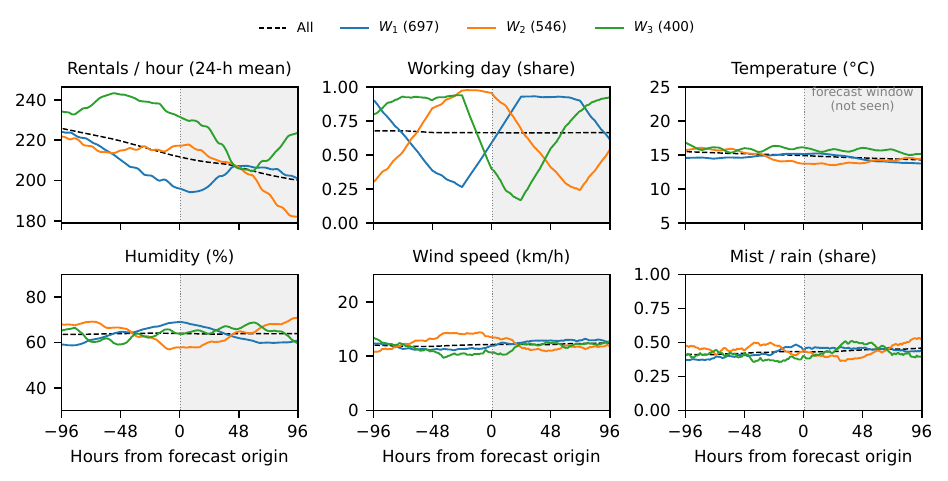}
\caption{Context conditions of the Bike Share test samples routed to each worker (one run; the same run as Figure~\ref{app:bike_timeline}). Each line is the mean over the samples routed to a worker at each hour relative to the forecast origin (dashed: all samples); the gray area is the forecast window, which the manager does not see. Rentals are shown as a trailing 24-hour mean. The workers correspond to different phases of the week (working-day share) and therefore to different demand trajectories, whereas the weather conditions are similar for all workers.}
\label{app:bike_feature_profiles}
\end{figure*}

\begin{figure}[tbp]
\centering
\includegraphics[width=\linewidth]{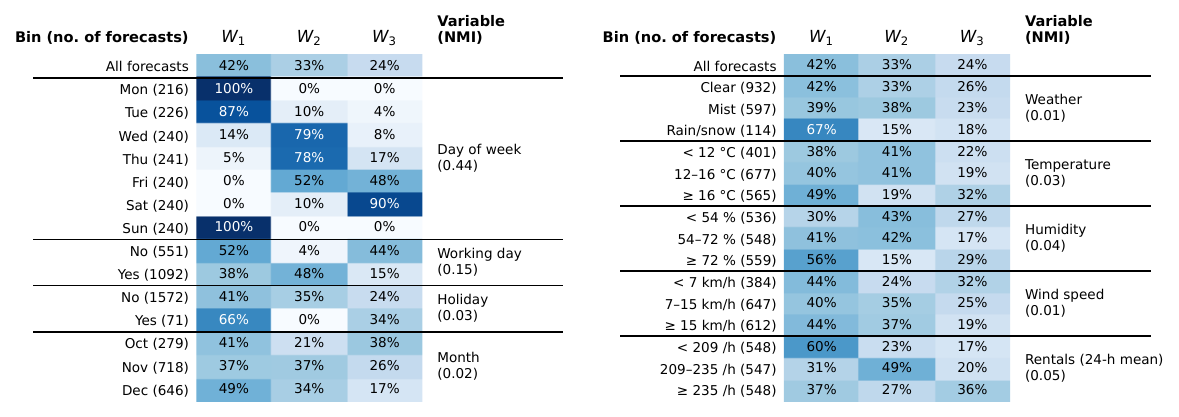}
\caption{Routing of the Bike Share test forecasts by each input variable (one run; the same run as Figure~\ref{app:bike_feature_profiles}). The variables are taken at the forecast origin; continuous variables are split into tertiles over the test period, and the number of forecasts in each bin is given in parentheses. Each row gives the percentage of the forecasts in the bin that are routed to each worker (rows sum to 100\%; first row: all forecasts). Left: day of the week, working day, holiday, and month; right: weather and recent demand. The column ``Variable (NMI)'' gives the variable of each group of rows and, in parentheses, the normalized mutual information (NMI) between its bin and the selected worker (0: unrelated; 1: the bin determines the worker). The routing depends strongly on the day of the week and the working-day indicator, and only weakly on the other variables.}
\label{app:bike_context_heatmap}
\end{figure}

\parasum{Context of the Workers.} Figure~\ref{app:bike_feature_profiles} shows the context of the samples routed to each Worker around the forecast origin. To compare the variables on a common scale, we compute, at each hour, the range of the three Worker means divided by the standard deviation of the variable over the test windows, and average it over the 192 hours. The Workers differ most in the variable that drives the routing, the working-day share (1.2 standard deviations), and accordingly face different demand trajectories (0.4). For the weather variables, which have low importance, the differences are much smaller than for the working-day share (temperature 0.4, humidity 0.4, wind speed 0.3, mist/rain 0.2; mean temperature within 2\,$^\circ$C, humidity within 10\%). Figure~\ref{app:bike_context_heatmap} relates the routing to every input variable at the forecast origin. Only the day of the week (NMI 0.44) and the working-day indicator (0.15) are strongly related to the routing; for all other variables, the NMI is at most 0.05. Some weather bins deviate from the overall Worker shares, e.g., 67\% of the forecasts made during rain or snow go to $W_1$ (42\% overall), but these bins are small and fall mostly on particular days: in the test period, most rainy hours occur on Sunday to Tuesday, the days of $W_1$. Together with the counterfactual analysis, this indicates that such deviations reflect the calendar rather than the weather itself.

\begin{figure*}[tbp]
\centering
\includegraphics[width=\textwidth]{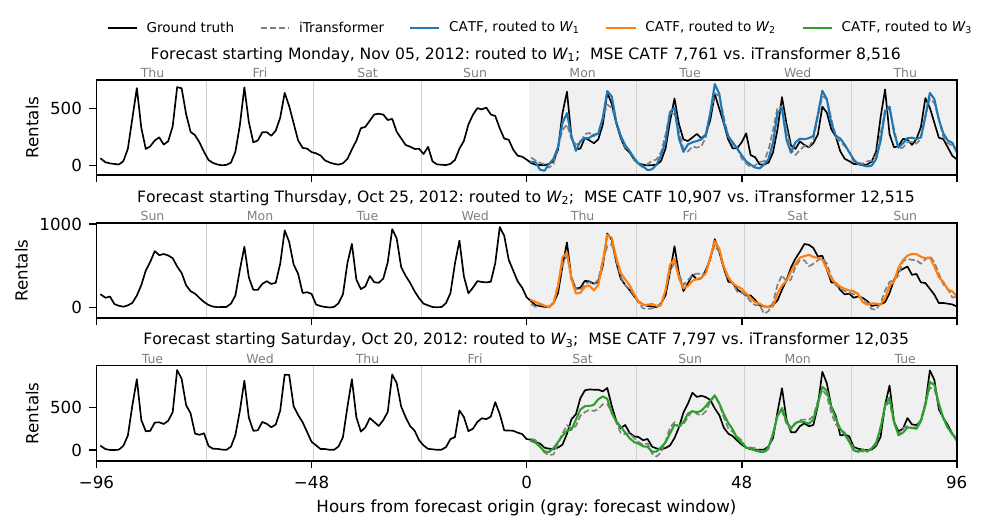}
\caption{Example forecasts on Bike Share for typical forecast origins of each worker (one run; the same run as Figure~\ref{app:bike_timeline}), together with the standalone iTransformer. For each case, the forecast with the median \sys~error among all forecasts starting at midnight on that day of the week (without holidays in the forecast window) is shown. The gray area is the forecast window. The largest difference occurs for the forecast starting on Saturday, for which $W_3$ captures the midday weekend peak that the standalone iTransformer underestimates.}
\label{app:bike_forecasts}
\end{figure*}

\parasum{Example forecasts.} Figure~\ref{app:bike_forecasts} compares typical forecasts of the three Workers with those of the standalone iTransformer. All three Workers capture the double commuting peak on working days and the single midday peak on weekends. The largest difference occurs for the forecast starting on Saturday: the standalone model underestimates the midday weekend peak, whereas $W_3$ captures it and reduces the MSE from $12{,}035$ to $7{,}797$.

\subsection{PM2.5-RV}
\label{app:context_rv}
We repeat the analysis of Section~\ref{sec:context} for the five runs of \sys-iTransformer on PM2.5-RV (global normalization, $L=L_y=96$). Here, the routing has little effect. With the same trained Workers, the learned routing (MSE $80.5 \pm 1.0$) is only slightly better than random routing with the same Worker shares ($81.1 \pm 0.7$; better in four of five runs) and an untrained Manager ($81.6 \pm 1.2$), and it is on par with the best single Worker ($80.4 \pm 1.0$). The Manager selects the most accurate Worker for 36\% of the test samples (untrained Manager: 34\%), and on the samples routed to a Worker, the other Workers are at most 5\% less accurate, i.e., the Workers barely specialize. Shuffling all context variables in the Manager input changes 47\% of the routing decisions, compared with 41\% for the untrained Manager, and no single variable stands out (19--27\% vs.\ 13--21\%). Unlike bike demand, which follows distinct patterns on working days and weekends, PM2.5 concentration varies with continuously changing meteorological conditions, so there may be no discrete context-driven regimes for the Workers to specialize in. The routing analysis therefore also indicates whether a dataset contains such regimes.


\end{document}